\documentclass{article}

\usepackage{microtype}
\usepackage{graphicx}
\usepackage{subfigure}
\usepackage{booktabs} 

\usepackage{hyperref}

\usepackage[accepted]{icml2024}

\usepackage{amsmath}
\usepackage{amssymb}
\usepackage{mathtools}
\usepackage{amsthm}

\usepackage[capitalize,noabbrev]{cleveref}

\theoremstyle{plain}
\newtheorem{theorem}{Theorem}[section]
\newtheorem{proposition}[theorem]{Proposition}

\theoremstyle{definition}

\theoremstyle{remark}

\usepackage{multirow}
\usepackage{graphicx,subfigure}
\usepackage{algorithm}
\usepackage{algpseudocode}

\usepackage{caption}
\usepackage{placeins}
\usepackage{wrapfig}
\usepackage{booktabs}       
\usepackage{amsfonts}       
\usepackage{nicefrac}       
\usepackage{microtype}      
\usepackage{xcolor}         
\usepackage{wrapfig}

\definecolor{cadmiumgreen}{rgb}{0.0, 0.42, 0.24}
\definecolor{custom}{cmyk}{0.1,0.48,0.49,0.2}
\definecolor{new}{rgb}{0.81,0.05,0.9}
\newcommand{\revision}[1]{\textcolor{black}{#1}} 

\usepackage[textsize=tiny]{todonotes}

\icmltitlerunning{CosPGD}

\begin{document}

\twocolumn[
\icmltitle{CosPGD: an efficient white-box adversarial attack for pixel-wise prediction tasks}



\icmlsetsymbol{equal}{*}
\icmlsetsymbol{dagger}{$\dagger$}

\begin{icmlauthorlist}
\icmlauthor{Shashank Agnihotri}{dagger,dws}
\icmlauthor{Steffen Jung}{dagger,mpi,dws}
\icmlauthor{Margret Keuper}{dws,mpi}

\end{icmlauthorlist}

\icmlaffiliation{dws}{Data and Web Science Group, University of Mannheim, Germany}
\icmlaffiliation{mpi}{Max-Planck-Institute for Informatics, Saarland Informatics Campus, Germany}

\icmlcorrespondingauthor{Shashank Agnihotri}{shashank.agnihotri@uni-mannheim.de}

\icmlkeywords{Machine Learning, ICML, Adversarial Attacks, Pixel-wise tasks, white-box}

\vskip 0.3in
]



\printAffiliationsAndNotice{\icmlModifiedContribution}  

\begin{abstract}
While neural networks allow highly accurate predictions in many tasks, their lack of robustness towards even slight input perturbations often hampers their deployment.
Adversarial attacks such as the seminal \emph{projected gradient descent} (PGD) offer an effective means to evaluate a model's robustness and dedicated solutions have been proposed for attacks on semantic segmentation or optical flow estimation. While they attempt to increase the attack's efficiency, a further objective is to balance its effect, so that it acts on the entire image domain instead of isolated point-wise predictions. This often comes at the cost of optimization stability and thus efficiency.  
Here, we propose CosPGD, an attack that encourages more balanced errors over the entire image domain while increasing the attack's overall efficiency.
To this end, CosPGD leverages a simple alignment score computed from any pixel-wise prediction and its target to scale the loss in a smooth and fully differentiable way. 
It leads to efficient evaluations of a model's robustness for semantic segmentation as well as regression models (such as optical flow, disparity estimation, or image restoration), and it allows it to outperform the previous SotA attack on semantic segmentation. 
We provide code for the CosPGD algorithm and example usage at \url{https://github.com/shashankskagnihotri/cospgd}.
\end{abstract}

\section{Introduction}
\label{sec:intro}
Deep Neural Networks~(DNNs) have been gaining popularity for estimating solutions to various complex tasks including numerous vision tasks like classification~\citep{alexnet, resnet, resnext, convnext,lukasik2023improving}, generative models~\cite{steffen,jung2021internalized,lukasik2022learning,jung2023happy}, image segmentation~\citep{unet, semsegzhao2017pspnet, jung2022optimizing,sommerhoff2023differentiable}, or disparity~\citep{sttr} and optical flow~\citep{flownet, flownet2, raft,schmalfuss2023distracting} estimation, due to their overall precise predictions. 
However, DNNs are inherently black-box function approximators~\citep{blackboxanalysis}, known to find shortcuts to map the input to a target~\citep{shortcut}, to learn biases~\citep{texturebias,gavrikov2024vision} and to lack robustness~\cite{szegedy2014intriguing,hoffmann2021towards}. 


\begin{figure}[t]
     \centering
     \scriptsize
     \begin{tabular}{@{}c@{\hspace{0.1cm}}c@{}}
         \includegraphics[width=3.8cm]{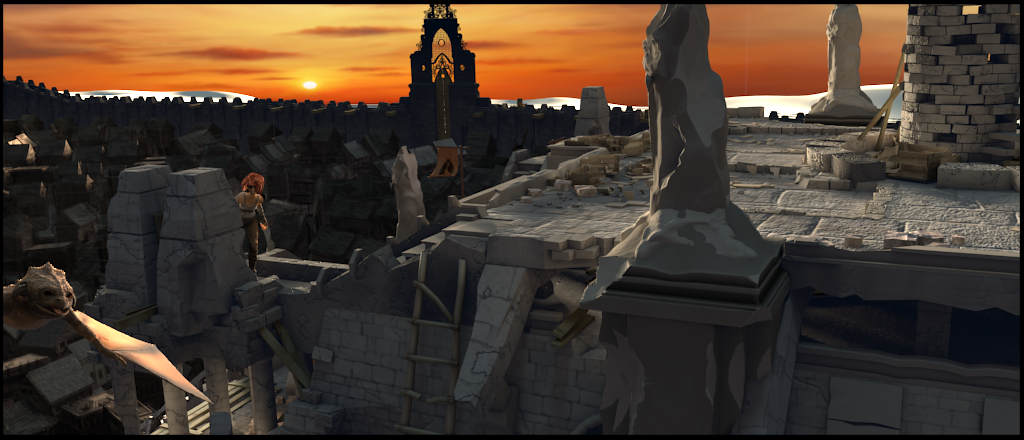}
         &\frame{\includegraphics[width=3.8cm]{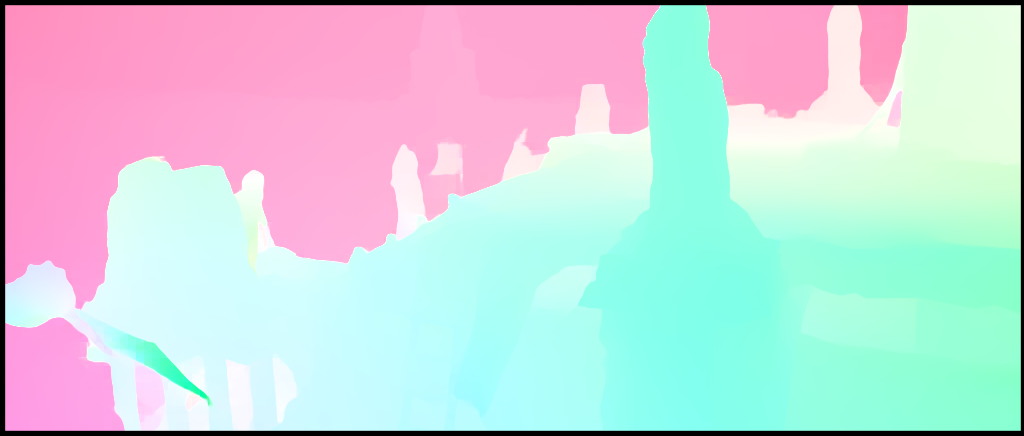}}\\
         (a) Input at $\mathit{time}=t$&      (d) Initial flow prediction\\
           \includegraphics[width=3.8cm]{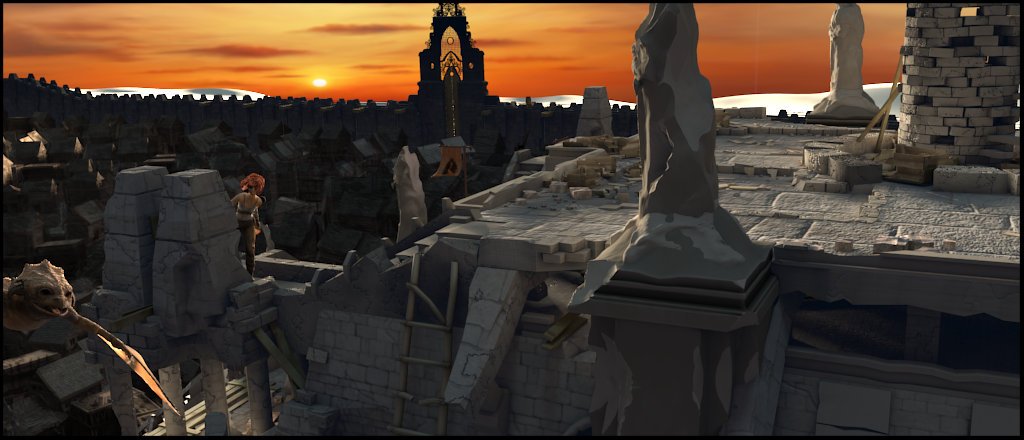}
         &
          \frame{\includegraphics[width=3.8cm]{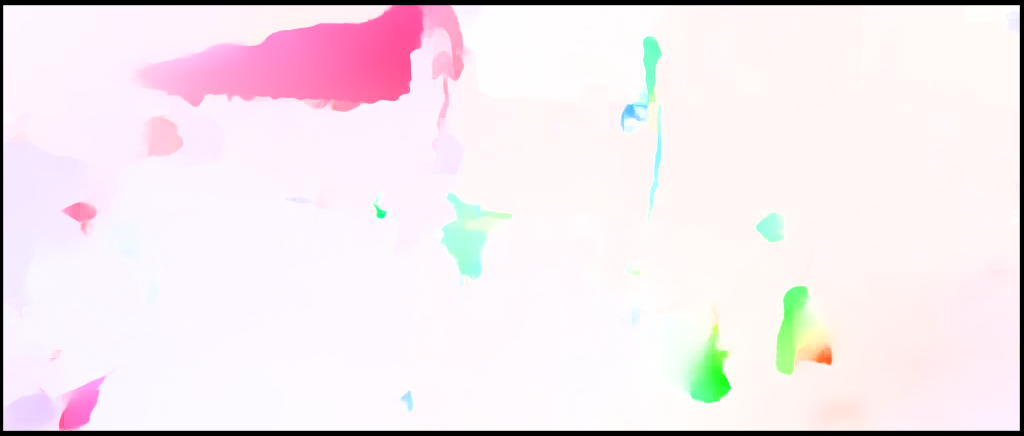}}\\
          (b) Input at $\mathit{time}=t+1$&(e) PGD, 40 iterations\\
         \includegraphics[width=3.8cm]{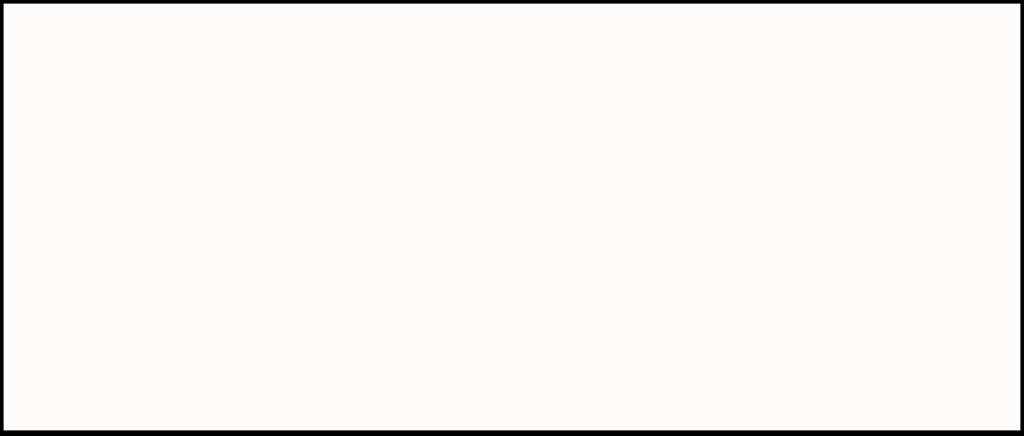}
         &
         \frame{\includegraphics[width=3.8cm]{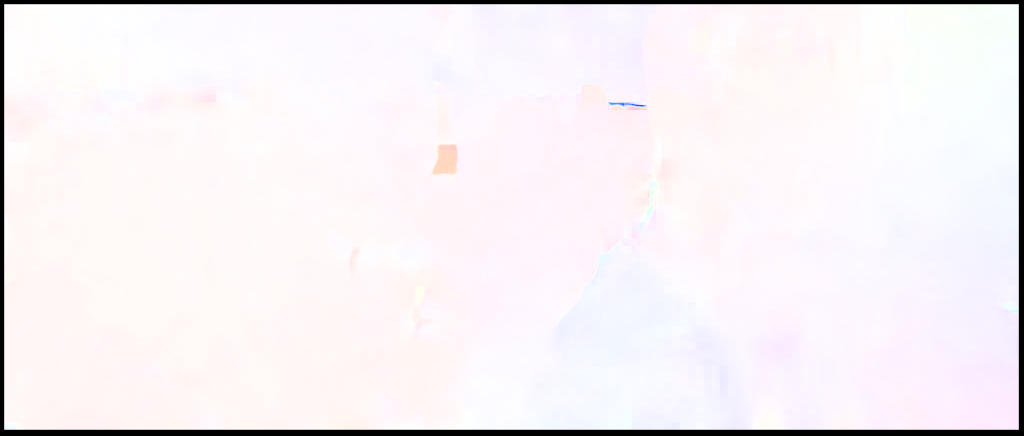}}
     \\
     (c) Target flow $\overrightarrow{0}$ & (f) CosPGD, 40 iterations
     \end{tabular}
        \caption{Optical flow predictions using RAFT~\citep{raft} on Sintel~\citep{sintel1, sintel2} validation.
        (a)~and~(b) show two consecutive frames for which the initial optical flow in (d) was predicted.
        The results of attacking the model with target $\overrightarrow{0}$ (c) are depicted in (e) for PGD and (f) for CosPGD.
        For the same perturbation magnitude and number of iterations, the proposed CosPGD alters the estimated optical flow more strongly and brings it closer to target (c).
        \label{fig:teaser}}
        \vspace{-1em}
\end{figure}

An adversarial attack adds a crafted, small (epsilon-sized) perturbation to the input of a neural network that aims to alter the prediction, thus assessing a network's robustness as in the benchmarks by \citet{croce2021robustbench,jung2023neural}. 
Due to the practical relevance to evaluating and analyzing DNN models, such attacks have been extensively studied~\citep{fgsm, pgd, apgd, pgdl2, deepfool, ifgsm,schrodi2022towards, agnihotri2023improving,grabinski2022frequencylowcut,grabinski2023fix,lukasik2023evaluation}. 

Existing approaches predominantly focus on attacking image classification models. However, arguably, the robustness of models for pixel-wise prediction tasks is highly relevant for many safety-critical applications such as motion estimation in autonomous driving or semantic segmentation. 
The application of existing attacks to pixel-wise prediction tasks such as semantic segmentation or optical flow estimation is possible in principle (e.g.~as in \citet{fgsm_segmentation}), albeit carrying only limited information since the pixel-specific loss information is not fully leveraged. 
In \autoref{fig:teaser}, we illustrate this effect for a targeted attack on optical flow estimation and show that classical classification attacks such as PGD (see \autoref{fig:teaser}(e)) only fool the network predictions to some extent: PGD tends to only fit the target (all zeros, i.e.~white) in parts of the optical flow, while a few predictions remain intact.

For semantic segmentation, \citet{segpgd} showed that harnessing pixel-wise information for adversarial attacks leads to much stronger attacks. They argue that, during the attack, the loss to be backpropagated needs to be altered such that already flipped pixel predictions are less important for the gradient computation. 
Thus, SegPGD~\citep{segpgd} makes a binary decision for each pixel based on the classification result at this location, to weigh the attack loss for incorrect and correct model predictions individually.
While this is intuitive for semantic segmentation, it can not extend to pixel-wise regression tasks by definition.
Furthermore, due to the discrete nature of the loss scaling, SegPGD faces stability issues and has to fade back in the loss of already incorrectly predicted pixels over time~\citep{segpgd}. 

In this work, we propose CosPGD, an efficient white-box adversarial attack that considers the cosine-alignment between the prediction and target for each pixel, leading to a smooth and fully differentiable attack objective. 
Due to its principled formulation, CosPGD can be used for a wide range of pixel-wise prediction tasks beyond semantic segmentation. \autoref{fig:teaser}(f) shows its effect on optical flow estimation, where, in contrast to PGD, it can fit the target at almost all locations. 
Since it leverages the (continuous) posterior distribution of the prediction to allow for a smooth and differentiable loss computation, it can significantly outperform SegPGD on semantic segmentation. 
The main contributions of this work are as follows:
\begin{itemize}
    \item{We propose CosPGD, an efficient white-box adversarial attack, that can be applied to any pixel-wise prediction task, and thus allows for an efficient evaluation of their robustness in a unified setting. 
    }
    \item{We provide theoretical and empirical proofs for the stability and spatial balancing of CosPGD during attack optimization.}
    \item{For semantic segmentation, we compare CosPGD to the recently proposed SegPGD which also uses pixel-wise information for generating attacks. CosPGD outperforms SegPGD by a significant margin.}
   
    \item{To demonstrate CosPGD's versatility, 
    we also evaluate it as a \emph{targeted} attack and as a \emph{non-targeted} attack, for both $\ell_2$ and $\ell_\infty$ bounds 
    on semantic segmentation, optical flow estimation 
    and image restoration in several settings and datasets.}
    
\end{itemize}
%
%
%
\section{Related work}
\label{sec:related}
The vulnerability of DNNs to adversarial attacks was first explored in \cite{fgsm} for image classification, proposing the Fast Gradient Sign Method (FGSM).
FGSM is a single-step (one iteration) white-box adversarial attack that perturbs the input in the direction of its gradient, generated from backpropagating the loss, with a small step size, such that the model prediction becomes incorrect. Due to its fast computation, it is still a widely used approach.
Numerous subsequent works have been directed towards generating effective adversarial attacks for diverse tasks including NLP~\citep{nlp_attack1, nlp_attack2, nlp_attack3}, or 3D tasks~\citep{3d_attack1, 3d_attack2}. Yet, the high input dimensionality of image classification models results in the striking effectiveness of adversarial attacks in this field~\citep{fgsm, jia2022multiguard}. 
A vast line of work has been dedicated to assessing the quality and robustness of representations learned by the network, including the curation of dedicated evaluation data for particular tasks~\citep{unforeseenadvasaries, commoncorruptions, naturaladversarial} or the crafting of effective adversarial attacks. 
These adversarial attacks can be image-wide or localized in a small region or patch.
These perturbations are in a small region of the image and are called Patch Attacks~(e.g.~\cite{advpatch1,scheurer2024detection}),
 while methods such as proposed in \cite{fgsm,pgd,pgdl2,apgd,deepfool,croce2020reliable,ACFH2020square,c_and_w,decoupling, mifgsm} argue in a Lipschitz continuity motivated way that a robust network's prediction should not change drastically if the perturbed image is within the epsilon-ball of the original image and thus optimize attacks globally within the epsilon neighborhood of the original input. Our proposed CosPGD follows this line of work.

White-box attacks assume full access to the model and its gradients~\citep{fgsm,pgd,pgdl2,apgd,segpgd,deepfool,proximal_splitting,mifgsm,schmalfuss2022attacking} while black-box attacks optimize perturbations in a randomized way~\citep{ACFH2020square,ilyas2018blackbox,qu2023certified}. 
The proposed CosPGD derives its optimization from PGD~\citep{pgd} and is a white-box attack.

Further, one distinguishes between \emph{targeted} attacks~(e.g.~\cite{wong2020targeted,9840784,pcfa}) that turn the network predictions towards a specific target and \emph{untargeted} (or non-targeted) attacks that optimize the attack to cause any incorrect prediction. PGD~\citep{pgd}, and CosPGD by extension, allows for both settings~\citep{pgd_target}. 

While previous attacks predominantly focus on classification tasks, only a few approaches specifically address the analysis of pixel-wise prediction tasks such as semantic segmentation, optical flow, or disparity estimation. 
For example, PCFA~\citep{pcfa} was applied to the estimation of optical flow and specifically minimizes the average end-point error ($AEE$) to a target flow field.
A notable exception of pixel-wise white-box adversarial attack is proposed in \cite{segpgd}. 
The SegPGD attack could showcase the importance of pixel-wise attacks for semantic segmentation. 
In this work, we propose CosPGD to provide a principled and efficient adversarial attack, that can be applied to a wide range of pixel-wise prediction tasks and provides stable optimization. CosPGD outperforms SegPGD by a significant margin when attacking semantic segmentation models while preserving its efficiency and extending it to other pixel-wise prediction tasks.

\section{Preliminaries}
\label{sec:preliminaries}
The projected gradient descent (PGD) \cite{pgd} attack is an iterative white box adversarial attack. It is known to be a strong attack and builds the basis for follow-up methods such as \cite{apgd}. Such methods leverage the gradients of a model's loss to create strong adversarial attacks, e.g.~the PGD update is given as
\begin{equation}
    \label{eqn:pgd_attack_1}
    \boldsymbol{X}^{\mathrm{adv}_{t+1}} = \boldsymbol{X}^{\mathrm{adv}_t}+\alpha \cdot \mathrm{sign}\nabla_{\boldsymbol{X}^{\mathrm{adv}_t}}L(f_{\theta}(\boldsymbol{X}^{\mathrm{adv}_t}), \boldsymbol{Y})
\end{equation}
\begin{equation}
    \delta = \phi^{\epsilon}(\boldsymbol{X}^{\mathrm{adv}_{t+1}} - \boldsymbol{X}^{\mathrm{clean}}), 
\end{equation}
\begin{equation}
\label{eqn:pgd_attack}
    \boldsymbol{X}^{\mathrm{adv}_{t+1}} = \phi^{r}(\boldsymbol{X}^{\mathrm{clean}}+ \delta)
\end{equation}
Here,  $L(\cdot)$ is a function (differentiable at least once) of the model prediction and the target, which defines the loss the model $f_\theta$ aims to minimize,  $\boldsymbol{X}^{\mathrm{adv}_{t+1}}$ is a new adversarial example for time step $t+1$, generated using $\boldsymbol{X}^{\mathrm{adv}_{t}}$, the adversarial example at time step $t$ and initial clean sample $\boldsymbol{X}^{\mathrm{clean}}$. 
$\boldsymbol{Y}$ is the ground truth label for non-targeted attacks and the target for targeted attacks, $\alpha$ is the step size for the perturbation ($\alpha$ is multiplied by $-1$ for targeted attacks to take a step in the direction of the target), and the function $\phi^{\epsilon}$ is clipping the $\delta$ in $\epsilon$-ball for $\ell_{\infty}$-norm bounded attacks or the $\epsilon$-projection in $l_{2}$-norm bounded attacks, complying with the $\ell_\infty$-norm or $l_2$-norm constraints, respectively. 
$\phi^{r}$ is clipping the generated example in the valid input range (usually between [0, 1]).
$\nabla_{\boldsymbol{X}^{\mathrm{adv}_t}}L(\cdot)$ denotes the gradient of $\boldsymbol{X}^{\mathrm{adv}_t}$ generated by backpropagating the loss and is used to determine the direction of the perturbation step.

Originally, PGD has been conceived to attack image classification models. For pixel-wise prediction tasks, its update in \autoref{eqn:pgd_attack_1} considers the sum of pixel-wise losses $\bar{L}$, i.e.~
\begin{align}
    \label{eqn:pgd_attack_3}
&\boldsymbol{X}^{\mathrm{adv}_{t+1}}&=&\boldsymbol{X}^{\mathrm{adv}_t}+\\&&&\alpha \cdot \mathrm{sign}\nabla_{\boldsymbol{X}^{\mathrm{adv}_t}}\sum_{i\in {H\times W}}\bar{L}\left(f_{\theta}(\boldsymbol{X}^{\mathrm{adv}_t})_i, \boldsymbol{Y}_i\right)\nonumber
\end{align}
where $i$ iterates over all positions in the prediction $f_{\theta}(\boldsymbol{X})$ with $f_{\theta}(\boldsymbol{X})$, $\boldsymbol{Y}\in\mathbb{R}^{H\times W\times M}$ for images of size $H\times W$ and $M$ output dimensions (e.g.~$M$ classes for semantic segmentation). The update in PGD thus aims to increase the overall loss maximally summing over all locations. It does not take into account that the prediction in some locations might remain correct while it further increases the loss in other locations (that might already be predicted incorrectly). 

\section{Prediction Alignment Scaling - CosPGD}
\label{sec:method}

We argue that the above formulation neglects an interesting aspect:
It does not facilitate inducing equally manipulated predictions in all locations.
This can be disadvantageous for targeted attacks, where one wants to ensure that the target is fit at all locations equally.
In particular, it is however problematic for, for example, attacks on semantic segmentation where models use cross-entropy-like losses that do not saturate.
Thus, after flipping a few point-wise label predictions, PGD-based attacks might continue to increase the overall loss even without altering any further labels.
Thus, we argue that the alignment between the current prediction and the target or ground truth has to be taken into account to efficiently compute strong adversaries.

In the following, we introduce CosPGD. Its goal is to employ a continuous pixel-wise measure of prediction alignment inside the computation of the attack update step so that the gradient-based CosPGD iterations smoothly converge to a strong adversary that acts on all pixel locations. 
The update step in CosPGD is defined as 
\begin{align}
    \label{eqn:cospgd_attack}
&\boldsymbol{X}^{\mathrm{adv}_{t+1}}=\boldsymbol{X}^{\mathrm{adv}_t}+\alpha \cdot \mathrm{sign}\nabla_{\boldsymbol{X}^{\mathrm{adv}_t}}\\
& \sum_{i\in H\times W}\mathrm{cos}\left(\psi(f_\theta(\boldsymbol{X}^{\mathrm{adv}_t})_i), \boldsymbol{Y}_{i}^{'}\right) \cdot \bar{L}\left(f_{\theta}(\boldsymbol{X}^{\mathrm{adv}_t})_i, \boldsymbol{Y}_i\right),
\end{align}

%

%

where $\psi$ is a continuously differentiable, monotonous function that can be used to normalize the model output, i.e.~we assume $\|\psi(f_{\theta}(\boldsymbol{X}))\| =1 \quad\forall f_{\theta}(\boldsymbol{X})$, and
\begin{equation}
    \mathrm{cos}(\boldsymbol{P},\boldsymbol{Y}) =  \frac{\boldsymbol{P}\cdot \boldsymbol{Y}}{\| \boldsymbol{P}\| \cdot \| \boldsymbol{Y}\|}
\end{equation}
is the cosine similarity between two vectors, in this case, a (normalized) network prediction $\boldsymbol{P}$ and the target or ground truth $\boldsymbol{Y}\in\mathbb{R}^M$. For the example of semantic segmentation, $\boldsymbol{Y}^{'}$ is one-hot encoded and therefore normalized, in other cases $\boldsymbol{Y}^{'}=\psi(\boldsymbol{Y})$. Cosine similarity provides a measure of similarity between the direction of two vectors and should therefore be well-suited to represent the alignment of the prediction with the target at the posterior level.
It scales in a fixed range [-1, 1], such that no further normalization of the scaling is needed.

As the loss in CosPGD is scaled with a pixel-wise measure of alignment between the current prediction and the target in \autoref{eqn:cospgd_attack}, the resulting gradient update emphasizes on changing those pixel-wise predictions that are correct in the current prediction.

This yields several desirable properties.   
First, it facilitates to optimize adversaries to pixel-wise tasks so that the prediction in all pixels is affected. As such, it is a stronger attack than PGD on tasks such as semantic segmentation. Further, it can be applied to pixel-wise classification and regression tasks in a principled way.
Second, the loss is scaled with a smooth scaling function, i.e.~if the prediction changes only a little, the change in the proposed alignment score will also be small, specifically
\begin{proposition}
\label{prop1}
    For any two pixel-wise network predictions $f_\theta(\boldsymbol{X})_i$ and $f_\theta(\boldsymbol{\bar{X}})_i \in\mathbb{R}^{M}$, a target $\boldsymbol{Y}_i\in\mathbb{R}^{M}$ and a continuously differentiable function $\psi:\mathbb{R}^{M}\rightarrow  \mathbb{R}^{M}$ with $\|\psi(f_{\theta}(\boldsymbol{X}))\| =1 \quad\forall f_{\theta}(\boldsymbol{X})$, it is 
  %
     \begin{align*}
    & d\cdot\|f_\theta(\boldsymbol{X})_i - f_\theta(\boldsymbol{\bar{X}})_i\|\geq\qquad\\
    &\qquad \|\mathrm{cos}\left(\psi(f_\theta(\boldsymbol{X})_i), \boldsymbol{Y}_{i}^{'}\right) -  \mathrm{cos}\left(\psi(f_\theta(\boldsymbol{\bar{X}})_i), \boldsymbol{Y}_{i}^{'}\right)\|
    \end{align*}
    for a real, constant $d\geq 0$.
\end{proposition}
The proof is given in the appendix. As a result of the above proposition, the gradient in \autoref{eqn:cospgd_attack} will change smoothly over the attack iterations for a sufficiently small step-size $\alpha$ and allow for fast convergence properties, i.e.~CosPGD should provide strong adversaries with relatively few iterations while providing a balance over the pixel locations.   

\paragraph{Untargeted versus Targeted Attacks.}
Untargeted attacks intend to drive the model's predictions away from the model's intended target (ground truth). Specifically, for non-targeted attacks, CosPGD, therefore, scales the loss pixel-wise in proportion to the pixel-wise predictions' similarity to the ground truth, while also accounting for the decrease in similarity over iterations.
Using cosine similarity as an alignment measure, pixels at which the network predictions are closer to the intended target (ground truth), have a higher similarity (approaching $1$) and thus higher loss.
Pixels with lower similarity, have a lower loss but are not rendered benign.
In contrast, for the targeted setting, 
the attack aims to drive predictions towards the target at all locations, such that pixels at which the network predictions are closer to the target and have higher similarity should have a lower loss than pixels with lower similarity.

To scale the loss by the dissimilarity of the prediction to the target prediction, for targeted settings, the targeted CosPGD update step is given by Eqn~\ref{eqn:cospgd_attack_2} in analogy to Eqn~\ref{eqn:cospgd_attack}.
\begin{align}
\label{eqn:cospgd_attack_2}
&\boldsymbol{X}^{\mathrm{adv}_{t+1}}=\boldsymbol{X}^{\mathrm{adv}_t}-\alpha \cdot \mathrm{sign}\nabla_{\boldsymbol{X}^{\mathrm{adv}_t}}\\
& \sum_{i}\left({\small 1-\mathrm{cos}\left(\psi(f_\theta(\boldsymbol{X}^{\mathrm{adv}_t})_i), \boldsymbol{Y}_{t}^{'}\right)}\right) \cdot\bar{L}\left(f_{\theta}(\boldsymbol{X}^{\mathrm{adv}_t})_i, \boldsymbol{Y}_t\right) \nonumber
\end{align}

\paragraph{Choice of $\psi$ and Algorithm Description.}
In \autoref{eqn:cospgd_attack}, we require $\psi$ to be monotonically increasing, differentiable, and, to ensure smooth convergence, smooth. To obtain a distribution over the predictions, 
we calculate the softmax of the predictions before taking the argmax
\begin{align}
    \label{eqn:sigmoid_pred}
    \psi(f_\theta(\boldsymbol{X})) = \textit{softmax}
    (f_\theta(\boldsymbol{X})), \quad \\ \text{where}, \quad \textit{softmax}(x_{i}) = \frac{\exp(x_i)}{\sum_j \exp(x_j)}.
\end{align}
Thus, in Algorithm~\ref{alg:cospgd} (given in Appendix~\ref{appendix:algorithm}) and \autoref{eqn:cospgd_attack}, $\psi$ is the softmax function. 
In the case of semantic segmentation, we obtain the distribution of the target $\boldsymbol{Y}_i$ for every point $i$ by generating a \textit{one-hot encoded vector} of the label (i.e.~encoding the argmax label) while we also apply softmax to compute $\boldsymbol{Y}_{i}^{'}$ from $\boldsymbol{Y}_{i}$ from continuous targets, e.g.~for optical flow or disparity estimation. One-hot encoding and softmax to represent $\boldsymbol{Y}_i$ are summarized by function $\Psi^{'}$  in Algorithm~\ref{alg:cospgd}.
$\boldsymbol{X}^\mathrm{adv}$ is initialized to the clean input sample $\boldsymbol{X}^{\mathrm{clean}}$ with added randomized noise in the range $[-\epsilon, +\epsilon]$, $\epsilon$ being the maximum allowed perturbation.
Over attack iterations $\boldsymbol{X}=\boldsymbol{X}^{\mathrm{adv}_t}$, the adversarial example generated at iteration $t$, such that $t\in[0, T)$, where $T$ is the total number of attack iterations.

%
%
%
\paragraph{Loss Scaling in Previous Approaches.}
When optimizing $\delta$ for an adversarial attack for semantic segmentation, \citet{segpgd} have argued before that pixels which are already misclassified by the model are less relevant than pixels correctly classified by the model, because the intention of the attack is to make the model misclassify as many pixels as possible while perturbing the $\delta$ inside the $\epsilon$-ball. As a consequence, they make a hard decision based on each pixels argmax prediction as of whether it is taken into account for attack computation. 
In \cite{segpgd}, the PGD update from \autoref{eqn:pgd_attack_3} is thus modified to
\begin{align}
\mathrm{sign}\nabla_{\boldsymbol{X}^{\mathrm{adv}_t}}\biggl((1-\lambda)\sum_{i\in P^T} \bar{L}\left(f_{\theta}(\boldsymbol{X}^{\mathrm{adv}_t})_j, \boldsymbol{Y}_j\right)  + \nonumber \\\lambda\sum_{k\in P^F} \bar{L}\left(f_{\theta}(\boldsymbol{X}^{\mathrm{adv}_t})_k, \boldsymbol{Y}_k\right)
\biggr),
\end{align}
where  $P^T$
is the set of correctly classified pixels and  $P^F$ 
is the set of wrongly classified pixels, $\lambda$ is a scaling factor between the two parts of the loss that is set heuristically, and $\boldsymbol{Y}$ is the one-hot encoded ground truth for semantic segmentation. See their equation (4) for details. 

For positive $\lambda$ and for categorical labels (i.e.~$\boldsymbol{Y}^{'}$ one-hot encoded), we can rewrite the SegPGD update as
\begin{align}
\mathrm{sign}\nabla_{\small \boldsymbol{X}^{\mathrm{adv}_t}}\biggl(\sum_{i}& \left(1- \left|\lambda - {\small \frac{|(\mathit{argmax}(f_\theta(\boldsymbol{X}^{\mathrm{adv}_t})_i)-\boldsymbol{Y}_{i}^{'}|}{2}}\right|\right) \nonumber \\ & \cdot \bar{L}\left(f_{\theta}(\boldsymbol{X}^{\mathrm{adv}_t})_i, \boldsymbol{Y}_i\right)\biggr) 
\label{eqn:segpgd}
\end{align}
for all locations $i\in P^T\cup P^F$, 
i.e.~$|\lambda - |(argmax(f(\boldsymbol{X}^{\mathrm{adv}_t}))-\boldsymbol{Y}|/2|$ equals $1-\lambda$ for incorrect predictions, it equals $\lambda$ for correct predictions.

Thus, the approach by \citet{segpgd} resembles a discrete approximation of the proposed CosPGD. Yet, the discrete nature of this weighting scheme has several disadvantages: First, it limits SegPGD to applications where the correctness of the prediction can be evaluated in a binary way, and it disregards the actual prediction scores. For pixel-wise regression tasks (like optical flow, or image reconstruction) there is no absolute measure of correctness, so SegPGD can not be directly applied.
Second, 
as the number of misclassified pixels increases, the attack loses effectiveness if it only focuses on correctly classified pixels in a binary way. The $\lambda$ scaling in \cite{segpgd} has been proposed as a heuristical remedy. 
It scales the loss over iterations such that the impact of the proposed scheme decays over time. 
At the end of the attack iterations, $\lambda \approx 1/2$.
This avoids the concern of the attack becoming benign after a few iterations, yet it fades out the effect of SegPGD and may reduce its efficiency.
CosPGD, operating on continuous predictions, does not require such a heuristic.

\begin{figure}[!ht]
    \centering 
  \includegraphics[width=\linewidth]{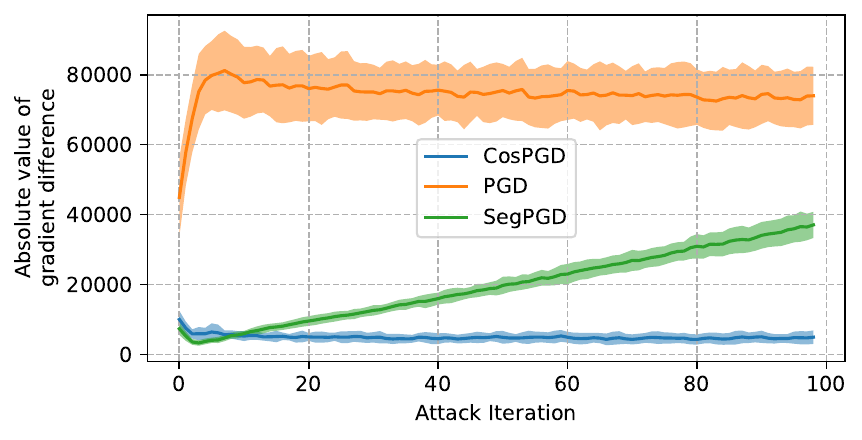}
  \includegraphics[width=\linewidth]{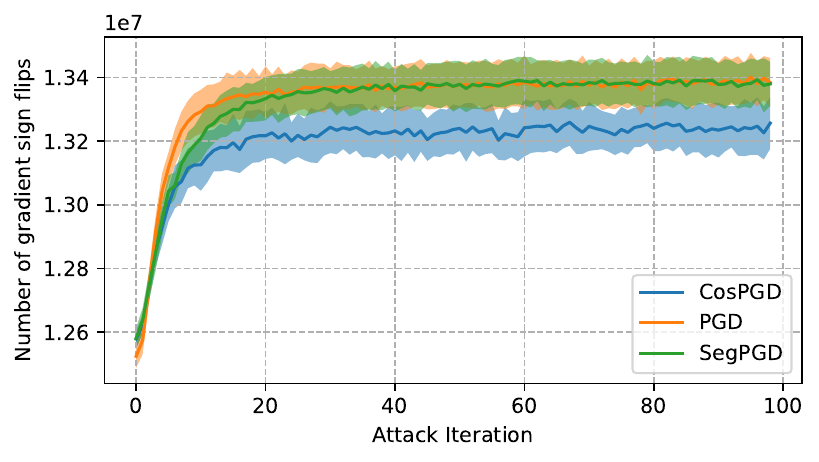}
\caption{%
Change in pixel-wise image gradients over attack iterations on DeepLabV3 performing semantic segmentation on PASCAL VOC 2012 validation subset.
We observe that the absolute difference between gradient values (top) is larger for PGD and increasing for SegPGD, while being stable for CosPGD.
Further, CosPGD has fewer changes in gradient direction over attack iterations (bottom) compared to PGD and SegPGD.
This shows CosPGD is more stable during optimization compared to PGD and SegPGD.
\vspace{-2em}
} \label{fig:semseg:nontarget:l_infinity:change_grad}
\end{figure}

Last, but maybe most importantly, the scaling based on discrete labels is not smooth, i.e. the $\mathit{argmax}$ operation in \autoref{eqn:segpgd} is not differentiable, such that, during the iterations, the direction of the gradient update can fluctuate, potentially leading to slower convergence of the SegPGD attack, compared to the proposed CosPGD. 
We show empirical evidence for this issue in \cref{fig:semseg:nontarget:l_infinity:change_grad} where we report the change in gradients and their directions during the attack optimization for PGD, SegPGD and the proposed CosPGD.
%
%
%
%
%
%
\section{Experiments}
\label{sec:exp}
To demonstrate the wide applicability of CosPGD, we conduct our experiments on distinct downstream tasks: semantic segmentation, optical flow estimation, and image restoration.
For semantic segmentation, we compare CosPGD to SegPGD and PGD and empirically validate its improved stability over the attack iterations. Further, we verify that CosPGD indeed encourages the attack to act on the entire image domain, with quantitative and qualitative results on non-targeted attacks on semantic segmentation and targeted attacks on optical flow. For optical flow estimation and other tasks (such as image deblurring and image denoising), we compare CosPGD to PGD in the main paper. The subsequent experiments provide evidence of CosPGD being a strong adversarial attack in diverse tasks and setups. In the main paper, we report $\ell_{\infty}$-norm constrained attacks with $\epsilon\approx\frac{8}{255}$ for CosPGD, SegPGD, and PGD. 
For $\alpha$, we follow \cite{segpgd} and set the step size to $\alpha=0.01$ \revision{(please refer to Appendix~\ref{subsec:exp_results:deeplabl2} for an ablation study)}. 
Further evaluations such as for different $\epsilon$ and $\alpha$ values for $\ell_{\infty}$ (\cref{subsubsec:appendix:semseg:segformer:epsilon_ablation}) and $\ell_2$ bounded attacks (\cref{subsec:exp_results:l_2}), \textbf{CosPGD for Adversarial Training} (\cref{subsec:appendix:adversarial_training}), Transfer Attacks (\cref{subsec:appendix:semseg:transfer_attacks}) including attacks on SAM~\cite{segment_anything} (\cref{subsec:appendix:semseg:attacking_sam}), Attack on Robust Models (\cref{subsec:appendix:semseg:defense_methods}),
comparison of CosPGD to recently proposed PCFA for optical flow estimation over various architectures (\cref{subsec:appendix:PCFA})
and Image Denoising (\cref{subsec:appendix:image_restoration}), are provided in the Appendix, \cref{tab:all_experiments} provides an overview. 
Please also refer to the Appendix~\ref{sup:details} for all details on the experimental setup. 

\label{subsec:exp:setup}

%
\subsection{Stability during Attack Optimization}
We evaluate the stability of CosPGD on semantic segmentation PASCAL VOC 2012~\citep{pascal-voc-2012}.
\cref{fig:semseg:nontarget:l_infinity:change_grad}(top) shows the change in gradients (i.e.~the absolute distance between gradients in two subsequent iterations) due to PGD, SegPGD and CosPGD over 100 iterations.
Both PGD and CosPGD gradients change constantly over time, with PGD having much stronger change.
\begin{figure}[t]
\vspace{-1em}
\scalebox{0.9}{
\begin{tabular}{@{}c@{}}
\scriptsize 
DeepLabV3  \\
  \includegraphics[width=\columnwidth]{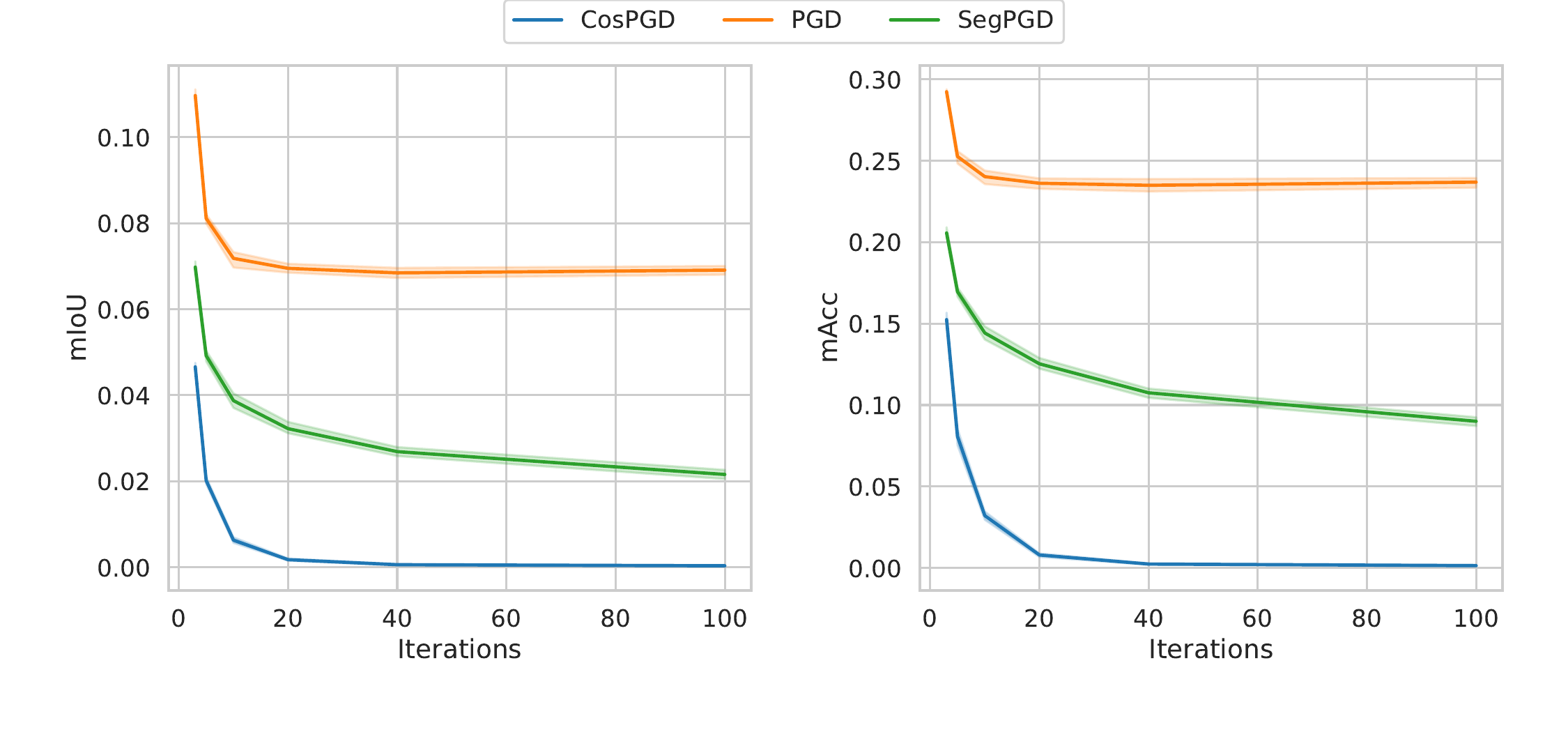}
    \vspace{-0.5cm}\\

\scriptsize PSPNet\\
  \includegraphics[width=\columnwidth]{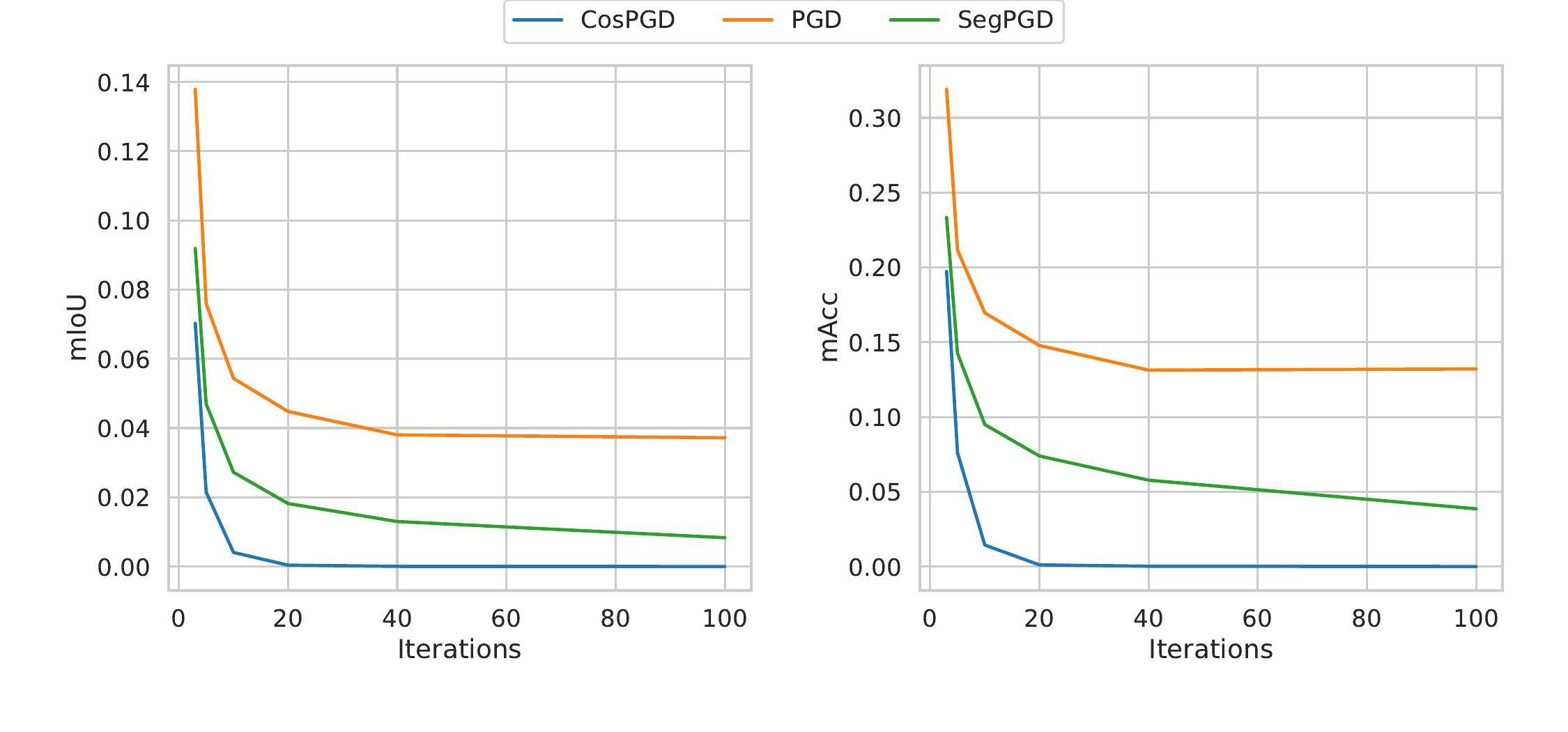}
  \vspace{-0.5cm}
\end{tabular}
}
\caption{CosPGD versus PGD and SegPGD ($\ell_{\infty}$-norm constrained) for semantic segmentation on PASCAL VOC2012 validation set on DeepLabV3  and PSPNet. CosPGD outperforms competing attacks even in early iterations by a large margin. See also \autoref{tbl:exp:semseg_pgd} in Appendix~\ref{subsec:appendix:semseg}. 
\label{fig:semseg:semseg_pgd}
\vspace{-2.5em}
}

\end{figure}
Yet, as expected, the change in gradients of SegPGD increases over the iterations, potentially leading to oscillations in the optimization.
To further analyze the effect on the optimization, \cref{fig:semseg:nontarget:l_infinity:change_grad} (bottom) shows the respective change in gradient direction (note that PGD, SegPGD, and CosPGD update all consider the sign of the gradient).
The evaluation verifies that the CosPGD updates are more stable over the iterations, such that we can expect faster convergence,~i.e.~a stronger attack at fewer iterations.

An indication of the potential benefit can be seen for example in Table~\ref{tbl:exp:semseg_pgd} (Appendix), where we observe that at low attack iterations~(iterations=3) SegPGD implies that PSPNet is more adversarially robust than DeepLabV3.
However, after more attack iterations (iterations$\geq$5), SegPGD reveals that DeepLabV3 is more robust than PSPNet.
Contrary to this, CosPGD even at low attack iterations correctly predicts DeepLabV3 to be more robust than PSPNet.
This is an insight that CosPGD provides with considerably fewer iterations, thus lower overall computation time, while compute costs per iteration are comparable, see \autoref{tbl:exp:time} (Appendix).

\begin{figure}[!ht]
    \centering 
    \begin{tabular}{@{}c@{\hspace{0.04cm}}c@{\hspace{0.04cm}}c@{\hspace{0.04cm}}c@{}}
    \scriptsize
    Ground Truth&  \scriptsize{PGD}&  \scriptsize{SegPGD} &  \scriptsize{CosPGD}\\
    &  \scriptsize{ mIoU$=6.79\%$}&  \scriptsize{mIoU$=2.69\%$} &  \scriptsize{mIoU$=0.08\%$}\\
  \includegraphics[width=0.118\textwidth]{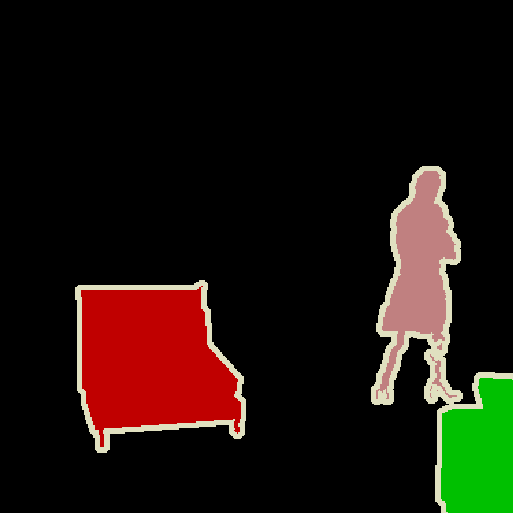}&
  \includegraphics[width=0.118\textwidth]{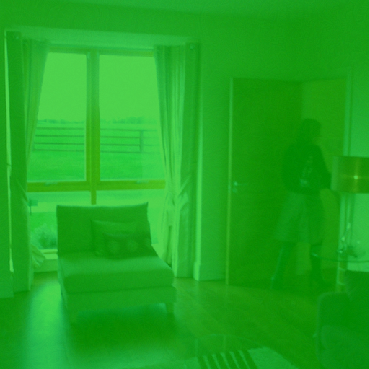}&
  \includegraphics[width=0.118\textwidth]{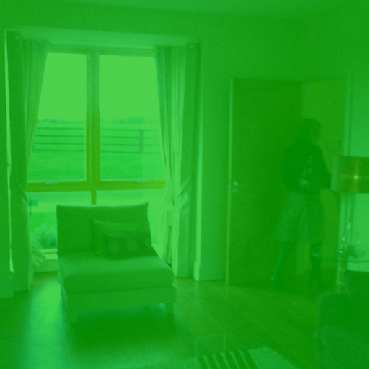}&
  \includegraphics[width=0.118\textwidth]{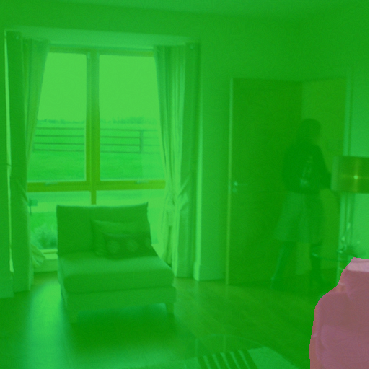}\\
  \includegraphics[width=0.118\textwidth]{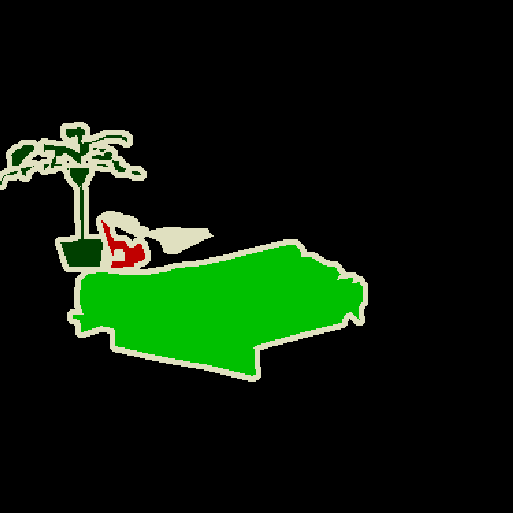}&
  \includegraphics[width=0.118\textwidth]{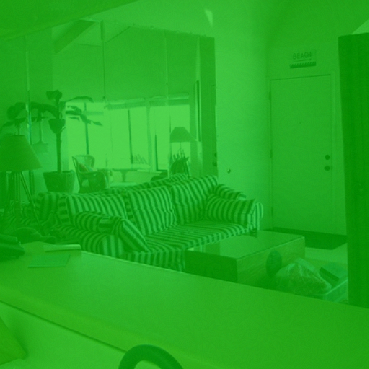}&
  \includegraphics[width=0.118\textwidth]{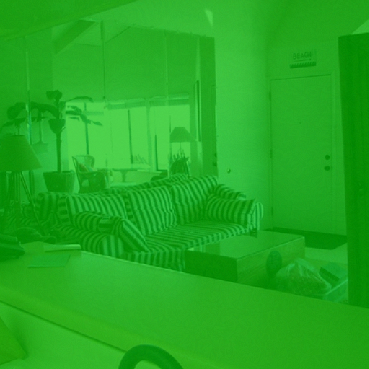}&
  \includegraphics[width=0.118\textwidth]{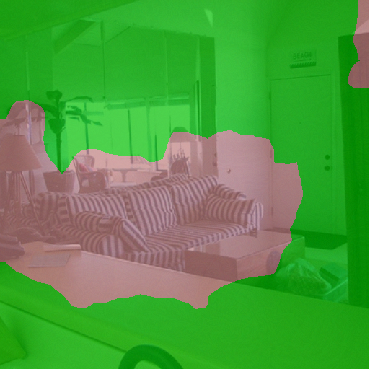}
\end{tabular}

\caption{Example predictions of DeepLabV3 on PASCAL VOC 2012 val set after $\ell_\infty$ PGD, SegPGD, and CosPGD attacks with 40 iters.
The ground truth segmentations are given on the left.
Both PGD and SegPGD are able to successfully change most of the predicted labels to one of the ground truth labels (here in green). Yet, the region with this label is predicted correctly.
Here, only CosPGD also changes the prediction in this region to a third class. 
}
\label{fig:deeplab_comparison}
\end{figure}
\subsection{Spatial Balancing of the Attack}
In the following, we show empirically that CosPGD encourages the attack to alter predictions over the entire image domain while PGD and SegPGD are weaker in this respect. 
\vspace{-1em}
\paragraph{Semantic Segmentation.}
\label{subsec:exp:semseg}
We first discuss the spatial balancing of CosPGD for untargeted attacks on semantic segmentation on PASCAL VOC2012, the standard setting evaluated in \cite{segpgd}. 

Therefore, we consider the mean Intersection over Union (mIoU) and mean accuracy (mACC) over the attack iterations as reported in \autoref{fig:semseg:semseg_pgd}. 
The first observation is that CosPGD yields a much stronger attack compared to PGD or SegPGD for both DeepLabV3~\cite{deeplabv3} and PSPNet~\cite{semsegzhao2017pspnet}. Second, we observe that CosPGD pushes the mIoU to values close to zero even in the first attack iterations, meaning that almost all pixel labels are flipped, while the mIoU for PGD stagnates at a high level as it decreases slowly for SegPGD, leading to significantly higher mIoUs even after 100 iterations, that for CosPGD.
%
%
%

For example in Figure~\ref{fig:deeplab_comparison} after 40 attack iterations, all attacks are considerably fooling the network into making incorrect predictions.
However, once the dominant class label is changed by SegPGD or PGD, they do not further optimize over small regions of correct predictions.
In contrast, CosPGD successfully fools the model into making incorrect predictions even in these small regions by either swapping the region prediction with an already existing class or forcing the model into predicting a different class. 

%
PGD can bring down the $mIoU$ of DeepLabV3 to 6.79\%.
SegPGD, by na\"{i}vely utilizing the pixel-wise segmentation error, deteriorates the model performance further to 2.69\%. However, CosPGD can fool the network into making incorrect predictions for almost all pixels, bringing down the model performance to almost 0\% after 100 iterations.
%

\begin{figure}[t]
    \centering 
  \includegraphics[width=\linewidth]{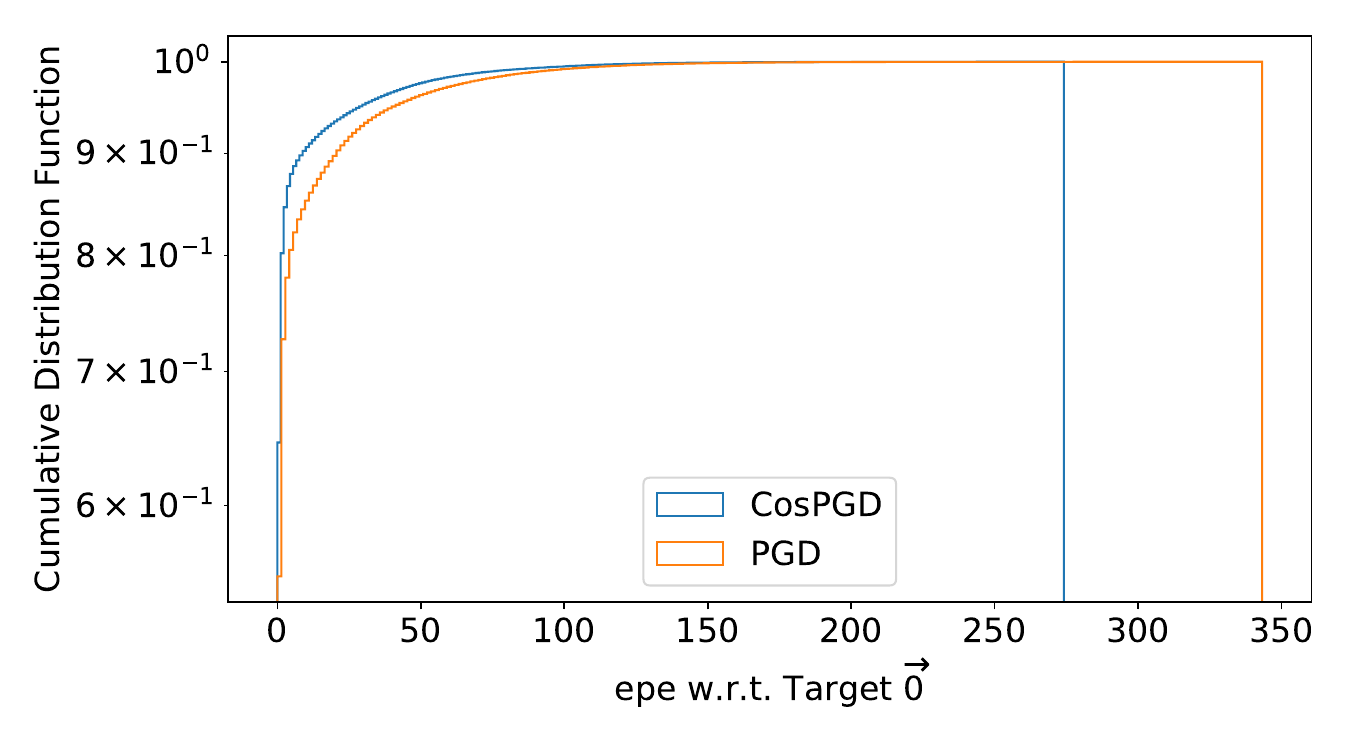}
\caption{Comparing the distributions of epe w.r.t.~Target flow $\overrightarrow{0}$ after $\ell_{\infty}$-norm constrained targeted 40 iterations CosPGD and PGD attacks on RAFT for optical flow estimation over KITTI-2015 validation dataset.
A lower epe w.r.t.~Target flow is desirable.
We observe that CosPGD can reduce the gap to Target for more pixels than the PGD attack.
Moreover, the highest epe w.r.t.~Target after a CosPGD attack is significantly lower than after a PGD attack.
\label{fig:optical:target:l_infinity:epe_distribution}
}
\end{figure}

\paragraph{Optical Flow.}
\label{subsec:exp:optical}
The evaluation of whether an attack alters the prediction in all regions is less trivial to conduct than for semantic segmentation, since there is no absolute measure of correctness. Therefore, in \cref{fig:optical:target:l_infinity:epe_distribution}, we evaluate CosPGD versus PGD for targeted attacks on optical flow (using RAFT~\cite{raft}) on the KITTI-2015 validation set such that we see how many of the point-wise flow predictions have an end point error (epe) to the target that is below a certain threshold. Ideally, we would see a curve that is rising to the maximum value very quickly, indicating that all predictions are very close to the target. \cref{fig:optical:target:l_infinity:epe_distribution} indicates that CosPGD achieves to bring more pixel-wise predictions very close to the target whereas only few predictions have larger epe. For PGD, more predictions remain with higher epe to the target.
SegPGD can not directly be compared to in this regard, since it is conceived for semantic segmentation and requires an absolute measure of correctness (i.e. is the predicted label correct).

\begin{figure}[!h]
    \centering 
  \includegraphics[width=0.9\linewidth]{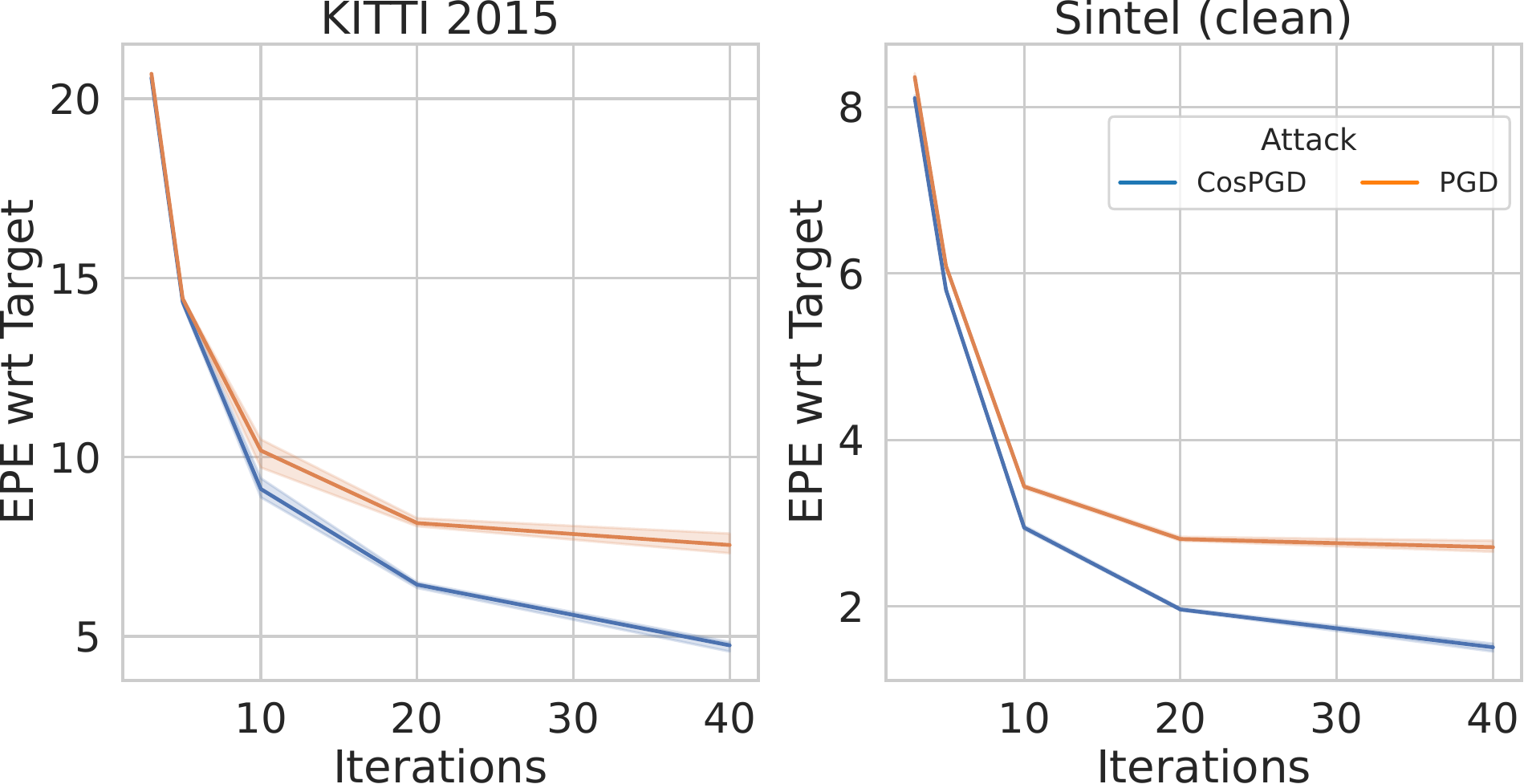}
\caption{Comparison of performance of CosPGD to PGD for optical flow estimation over KITTI-2015~(left) and Sintel~(clean $\rightarrow$ right)
 validation datasets as $\ell_{\infty}$-norm constrained targeted attacks using RAFT.
CosPGD is a stronger targeted attack than PGD for optical flow. We also report these results in \autoref{tbl:exp:optical_pgd} in Appendix~\ref{subsec:appendix:optical}. \label{fig:optical:target:l_infinity}
}
\end{figure}

A comparison of CosPGD to PGD in terms of epe over the iterations is shown in \autoref{fig:optical:target:l_infinity}.
Here, we quantitatively observe better performance of CosPGD compared to PGD.
As this is the targeted setting, we intend to close the gap between the target prediction and the model predictions, thus a lower $epe$ of the model prediction w.r.t.~the target prediction is desired. As the attack iterations increase, across datasets, CosPGD can significantly fool the network into making predictions closer to the target, bringing down the $epe$ to as low as $1.55$ for Sintel~(final) (see Appendix \ref{subsec:appendix:optical}).

\begin{figure*}[t]
    \centering 
    \scalebox{1}{
    \begin{tabular}{@{}c@{}c@{}c@{}}
  \includegraphics[width=0.32\linewidth]{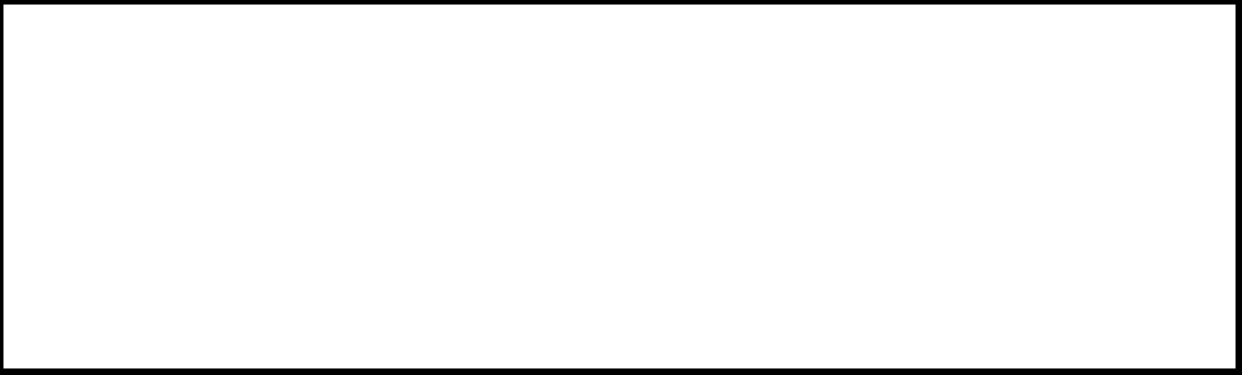}
  &
  \frame{\includegraphics[width=0.32\linewidth]{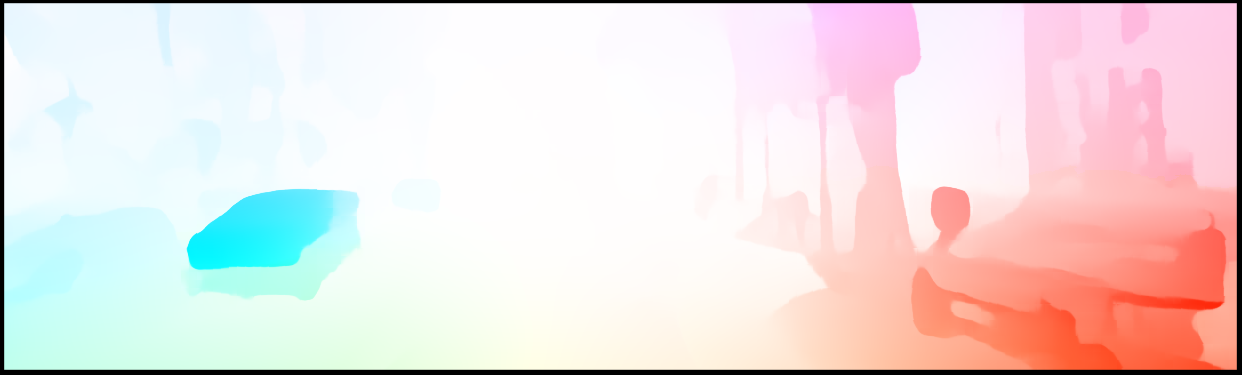}}
&
  \frame{\includegraphics[width=0.32\linewidth]{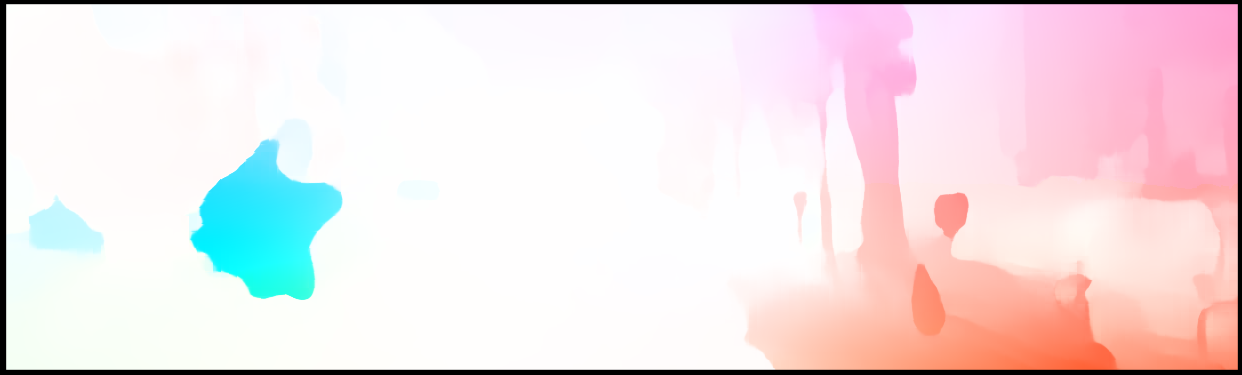}}
\\
(a) Target flow & (b) PGD 5~itrs $epe=14.42$ &(c) PGD 40~itrs $epe=7.32$ 
\\
  \frame{\includegraphics[width=0.32\linewidth]{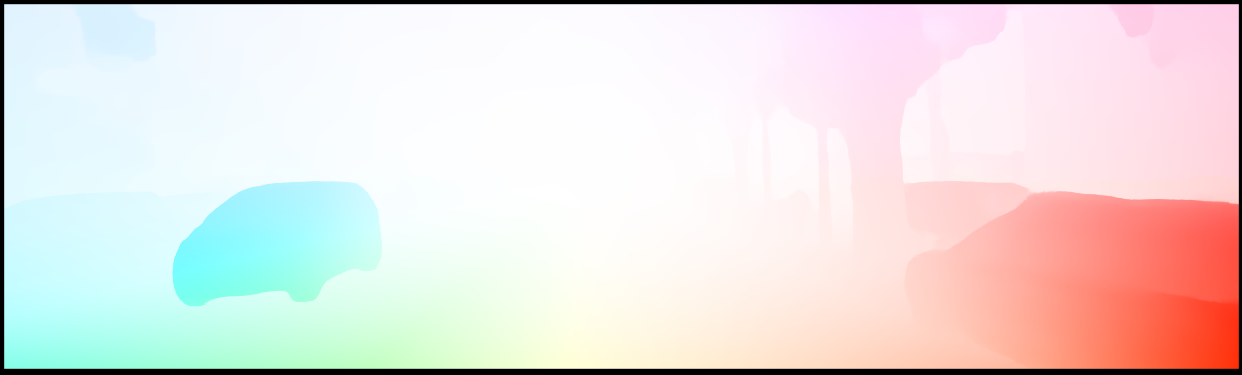}}
&
  \frame{\includegraphics[width=0.32\linewidth]{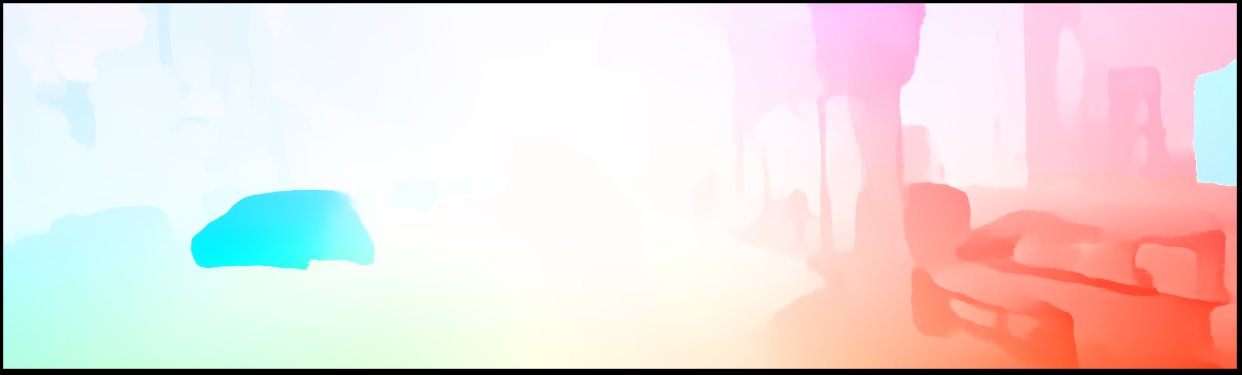}}
%
&
 \frame{ \includegraphics[width=0.32\linewidth]{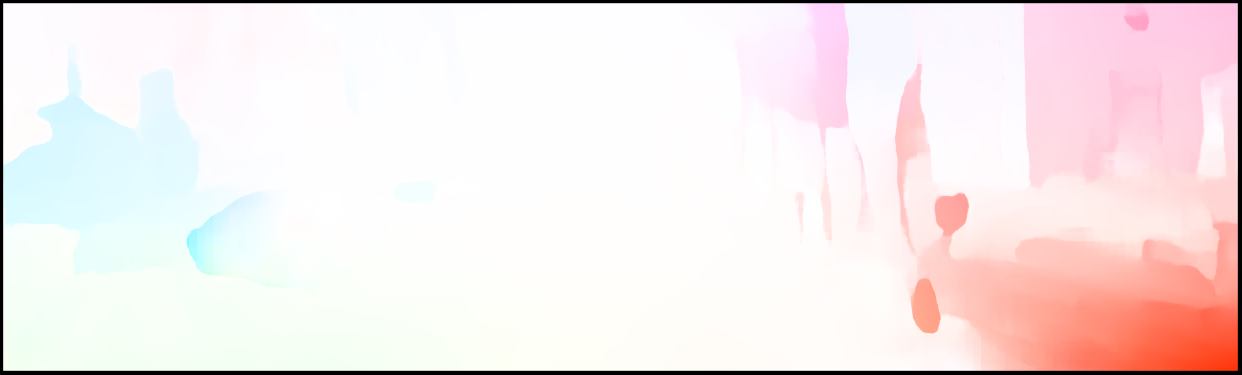}}
  \\
  (d) Initial~flow~$epe=31.1$&
  (e) CosPGD~5~itrs~$epe=14.28$&
  (f) CosPGD~40~itrs~$epe=4.84$\\
\end{tabular}
}
\caption{Comparing PGD and CosPGD as a targeted $\ell_{\infty}$-norm constrained attack on RAFT using KITTI15 validation set over various iterations. (a) shows the targeted prediction, a $\overrightarrow{0}$, and (d) shows the initial optical flow estimation by the network before adversarial attacks.
EPEs between the target and the final prediction are reported, thus lower epe is better.
(b) and (c) show flow predictions after PGD attack over 5 and 40 iterations respectively, while figures (e) and (f) show flow predictions after CosPGD attack over 5 and 40 iterations respectively. CosPGD significantly reduces the gap to target (a).}
\label{fig:kitti_comparison}
\end{figure*}

We qualitatively observe in \autoref{fig:kitti_comparison} that the initial optical flow estimation by the model (which is substantially different to the target) is only moderately changed when the model is attacked with PGD.
As the attack was designed for classification tasks, the model is not substantially fooled even as the intensity of the attack is increased to 40 iterations.

\autoref{fig:kitti_comparison}(b), shows qualitatively that the model predictions are not significantly different from the initial predictions.
The shape of the moving car is preserved to a considerable extent.
The limited effectiveness of the PGD attack is further highlighted by increasing attack iterations to 40 (see \autoref{fig:kitti_comparison}(c)).
Here, some initial predictions are still preserved, for example, the bark of the tree. 
This is in contrast to when the model is attacked with CosPGD, a method that utilizes pixel-wise information.
In \autoref{fig:kitti_comparison}(e), we observe that even at a small number of attack iterations (5), the model predictions are significantly different from the initial predictions, especially in the background and the shape of the moving car.
The model is incorrectly predicting the motion of the pixels around the moving car.
At high attack intensity, as shown in \autoref{fig:kitti_comparison}(f) with 40 iterations, the model's optical flow predictions are significantly inaccurate and exceedingly different from the initial predictions and very close to the target of $\overrightarrow{0}$.
The model fails to differentiate the moving car from its background, moreover, the bark of the tree has completely vanished.
In a real-world scenario, this vulnerability of the model to a relatively small perturbation~($\epsilon=\frac{8}{255}$) could be hazardous. 
\revision{CosPGD provides us with this new insight.}
A similar observation is made for the Sintel dataset as shown in \autoref{fig:teaser}.
The benefit of CosPGD over PGD for optical flow can be quantitatively seen in \autoref{fig:optical:target:l_infinity} and \autoref{tbl:exp:optical_pgd} in Appendix~\ref{subsec:appendix:optical}. 
%
%
\subsection{Benchmarking on Further Tasks and Settings}
\vspace{-2em}
\label{subsec:exp:gopro}
\noindent\paragraph{Semantic Segmentation.}
We observed the strength of CosPGD as a $\ell_{\infty}$-norm constrained attack in Figures~\ref{fig:semseg:semseg_pgd} \& \ref{fig:deeplab_comparison}.
Furthermore, we show that the improved performance of CosPGD is not limited to $\ell_{\infty}$-norm constrained attacks.
\cref{fig:all_deeplab_l2} in Appendix~\ref{subsec:exp_results:l_2} demonstrates the versatility of CosPGD as an $\ell_{2}$-norm constrained attack.

We observe that across $\ell_{p}$-norm constraints, the gap in performance of CosPGD w.r.t~other adversarial attacks significantly increases when increasing the number of attack iterations.
This demonstrates that CosPGD can utilize the increase in attack iterations best and highlights the significance of scaling the pixel-wise loss with the cosine alignment of predictions rather than using a heuristic, argmax-based scaling as in SegPGD.

Thus, we successfully demonstrate the benefit of CosPGD over existing adversarial attacks for semantic segmentation.
We provide more results on $\ell_{\infty}$-norm and $\ell_2$-norm constrained non-targeted adversarial attacks for semantic segmentation using UNet~\citep{unet} with ConvNeXt backbone on \textbf{CityScapes}~\citep{cordts2016cityscapes} in Appendix~\ref{subsec:exp_results:unet}, further confirming the benefit of CosPGD.

Additionally, we ablate over the attack step size $\alpha$ for $\ell_{\infty}$-norm constrained attacks on DeepLabV3 using PASCAL VOC2012 validation dataset in Appendix~\ref{subsec:exp_results:l_inf} and over multiple attack step size $\alpha$ and permissible perturbation $\epsilon$ for $l_2$-norm constrained attacks on DeepLabV3 using PASCAL VOC2012 validation dataset in Appendix~\ref{subsec:exp_results:deeplabl2}.
We show in Appendix~\ref{subsec:exp_results:l_2} that CosPGD outperforms both PGD and SegPGD (for segmentation) in the $\ell_{2}$-norm constraint settings under all commonly used $\epsilon$ and $\alpha$ values.
%
\noindent\paragraph{Optical Flow.}
In addition to the results discussed in \cref{subsec:exp:optical}, we provide results comparing CosPGD to PGD as a $\ell_{\infty}$-constrained non-targeted attack for optical flow estimation in Appendix~\ref{subsec:appendix:limitations:l_inf_kitti}.
We also provide a comparison to PCFA~\citep{pcfa} in Appendix.~\ref{subsec:appendix:PCFA}.
\noindent\paragraph{Image Deblurring.}
 \begin{figure}[htb]
    \centering 
 \includegraphics[width=\linewidth]{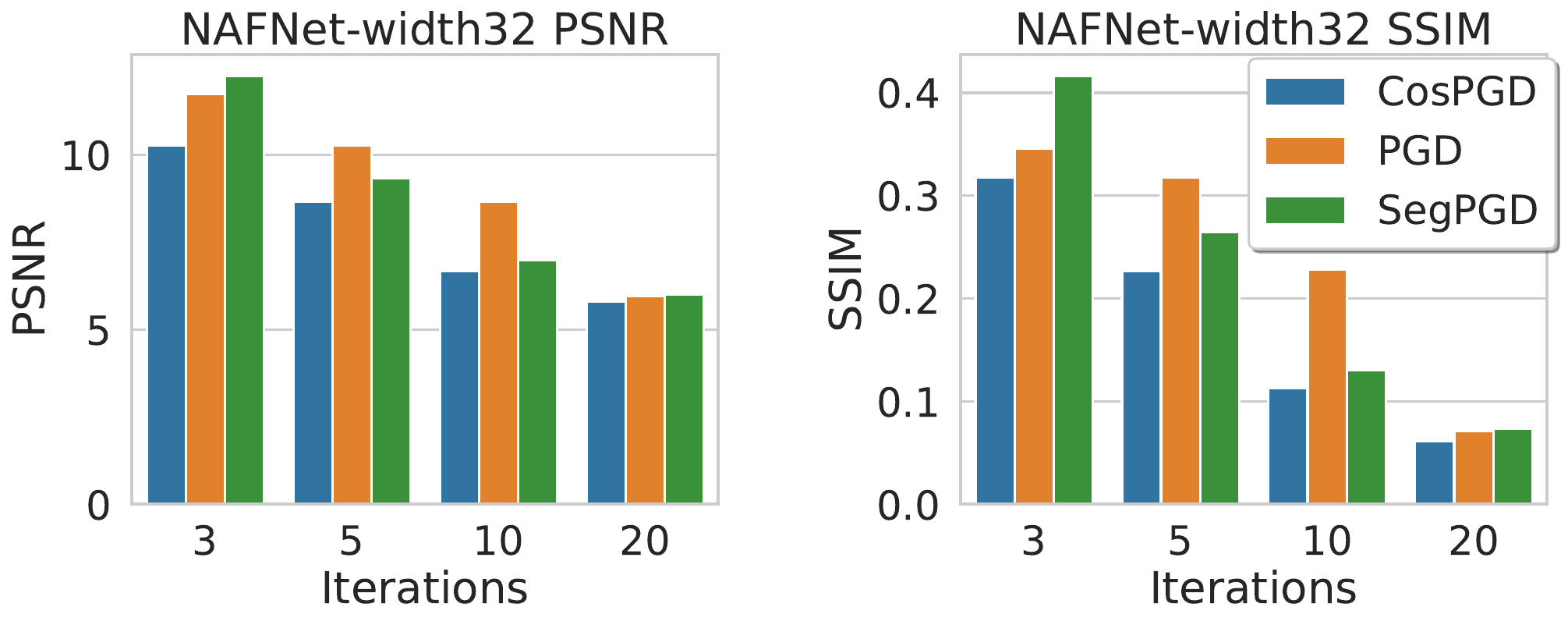}
    \caption{Non-targeted $\ell_{\infty}$-norm constrained  CosPGD, PGD, and SegPGD attacks on NAFNet, recently proposed by \cite{nafnet} as the state-of-the-art network for image de-blurring on the GoPro dataset.CosPGD significantly outperforms the other attacks.
    Lower PSNR and SSIM indicate a worse restoration and thus a stronger attack.
    }
    \label{fig:gopro}
\end{figure}
To demonstrate CosPGD's versatility, we last consider the vision transformer-based image restoration model NAFNet~\citep{nafnet}.
NAFNet outperforms Restormer~\citep{Zamir2021Restormer} for image restoration tasks like image de-blurring and image denoising on clean data, thus implying that NAFNet learns good representations.
 %
\autoref{fig:gopro} depicts results for NAFNet on image deblurring of the GoPro dataset images.
We observe that CosPGD is a significantly stronger attack than both PGD and SegPGD on this task.
We provide further discussion and results on Restormer~\citep{Zamir2021Restormer} and the ``Baseline network" \citep{nafnet} in Appendix~\ref{subsec:appendix:image_deblurring}.
%
%
%
%

%
%
%
%
\section{Conclusion}
\label{sec:conclusion}
In this work, we demonstrated across different downstream tasks and architectures that our proposed adversarial attack, CosPGD, is significantly more effective than other existing and commonly used adversarial attacks on several pixel-wise prediction tasks.
We provide a new algorithm for evaluating the adversarial robustness of models on pixel-wise tasks.
By comparing CosPGD to attacks like PGD, which were originally proposed for image classification tasks, we expanded on the work by \citet{segpgd} and highlighted the need and effectiveness of attacks specifically designed for pixel-wise prediction tasks beyond segmentation.
We illustrated the intuition behind using cosine similarity as a measure for generating stronger adversaries and leveraging more information from the model and backed it with experimental results from different downstream tasks.
This further highlights the simplicity and principled formulation of CosPGD, making it applicable to a wide range of pixel-wise prediction tasks and in principle extendable to all Lipschitz continuous bounds as a targeted as well as a non-targeted attack.


\paragraph{Limitations.} 
Most white-box adversarial attacks require access to ground truth labels ~\citep{fgsm, pgd, pgdl2, apgd, segpgd}.
While this is beneficial for generating adversaries, it limits the applications of the non-targeted attacks like SegPGD as many benchmark datasets~\citep{kitti15, sintel1, sintel2, pascal-voc-2012} do not provide the ground truth for test data. 
The wide-applicability of CosPGD allows it to be used as a targeted attack thus mitigating this limitation to a great extent.
Yet, it would be interesting to study the attack on the ground truth test images in the non-targeted setting as well, due to the potential slight distribution shifts pre-existing in the test data.
We discuss additional limitations of CosPGD in \Cref{sec:appendix:limitations}.

\section*{Acknowledgements}
S.J and M.K acknowledge funding by the DFG Research Unit 5336 - Learning to Sense.
The OMNI cluster of University of Siegen was used for some of the initial computations.

\section*{Impact Statement} We have carefully read the ICML 2024 Code of Ethics and confirm that we adhere to it. 
The proposed work is original and novel.
To the best of our knowledge, all literature used in this work has been referenced correctly.
Our work did not involve any human subjects and does not pose a threat to humans or the environment.

Assessing the quality of representations learned by a machine learning model is of paramount importance.
This makes sure that the model is not learning shortcuts from the input distribution to the target distribution~\cite{shortcut} but learning something meaningful.
Adversarial attacks are a reliable tool for gauging the quality of a model's learned representations. 
However adversarial attacks are time and computation exhaustive.
Thus, our proposed adversarial attack, CosPGD helps in this regard as it can provide new insights into a model's robustness and vulnerabilities with much less time and thus computation and is theoretically motivated. 
Thus, our work helps advance the field of machine learning.

\section*{Author Contribution}
The idea for CosPGD was conceptualized by Shashank Agnihotri and improved by discussions with Steffen Jung and Margret Keuper.
Shashank Agnihotri led the development, with inputs from Steffen Jung and Margret Keuper.
Margret Keuper provided supervision and contributed significantly to the writing.
Steffen Jung additionally made notable and significant contributions with experiments for non-targeted attacks on semantic segmentation, especially experiments with PSPNet, DeepLabV3, and Robust UPerNet.
Shashank Agnihotri performed the remaining experiments.

\bibliography{main_paper}
\bibliographystyle{icml2024}

\newpage
\appendix
\onecolumn
{
    \centering
    \Large
    \textbf{CosPGD: an efficient and unified white-box adversarial attack for pixel-wise prediction tasks} \\
    \vspace{0.5em}Supplementary Material \\
    \vspace{1.0em}
}

We include the following information in the supplementary material:

\begin{itemize}
    \item Section~\ref{sec:appendix} Additional Details:
    \begin{itemize}
        \item{Section~\ref{proof1}: We provide the proof for proposition \ref{prop1}.}
        \item  Section~\ref{appendix:algorithm}: Algorithm of CosPGD.
        \item Section~\ref{sup:details}: Hardware details
        \item Section~\ref{subsec:appendix:code}: Implementation details including code and example usage.
        \item Section~\ref{subsec:exp_details:image_restoration}: We provide additional experimental details for the image deblurring experiments.
        \item Section~\ref{subsec:appendix:time_taken}: We compare the time taken by different adversarial attacks for different tasks.
        \item Section~\ref{subsec:appendix:epe_f1_all}: Details on calculating \emph{epe-f1-all}.
    \end{itemize}
    \item Section~\ref{subsec:appendix:semseg}: Semantic Segmentation Additional Results:
    \begin{itemize}
        \item Section~\ref{subsec:appendix:semseg:segformer}: We provide additional experimental results using SegFormer~\cite{xie2021segformer} on ADE20K~\cite{ade20k_cvpr,ade20k_journal}.
        \begin{itemize}
            \item Section~\ref{subsubsec:appendix:semseg:segformer:epsilon_ablation}: We report an ablation study over multiple $\epsilon$ values for $\ell_{\infty}$-norm bounded attacks
        \end{itemize}
        \item Section~\ref{subsec:appendix:semseg:transfer_attacks}: We provide evaluations on transferring adversarial attacks between a DeepLabV3 and a PSPNet model on PASCALVOC2012 dataset.
        \item Section~\ref{subsec:appendix:semseg:defense_methods}:  We report the performance of adversarial attacks against some SotA defense methods.
        \item Section~\ref{subsec:appendix:semseg:attacking_sam}: Here we report transfer attacks from a DeepLabV3 to Segment Anything Model (SAM)~\cite{segment_anything}.
        \item Section~\ref{subsec:exp_results:unet}: We provide extra $l_{\infty}$-norm and $l_2$-norm constrained non-targeted adversarial attack results from Semantic Segmentation using the UNet architecture with ConvNeXt backbone on the \textbf{CityScapes} dataset~\citep{cordts2016cityscapes}.
        \item Section~\ref{subsec:exp_results:deeplabl2}: We provide an ablation study on attack step size $\alpha$ and $\epsilon$ for $l_2$-norm bounded for non-targeted adversarial attack results from Semantic Segmentation using DeepLabV3 on the PASCAL VOC 2012 dataset.
        \item Section~\ref{subsec:exp_results:l_inf}: We provide an ablation study on attack step size $\alpha$ for $l_{\infty}$-norm bounded for non-targeted adversarial attack results from Semantic Segmentation using DeepLabV3 on the PASCAL VOC 2012 dataset.
        \item Section~\ref{subsec:appendix:results:tables}: We report results from Figure~\ref{fig:semseg:semseg_pgd} in a tabular form.
        \item \textcolor{black}{Section~\ref{subsec:appendix:adversarial_training}: We report the results of adversarial training for semantic segmentation.}
    \end{itemize}
    \item Section~\ref{subsec:appendix:optical}: Optical Flow Additional Results:
    \begin{itemize}
        \item Section~\ref{subsec:appendix:optical_tabular}: We report results from Figure~\ref{fig:optical:target:l_infinity} in a tabular form.
        \item Section~\ref{subsec:appendix:limitations:l_inf_kitti}: We provide extra results comparing CosPGD to PGD as a $l_{\infty}$-norm constrained non-targeted adversarial attack for optical flow estimation.
    \item Section~\ref{subsec:appendix:PCFA}: We provide a comparison to the $l_2$-constrained PCFA~\citep{pcfa}, which is a dedicated attack for optical flow.
    \end{itemize}
    \item Section~\ref{subsec:appendix:image_restoration}: Image Restoration Results:
    \begin{itemize}
        \item Section~\ref{subsec:appendix:image_deblurring}: We report the findings on the adversarial robustness of many recently proposed transformer-based image deblurring models.
        \item Section~\ref{subsec:appendix:limitations:l_inf_ssid}: We report the results on many recently proposed transformer-based image denoising models.
    \end{itemize}
    \item Section~\ref{sec:appendix:limitations}: A detailed discussion on limitations of CosPGD
\end{itemize}

In Table~\ref{tab:all_experiments}, we provide a look-up table for all experiments considered in this supplementary material. 
We provide details on the downstream tasks, models, targeted and non-targeted attack settings, and $l_{\infty}$-norm constrained and $l_{2}$-norm constrained settings considered respectively do demonstrate the wide-applicability of CosPGD.


\section{Appendix}
\label{sec:appendix}
\begin{table*}[ht]
    \centering
    \caption{Look-up table for considered experiments in this appendix.}
    \scalebox{0.5}{
    \begin{tabular}{@{}lccc|cc|cc}
    \toprule
    \multirow{2}{*}{Downstream Task} & \multirow{2}{*}{Networks} & \multirow{2}{*}{Dataset} & \multirow{2}{*}{Study} & \multicolumn{2}{c}{Non-targeted Attack} & \multicolumn{2}{c}{Targeted Attack} \\

    & & & & $l_{\infty}$-norm constraint & $l_{2}$-norm constraint & $l_{\infty}$-norm constraint & $l_{2}$-norm constraint \\

    \midrule
    
    \multirow{8}{*}{Semantic Segmentation} & DeepLabV3 & \multirow{3}{*}{PASCAL VOC 2012, Cityscapes} & various $\epsilon$ and $\alpha$ values & \multirow{3}{*}{Sec.~\ref{subsec:exp_results:l_inf}} & Sec.~\ref{subsec:exp_results:l_2} \\
    & PSPNet & &  Non-targeted Attacks   & \\
    & UNet & &   Non-targeted Attacks   & & & \\
    & SegFormer & ADE20K &   various $\epsilon$ values   & Sec.~\ref{subsubsec:appendix:semseg:segformer:epsilon_ablation} & & \\
    & Robust UPerNet \cite{croce2023robust} & PASCAL VOC 2012 &   Performance against Defense Methods     & Sec.~\ref{subsec:appendix:semseg:defense_methods} & \\
    & Robust PSPNet \cite{9710783} & PASCAL VOC 2012 &   Performance against Robust Models  & Sec.~\ref{subsec:appendix:semseg:defense_methods} & \\
    & DeepLabV3 $\rightarrow$ SAM & PASCAL VOC 2012 &   Transfer Attack on SAM    & Sec.~\ref{subsec:appendix:semseg:attacking_sam} & \\
    & DeepLabV3 $\rightarrow$ PSPNet & PASCAL VOC 2012 &   Transfer Attacks    & Sec.~\ref{subsec:appendix:semseg:transfer_attacks} & \\
    & PSPNet $\rightarrow$ DeepLabV3 & PASCAL VOC 2012 &  Transfer Attacks    & Sec.~\ref{subsec:appendix:semseg:transfer_attacks} & \\

    \midrule

    \multirow{2}{*}{Optical Flow Estimation} & RAFT & \multirow{2}{*}{KITTI 2015, Sintel (clean and final)} &   Targeted Attacks  & Sec.~\ref{subsec:appendix:limitations:l_inf_kitti} & &  Sec.~\ref{subsec:appendix:optical} & \multirow{2}{*}{Sec.~\ref{subsec:appendix:PCFA}} \\

    & PWCNet, GMA, SpyNet &  &     Comparison to PCFA   &            \\
    \midrule

    Image Deblurring & Restormer, Baseline net, NAFNet & GoPro &       Non-targeted Attacks       & Sec.~\ref{subsec:appendix:image_deblurring} & \\

    Image Denoising & Baseline net, NAFNet & SSID &    Non-targeted Attacks       & Sec.~\ref{subsec:appendix:limitations:l_inf_ssid} & \\

    \bottomrule

    \end{tabular}
    }
    
    \label{tab:all_experiments}
\end{table*}

\subsection{Proof of Proposition \ref{prop1}}\label{proof1}

    We are to show that, for any two pixel-wise network predictions $f_\theta(\boldsymbol{X})_i$ and $f_\theta(\boldsymbol{\bar{X}})_i \in\mathbb{R}^{M}$, a target $\boldsymbol{Y}_i\in\mathbb{R}^{M}$ and a continuously differentiable function $\psi:\mathbb{R}^{M}\rightarrow  \mathbb{R}^{M}$ with  $ \| \psi(f_{\theta}(\boldsymbol{X})) \| =1 \quad\forall f_{\theta}(\boldsymbol{X})$, there exists a real, constant $d\geq 0$ so that
  %
    \begin{align*}
    & d\cdot\|f_\theta(\boldsymbol{X})_i - f_\theta(\boldsymbol{\bar{X}})_i\|\geq\qquad\\
    &\qquad \|\mathrm{cos}\left(\psi(f_\theta(\boldsymbol{X})_i), \boldsymbol{Y}_i\right) -  \mathrm{cos}\left(\psi(f_\theta(\boldsymbol{\bar{X}})_i), \boldsymbol{Y}_i\right)\|.
    \end{align*}
\begin{proof}
The function $\psi:\mathbb{R}^{M}\rightarrow  \mathbb{R}^{M}$ as well as the cosine similarity $\mathrm{cos}:\mathbb{R}^{M}\times \mathbb{R}^{M} \rightarrow [-1,1]$ are both continuously differentiable functions. From the continuous differentiability of $\psi$, it follows that is it Lipschitz continuous, i.e.~there exists a real constant $d_1\geq 0$ so that
 \begin{align*}
     d_1\cdot\|f_\theta(\boldsymbol{X})_i - f_\theta(\boldsymbol{\bar{X}})_i\|\geq
     \|\psi(f_\theta(\boldsymbol{X})_i) - \psi(f_\theta(\boldsymbol{\bar{X}})_i)\|
    \end{align*}
    for any $f_\theta(\boldsymbol{X})_i$ and $f_\theta(\boldsymbol{\bar{X}})_i \in\mathbb{R}^{M}$.
    Further, the cosine similarity effectively computes the norm of the projection of the normalized model predictions onto the target vector, which is again a continuously differentiable operation, i.e.~is again Lipschitz continuous
 \begin{align*}
    & d_2\cdot\|\psi(f_\theta(\boldsymbol{X})_i) - \psi(f_\theta(\boldsymbol{\bar{X}})_i)\|\geq\qquad\\
    &\qquad \|\mathrm{cos}\left(\psi(f_\theta(\boldsymbol{X})_i), \boldsymbol{Y}_i\right) -  \mathrm{cos}\left(\psi(f_\theta(\boldsymbol{\bar{X}})_i), \boldsymbol{Y}_i\right)\|.
    \end{align*}
    for a real constant $d_2\geq 0$.
\end{proof}

\subsection{Algorithm for CosPGD}
\label{appendix:algorithm}
Following we present the algorithm for CosPGD.
Algorithm~\ref{alg:cospgd} provides a general overview of the implementation of CosPGD.
It demonstrates that CosPGD is downstream-task agnostic, $l_p$-norm agnostic, and agnostic to targeted or non-targeted application.
\begin{algorithm*}[tb]
\caption{Algorithm for generating adversarial examples using CosPGD.}
\label{alg:cospgd}
\begin{algorithmic}
\footnotesize
\Require model $f_{\mathrm{net}}(\cdot)$, clean samples $\boldsymbol{X}^{\mathrm{clean}}$, perturbation range $\epsilon$, step size $\alpha$, attack iterations $T$, ground truth/target $\boldsymbol{Y}$
\vspace{0.1cm}
\State $\boldsymbol{X}^{\mathrm{adv}_0} = \boldsymbol{X}^{\mathrm{clean}}+ \mathcal{U}(-\epsilon, +\epsilon)$ \Comment{initialize adversarial example and clip to valid $\ell_{\infty}$ or $l_{2}$ bound}
\For{t ← 0 to T-1} \Comment{loop over attack iterations}
    \vspace{0.15cm}
    \State $P = f_{\mathrm{net}}(\boldsymbol{X}^{\mathrm{adv}_{t}})$ \Comment{make predictions}
    \vspace{0.15cm}
    \State $\mathrm{cossim} \gets CosineSimilarity(\psi(P), \boldsymbol{Y'})$ \Comment{compute cosine similarity}
    \vspace{0.05cm}
    \State if targeted attack: 
    \vspace{0.05cm}
    \State \hspace{1.05cm} $\mathrm{cossim} \gets 1 - \mathrm{cossim}$ \Comment{punish dissimilarity to target}
    \State \hspace{1.05cm} $\alpha \gets -\alpha$ \Comment{opposite direction for targeted attack}
    \vspace{0.05cm}
    \State $L_{\mathrm{cos}} \gets \mathrm{cossim} \cdot \Bar{L}(P, \boldsymbol{Y})$ \Comment{scaling the pixel-wise loss for sample updates}
    \vspace{0.15cm}
    \State $\boldsymbol{X}^{\mathrm{adv}_{t+1}} \gets \boldsymbol{X}^{\mathrm{adv}_{t}} + \alpha \cdot \textit{sign}(\nabla_{\boldsymbol{X}^{\mathrm{adv}_{t}}} L_\mathrm{cos})$ \Comment{update adversarial examples}
    \State $\delta \gets \phi^{\epsilon}(\boldsymbol{X}^{\mathrm{adv}_{t+1}} - \boldsymbol{X}^{\mathrm{clean}})$ \Comment{clip $\delta$ to valid $\ell_{\infty}$ or $l_2$ bound}
    \State $\boldsymbol{X}^{\mathrm{adv}_{t+1}} = \phi^{\epsilon}(\boldsymbol{X}^{\mathrm{clean}} + \delta)$  \Comment{add $\delta$ to $\boldsymbol{X}^{\mathrm{clean}}$ and clip into valid image range}
\EndFor
\State $P = f_{\mathrm{net}}(\boldsymbol{X}^{\mathrm{adv}_{T}})$ \Comment{make predictions on adversarial examples}
\end{algorithmic}
\end{algorithm*}

\subsection{Further Experimental Details on Hardware and Metrics}\label{sup:details}

\paragraph{Semantic Segmentation}
We use PASCAL VOC 2012~\citep{pascal-voc-2012}, which contains 20 object classes and one background class, with 1464 training images, and 1449 validation images. 
We follow common practice \citep{contouring, segpgd, semseg2019, semsegzhao2017pspnet}, and use work by~\citet{SBD_BharathICCV2011}, augmenting the training set to 10,582 images. 
We evaluate on the validation set. 
Architectures used for our evaluations are PSPNet~\citep{semsegzhao2017pspnet} and DeepLabV3~\citep{deeplabv3}, both with ResNet50~\citep{resnet} encoders, and UNet~\citep{unet} with a ConvNeXt tiny encoder~\citep{convnext}.
Results are reported in Appendix~\ref{subsec:exp_results:unet}. 
We report mean Intersection over Union~(mIoU) and mean pixel accuracy~(mAcc).

\textbf{Hardware. } For the experiments on DeepLabV3, we used NVIDIA Quadro RTX 8000 GPUs. For PSPNet, we used NVIDIA A100 GPUs. For the experiments with UNet, we used NVIDIA GeForce RTX 3090 GPUs.

\paragraph{Optical Flow}
We use RAFT~\citep{raft} and follow the evaluation procedure used therein. 
Evaluations are performed on KITTI2015~\citep{kitti15} and MPI Sintel~\citep{sintel1, sintel2} validation sets. 
We use the networks pre-trained on FlyingChairs~\citep{flyingchairs} and FlyingThings~\citep{flyingthings_MIFDB16} and fine-tuned on training datasets of the specific evaluation, as provided by \citet{raft}. 
For Sintel we report the end-point error~($epe$) on both clean and final subsets, while for KITTI15 we report the $epe$ and $epe$-$f1$-$all$.
In Appendix~\ref{subsec:appendix:PCFA} we compare CosPGD to PCFA across different networks.

\textbf{Hardware. } We used NVIDIA V100 GPUs, a single GPU was used for each run.

\paragraph{Image Restoration}
Following the regime of \cite{nafnet,Zamir2021Restormer,agnihotri2023unreasonable},
for the image de-blurring task we use the GoPro dataset~\citep{gopro} as in \cite{nafnet}. 
The images are split into 2103 training images and 1111 test images.
We consider the ``Baseline network" and NAFNet as proposed by \cite{nafnet}.
For the image restoration tasks we report the $PSNR$ and $SSIM$ scores of the reconstructed images w.r.t.~to the ground truth images, averaged over all images.
We provide further details in Appendix~\ref{subsec:appendix:image_deblurring}.

\textbf{Hardware. } For the experiments on Image de-blurring tasks, we used NVIDIA GeForce RTX 3090 GPUs. 
A single GPU was used for each run.

\subsubsection{\revision{Code for the attack}}
\label{subsec:appendix:code}
\revision{The code for the functions used for generating adversarial samples using CosPGD and other considered adversarial attacks in the main paper is available at \url{https://github.com/shashankskagnihotri/cospgd}.}



\revision{Additionally, we provide sample code demonstrating the usage of the packages for a UNet-like architecture with detailed instructions at \url{https://github.com/shashankskagnihotri/cospgd}.
}

\subsubsection{Calculating epe-f1-all}
\label{subsec:appendix:epe_f1_all}
Following the work by \citet{raft}, $f1-all$ is calculated by averaging $out$ over all the predicted optical flows. $out$ is calculated using Equation~\eqref{eqn:out},
\begin{equation}
\label{eqn:out}
    out = epe > 3.0 \cup \dfrac{epe}{mag} > 0.05
\end{equation}
Where, $mag$ = $\sqrt{flow~ground~truth^2}$
\revision{and $epe$ is the Euclidean distance between the two vectors.}

\subsubsection{\revision{Image Deblurring Experimental Details}}
\label{subsec:exp_details:image_restoration}
\revision{\citet{nafnet} simplify a transformer-based architecture Restormer~\citep{Zamir2021Restormer} for image restoration tasks and first propose a simplified architecture as a Baseline network, and then improve upon it with intuitions backed by reasoning and ablation studies to propose Non-linear Activation Free Networks abbreviated as NAFNet.
In this work, we perform adversarial attacks on both the Baseline network and NAFNet.}

\paragraph{\revision{Dataset.} }\revision{Similar to \cite{nafnet}, for the image de-blurring task, we use the GoPro dataset~\citep{gopro} which consists of 3124 realistically blurry images of resolution 1280$\times$720 and corresponding ground truth sharp images obtained using a high-speed camera.
The images are split into 2103 training images and 1111 test images.
For the image denoising task, we use the Smartphone Image Denoising Dataset (SSID)~\citep{ssid}.
This dataset consists of 160 noisy images taken from 5 different smartphones and their corresponding high-quality ground truth images.}

\paragraph{\revision{Metrics.} }\revision{For both the image restoration tasks, we report the $PSNR$ and $SSIM$ scores of the reconstructed images w.r.t. to the ground truth images, averaged over all images.
$PSNR$ stands for Peak Signal-to-Noise ratio, a higher $PSNR$ indicates a better quality image or an image closer to the image to which it is being compared.
$SSIM$ stands for Structural similarity~\citep{ssim}.}

\subsubsection{\revision{Comparing Time taken by different Adversarial Attacks}}
\label{subsec:appendix:time_taken}
\revision{Following, we report the approximate time taken by each attack in minutes.
Please note, this time includes time taken for data-loading and saving of experimental results including images.
For a given task, network, and dataset, the time taken by different attacks is comparable and representative of the time taken by the attacks as they followed the same attack procedures.
We observe in \autoref{tbl:exp:time} that the difference in time taken by the different attacks at the same number of iterations is negligible.
This is because operations like one-hot encoding and softmax take negligible time.}

\revision{Thus, the ability of CosPGD to provide valuable insights into model robustness with significantly less iterations than other methods, as discussed in Section~\ref{subsec:exp:semseg} and Section~\ref{subsec:exp:gopro} is a compelling advantage.}
\begin{table*}[htb]
\caption{
Comparison of time taken in minutes by different attacks on different downstream tasks for different amount of iterations.
The computation times are comparable.
}
\label{tbl:exp:time}
\centering
\scalebox{.7}{
\begin{tabular}{@{}p{2.5cm}@{\hspace{0.1cm}}c@{\hspace{0.1cm}}c@{\hspace{0.1cm}}c@{\hspace{0.1cm}}c@{\hspace{0.1cm}}|@{\hspace{0.1cm}}c@{\hspace{0.1cm}}|@{\hspace{0.1cm}}c@{\hspace{0.1cm}}|@{\hspace{0.1cm}}c@{\hspace{0.1cm}}|@{\hspace{0.1cm}}c@{}}
\toprule
 \multirow{3}{2.5cm}{\textbf{Task}} & \multirow{3}{1.5cm}{\textbf{Network}} & \multirow{3}{3cm}{\textbf{Dataset}} & \multirow{3}{3cm}{\textbf{Attack method}} &  \multicolumn{5}{c}{\textbf{Attack iterations}} \\
 & & & & 3 & 5 & 10  &  20 &  40   \\
 & & &  & Time (mins) &  Time (mins) & Time (mins) & Time (mins) & Time (mins) \\
\midrule
 \multirow{2}{2.3cm}{\textbf{Semantic Segmenation}} & \multirow{2}{1.5cm}{\textbf{UNet}} & \multirow{2}{2.5cm}{\textbf{PASCAL VOC 2012}} & \textbf{SegPGD} & 28.73 &	36.33 & 58.72 & 88.93 & 163.15 \\
  & & & \textbf{CosPGD} &  26.67 & 36.75 & 54.45 & 97.08 & 165.35 \\
\midrule
\multirow{4}{2.3cm}{\textbf{Optical Flow}} & \multirow{4}{1cm}{\textbf{RAFT}} & \multirow{2}{2.5cm}{\textbf{KITTI2012}} & \textbf{PGD} & 5.90 & 7.73 & 12.23 & 20.98 & 37.45 \\
  & & & \textbf{CosPGD} &  6.00 & 7.85 & 12.15 & 21.03 & 38.28 \\
\cmidrule{3-9}
  & & \multirow{2}{2.5cm}{\textbf{Sintel~(clean + final)}} & \textbf{PGD} & 69.87 & 97.47 & 158.28 & 297.40 & 557.97 \\
   & & & \textbf{CosPGD} & 73.68 & 102.77 & 160.40 & 287.82 & 602.08 \\
\bottomrule

\end{tabular}
}
\end{table*}

\section{Semantic Segmentation}
\label{subsec:appendix:semseg}
Following we provide additional Semantic Segmentaion evaluations, including study on different $\epsilon$ values, different $\alpha$ values, using different tasks and transfer attacks on SAM using a DeepLabV3.

\subsection{Semantic Segmentation with SegFormer on ADE20k}
\label{subsec:appendix:semseg:segformer}

\subsubsection{Implementation Details}
For experiments with SegFormer~\cite{xie2021segformer} with MIT-B0 backbone, we use the ADE20k dataset~\cite{ade20k_journal}.
This dataset has 150 classes and is split into 25,574 training images and 2,000 validation images.

We perform $\ell_{\infty}$-bounded  PGD, SegPGD and CosPGD with various $\epsilon$ values $\in$ \{$\frac{2}{255}$,$\frac{4}{255}$,$\frac{6}{255}$,$\frac{8}{255}$,$\frac{10}{255}$,$\frac{12}{255}$\}, over various attack iterations $\in$ \{3, 5, 10, 20, 40, 100\}.

\subsubsection{Ablation over multiple $\epsilon$ values for $\ell_{\infty}$-norm bounded attacks}
\label{subsubsec:appendix:semseg:segformer:epsilon_ablation}
\begin{table}[htb]
    \centering
    \caption{Attacking SegFormer with a MIT-B0 backbone using ADE20K with different $\ell_{\infty}$ bounded $\epsilon$ values and with different adversarial attacks.}
    \scalebox{0.63}{
    \begin{tabular}{cc|cc|cc|cc|cc|cc|cc}
         \toprule
         \multirow{3}{*}{Attack Method} & \multirow{3}{*}{$\frac{\epsilon}{255}$ value} & \multicolumn{12}{c}{Attack Iterations} \\
         & & \multicolumn{2}{c}{3} & \multicolumn{2}{c}{5} & \multicolumn{2}{c}{10} & \multicolumn{2}{c}{20} & \multicolumn{2}{c}{40} & \multicolumn{2}{c}{100} \\
         & & mIoU (\%) & mAcc (\%) & mIoU (\%) & mAcc (\%) & mIoU (\%) & mAcc (\%) & mIoU (\%) & mAcc (\%) & mIoU (\%) & mAcc (\%) & mIoU (\%) & mAcc (\%) \\

         \midrule
         
         PGD & \multirow{3}{*}{2} & 8.45 & 14.44 &	6.62 & 11.49	& 5.36 & 9.45 &	4.21 & 7.51	 &	3.8 & 6.73 &		3.3 & 6.12 \\
         SegPGD & & 5.80 & 10.15 &		4.88 & 8.68 &		3.69 & 6.56 &		2.91 & 5.18	 &	2.41 & 4.49 &		2.19 & 4.02 \\
         \textbf{CosPGD} & & \textbf{5.37} & \textbf{10.06} &		\textbf{3.75} &\textbf{ 7.26} &		\textbf{2.18} & \textbf{4.3} &		\textbf{1.87} & \textbf{3.55} &		\textbf{1.68} & \textbf{3.01} &		\textbf{1.37} & \textbf{2.46} \\
        
        \midrule
         PGD & \multirow{3}{*}{4} &   5.11 & 9.48	 &   2.94 & 5.63 &   	1.66 & 3.34	& 1.01 & 2.21 &   	0.79 & 1.79 &   	0.6 & 1.38    \\
         SegPGD & &  3.29 & 6.15 &   	1.83 & 3.7 &   	0.89 & 1.9 &   	0.47 & 1.18	 &   0.3 & 0.86	 &   0.26 & 0.68    \\
         \textbf{CosPGD} & &   	\textbf{1.66} & \textbf{3.45}	 &	\textbf{0.55} & \textbf{1.28} &		\textbf{0.09} & \textbf{0.22}	 &	\textbf{0.05} & \textbf{0.09} &		\textbf{0.05} & \textbf{0.09} &		\textbf{0.04} & \textbf{0.06}   \\

         \midrule
         PGD & \multirow{3}{*}{6} &   	3.97 & 7.5	 &   2.05 & 4.1	 &   1.07 & 2.28 &   	0.67 & 1.57 &   	0.41 & 1.14	& 0.36 & 0.88    \\
         SegPGD & &  2.64 & 5.10	 &   1.22 & 2.71 &   	0.47 & 1.24	 &   0.21 & 0.7 &   	0.13 & 0.49	 &   0.09 & 0.35    \\
         \textbf{CosPGD} & &   \textbf{1.11} & \textbf{2.39} &		\textbf{0.18} & \textbf{0.52} &		\textbf{0.01} & \textbf{0.04} &		\textbf{0.0} & \textbf{0.01} &		\textbf{0.0} & \textbf{0.0}	 &	\textbf{0.0} & \textbf{0.0}   \\

         \midrule
         PGD & \multirow{3}{*}{8} &   3.38 & 6.48	 &   1.76 & 3.63 &   	0.82 & 1.95	 &   0.46 & 1.28 &   	0.37 & 1.04	 &   0.2 & 0.7    \\
         SegPGD & &  2.31 & 4.54	 &   0.90 & 2.06	 &   0.33 & 1.03 &   	0.15 & 0.61 &   	0.09 & 0.35	 &   0.05 & 0.28    \\
         \textbf{CosPGD} & &  \textbf{0.98} &\textbf{2.21} &		\textbf{0.08} & \textbf{0.25} &		\textbf{0.00} & \textbf{0.02} &		\textbf{0.00} & \textbf{0.00}	 &	\textbf{0.00} & \textbf{0.00} &		\textbf{0.00} & \textbf{0.00}    \\

         \midrule
         PGD & \multirow{3}{*}{10} &    3.29 & 6.28	 &   1.74 & 3.58 &   	0.79 & 1.99 &   	0.47 & 1.27 &   	0.34 & 1.01	 &   0.24 & 0.74   \\
         SegPGD & &  1.91 & 3.88 &   	0.89 & 2.09 &   	0.32 & 0.96 &   	0.18 & 0.65	 &   0.08 & 0.38 &   	0.05 & 0.27    \\
         \textbf{CosPGD} & &  \textbf{0.81} & \textbf{1.82}	 &	\textbf{0.11} & \textbf{0.41} &		\textbf{0.00} & \textbf{0.01} &		\textbf{0.00} & \textbf{0.00} &		\textbf{0.00} & \textbf{0.00} &		\textbf{0.00} & \textbf{0.00}    \\

         \midrule
         PGD & \multirow{3}{*}{12} & 3.16 & 5.95 &   	1.49 & 2.98	 &   0.72 & 1.79 &   	0.45 & 1.27	 &   0.31 & 0.93	 &   0.24 & 0.69      \\
         SegPGD & &   1.83 & 3.77	 &   1.83 & 3.77	 &   0.26 & 0.83 &   	0.14 & 0.6	 &   0.1 & 0.44 &   	0.04 & 0.26   \\
         \textbf{CosPGD} & &   \textbf{0.72} & \textbf{1.68} &		\textbf{0.08} & \textbf{0.22} &		\textbf{0.00} & \textbf{0.00}	 &	\textbf{0.00} & \textbf{0.00} &		\textbf{0.00} & \textbf{0.00}	 &	\textbf{0.00} & \textbf{0.00}   \\

         \bottomrule
    \end{tabular}    
    }
    \label{tab:segformer_epsilon_ablation}
\end{table}
Since ADE20K has 150 classes, making it a more difficult distribution to learn, it is not usually considered to evaluate attack methods. 
We expect CosPGD to be a significantly stronger attack than SegPGD or the simple PGD on this data because it can smoothly align the loss to the posterior distribution.
In \Cref{tab:segformer_epsilon_ablation} we confirm this by providing additional experiments using SegFormer with $\ell_{\infty}$-norm bounded $\epsilon=\frac{8}{255}$ attacks with $\alpha$=0.01 for Untargeted Attacks. 
Note that the chosen attack settings are the default values proposed in SegPGD.

We observe that CosPGD is a significantly stronger attack than SegPGD for ADE20K and SegFormer.
Please also note that white-box attacks are extremely useful in exposing a model’s vulnerabilities, however, they are very expensive to run, and thus 40 or more attack iterations are generally considered to be a very high number of attack iterations in white-box attack literature (please refer to PGD, APGD, PCFA, SegPGD, AutoAttack, MI-FGSM).
Here, CosPGD required merely 10 attack iterations to bring the model mIoU to absolute $0.00$, whereas SegPGD is not able to achieve this even when using 100 iterations (increasing the attack cost by a factor of 10).
Our current understanding is that given a reasonable perturbation attack, and step size smaller than this budget (so that the perturbations are not clipped away by the budget), all attacks should optimize the adversary in the best possible way. We have shown that CosPGD is better at this optimization than the other white-box attacks for various step-sizes($\alpha$) and various $\epsilon$ values.

For $\ell_{\infty}$-norm we have shown this for $\epsilon=\frac{8}{255}$. 
The maximum permissible perturbation budget should not affect the relative performance of different attacks. 
We further solidify this claim here by providing additional experiments using SegFormer on ADE20K with $\ell_{\infty}$-norm bounded $\epsilon=\{\frac{2}{255}, \frac{4}{255}, \frac{6}{255}, \frac{8}{255}, \frac{10}{255}, \frac{12}{255}\}$ attack settings with $\alpha$=0.01 for Untargeted Attacks in \Cref{tab:segformer_epsilon_ablation}.

\subsection{Evaluating Transfer Attacks}
\label{subsec:appendix:semseg:transfer_attacks}
\begin{table}[htb]
    \centering
    \caption{Transfer Attacks on DeepLabV3 and PSPNet using 20 iterations attacks with $\ell_{\infty}$-norm bounded $\epsilon=\frac{8}{255}$ and $\alpha$=0.01 using PASCAL VOC 2012 validation dataset.}
    \scalebox{1.0}{
    \begin{tabular}{ccc|cc}
         \toprule
         Attacked Model & Attacking Model & Attack Method & mIoU (\%) & mAcc (\%) \\
         \midrule
         \multirow{2}{*}{DeepLabV3 ResNet50} &	\multirow{3}{*}{PSPNet ResNet50}	& \textbf{CosPGD}	& \textbf{1.67} & \textbf{3.59} \\
         & & SegPGD &	1.93 & 5.72 \\
         (Clean mIoU: 76.17) & & PGD &	5.11 & 12.75 \\
         \midrule
         \multirow{2}{*}{PSPNet ResNet50} &	\multirow{3}{*}{DeepLabV3 ResNet50}	& \textbf{CosPGD}	& \textbf{1.21} & \textbf{3.33} \\
         & & SegPGD &	1.77 & 5.62 \\
         (Clean mIoU: 76.78) & & PGD & 4.58 & 12.07 \\
         \bottomrule
    \end{tabular}  
    }
    \label{tab:transfer_attack_itrs_20}
\end{table}

CosPGD, like PGD, SegPGD, and FGSM, is a white box attack. They are designed to optimize attacks for a specific model and generalizability of the attacks to other models i.e. using them in a black-box setting is not a requirement for them at least not something they are optimized to do.
However, it could be interesting to see if the adversarial examples that are optimized on a particular network, also cause a failure in the other.
Thus in \Cref{tab:transfer_attack_itrs_20}, we report results for the PASCAL VOC 2012 dataset when attacking PSPNet using DeepLabV3, and vice versa, both with a ResNet50 encoder. 
We observe that CosPGD is a significantly better attack even in this black-box setting. 
Here we consider $\ell_{\infty}$-norm bounded $\epsilon=\frac{8}{255}$ attacks with $\alpha$=0.01.
\begin{table}[htb]
    \centering
    \caption{Transfer Attacks from DeepLabV3 on PSPNet over various iterations with $\ell_{\infty}$-norm bounded $\epsilon=\frac{8}{255}$ and $\alpha$=0.01 using PASCAL VOC 2012 validation dataset.}
    \scalebox{0.67}{
    \begin{tabular}{ccc|cc|cc|cc|cc}
         \toprule
            \multirow{3}{*}{Attacked Model} & \multirow{3}{*}{Attacking Model} & \multirow{3}{*}{Attack Method} & \multicolumn{8}{c}{Attack Iterations} \\
            
            & & & \multicolumn{2}{c}{3} & \multicolumn{2}{c}{10} & \multicolumn{2}{c}{20} & \multicolumn{2}{c}{40} \\
            & & & mIoU (\&) & mAcc (\%) & mIoU (\&) & mAcc (\%) & mIoU (\%) & mAcc (\%) & mIoU (\&) & mAcc (\%) \\
            \midrule
            \multirow{2}{*}{PSPNet ResNet50} & \multirow{3}{*}{DeepLabV3 ResNet50} & \textbf{CosPGD} & \textbf{9.66} & \textbf{19.39}	& \textbf{2.39} & \textbf{5.91} &	\textbf{1.21} & \textbf{3.33} &	\textbf{1.00} & \textbf{2.59} \\
            & & SegPGD & 9.92 & 19.79 &	2.40 & 6.67	 & 1.77 & 5.62 &	1.23 & 4.40 \\
            (Clean mIoU: 76.78) & & PGD & 14.67 & 27.79	& 5.56 & 13.60	& 4.58 & 12.07 & 	4.35 & 11.81 \\
         \bottomrule
    \end{tabular}
    }
    \label{tab:transfer_attack_attacking_pspnet_multi_itrs}
\end{table}
The benefit of CosPGD over previous methods becomes more significant as the number of attack iterations increases, but is measurable across attack iterations.
We show this in \Cref{tab:transfer_attack_attacking_pspnet_multi_itrs}.

\subsection{Evaluating against Defense Methods}
\label{subsec:appendix:semseg:defense_methods}

\begin{table}[htb]
    \centering
    \caption{Comparing the ``Robust'' PSPNet from \cite{9710783} against white-box adversarial attacks over different number of iterations. Here, same as \cite{9710783}, $\epsilon=\frac{8}{255}$ and $\alpha$=0.01. We use the model weights provided by \cite{9710783} in their \href{https://github.com/dvlab-research/Robust-Semantic-Segmentation}{official GitHub repository}. }
    \scalebox{0.67}{
    \begin{tabular}{cccc|cc|cc|cc|cc}
         \toprule
        \multirow{3}{*}{Training Method} & \multicolumn{2}{c}{Clean Performance} & \multirow{3}{*}{Attack Method} & \multicolumn{8}{c}{Attack Iterations}  \\

        & & & & \multicolumn{2}{c}{2} & \multicolumn{2}{c}{4} & \multicolumn{2}{c}{6} & \multicolumn{2}{c}{10} \\

        & mIoU (\%) & mAcc (\%) & & mIoU (\%) & mAcc (\%) & mIoU (\%) & mAcc (\%) & mIoU (\%) & mAcc (\%) & mIoU (\%) & mAcc (\%) \\

        \midrule

        \multirow{3}{*}{No Defense} &  \multirow{3}{*}{76.90} &  \multirow{3}{*}{84.60} & \textbf{CosPGD} & \textbf{9.11} & \textbf{20.77}	& \textbf{1.56} & \textbf{5.02}	& \textbf{0.54} & \textbf{2.03} & 	\textbf{0.13} & \textbf{0.40} \\
        & & & SegPGD & 10.39 & 22.14 &	3.86 & 9.69	& 2.62 & 6.97 &	1.88 & 5.36 \\

        & & & BIM & 18.90 & 34.92 &	7.59 & 18.61 &	5.57 & 14.98 &	4.14 & 12.22 \\

        \midrule

        \multirow{3}{*}{SAT~\cite{9710783}} & \multirow{3}{*}{74.78} & \multirow{3}{*}{83.36} & \textbf{CosPGD} &  \textbf{64.68} & \textbf{80.13} &	\textbf{42.74} & \textbf{64.96}	 & \textbf{29.17} & \textbf{52.66} & 	\textbf{17.05} & \textbf{38.75}  \\
        & & & SegPGD &  66.24 & 81.72 &	42.71 & 65.75 &	30.74 & 54.31	 & 20.59 & 43.13 \\
        & & & BIM &   69.89 & 86.68 & 	48.62 & 67.34 &	31.54 & 50.80 &	20.67 & 40.05   \\

        \midrule

        \multirow{3}{*}{DDC-AT~\cite{9710783}} & \multirow{3}{*}{75.98} & \multirow{3}{*}{84.72} & \textbf{CosPGD} &  \textbf{66.93} & \textbf{77.60} &	\textbf{50.79} & \textbf{65.13} &	\textbf{36.12} & \textbf{53.26}	 & \textbf{23.04} & \textbf{41.02}  \\
        & & & SegPGD & 67.09 & 78.36 &	50.89 & 65.14 &	37.70 & 54.48 & 25.40 & 42.72  \\
        & & & BIM &   74.04 & 83.09 &	51.57 & 65.67 &	39.07 & 55.97 &	26.90 & 45.27   \\

         \bottomrule
    \end{tabular}
    }
    \label{tab:ddcat_robust_pspnet}
\end{table}
In \Cref{tab:ddcat_robust_pspnet}, we report the results on the evaluation of CosPGD on \cite{9710783}.
Here we observe that defense methods as in \cite{9710783} might help in reducing some effect of the attacks but not nearly strong enough to negate them and CosPGD is still the strongest adversarial attack.

Please note, we observed some errors in the white-box attack implementation in the \href{https://github.com/dvlab-research/Robust-Semantic-Segmentation}{official GitHub repository} of \cite{9710783}. 
Thus, we were able to reproduce their reported clean accuracies of the three models, i.e. PSPNet with No Defense during training, PSPNet trained with SAT and PSPNet trained with DDC-AT~\cite{9710783}. 
However, as their attack implementation code is wrong, specifically, the normalization done assumes the images to be in the space [0, 1], but in reality they are in [0, 255]. 
Thus, the performance reported by \cite{9710783}, under white-box adversarial attacks is incorrect.
Therefore, we correct these errors and re-run their experiments and extend to them, going as far as 10 attack iterations.
We correct the code from \cite{9710783} and provide the corrected code here: \url{https://github.com/shashankskagnihotri/adv-corrected-ddcat-cospgd}.

\begin{table}[htb]
    \centering
    \caption{Attacking Robust UPerNet~\cite{xiao2018unified} with ConvNeXt-tiny encoder from \cite{croce2023robust} with different fixed attacks in the Segmentation Ensemble Attack (SEA) over different permissible perturbation budgets ($\epsilon$) and attack iterations. $\mathbf{Bold}$ results are the strongest attacks, while $\underline{\textrm{Underlined}}$ results are second strongest.}
    \scalebox{0.63}{
    \begin{tabular}{ccc|cc|cc|cc|cc}
    \toprule
    \multirow{3}{*}{Attack Used} & \multirow{3}{*}{Optimizer Used} & \multirow{3}{*}{Attack Iterations} & \multicolumn{8}{c}{\huge $\frac{\epsilon}{255}$} \\
    & & & \multicolumn{2}{c}{4} & \multicolumn{2}{c}{8} & \multicolumn{2}{c}{12} & \multicolumn{2}{c}{16} \\
    & & & mIoU (\%) & mAcc (\%) & mIoU (\%) & mAcc (\%) & mIoU (\%) & mAcc (\%) & mIoU (\%) & mAcc (\%) \\
    \midrule

    \multirow{7}{*}{\shortstack{SEA: \\only $\textbf{CosPGD (with Softmax) (OURS)}$\\ in \cite{robustbench}}} & \multirow{7}{*}{APGD} & 10 &  $\mathbf{64.17}$ &	$\mathbf{88.52}$	& $\mathbf{43.73}$& 	$\mathbf{76.36}$ &	$\mathbf{21.51}$	 & $\mathbf{55.27}$	& $\mathbf{11.20}$ &	$\mathbf{41.40}$   \\
    & & 20 &  $\mathbf{64.15}$ &	$\mathbf{88.53}$ &	$\mathbf{41.94}$ &	$\mathbf{74.89}$ &	$\mathbf{16.27}$ &	$\mathbf{45.71}$ &	$\mathbf{6.54}$ &	$\mathbf{24.93}$  \\
    & & 30 &  $\mathbf{64.15}$ &	$\mathbf{88.51}$ &	$\mathbf{40.90}$ &	$\mathbf{74.36}$ &	$\mathbf{14.79}$ &	$\mathbf{42.05}$ &	$\mathbf{5.05}$ &	$\mathbf{18.31}$  \\
    & & 40 &  $\underline{64.13}$ &	$\mathbf{88.50}$ &	$\mathbf{40.61}$ &	$\mathbf{74.08}$ &	$\mathbf{14.01}$ &	$\mathbf{39.99}$ &	$\mathbf{4.80}$ &	$\mathbf{16.53}$  \\
    & & 50 &  $\underline{64.10}$ &	$\mathbf{88.50}$ &	$\underline{40.77}$ &	$\mathbf{73.97}$ &	$\mathbf{13.74}$ &	$\mathbf{39.12}$ &	$\mathbf{4.30}$ &	$\mathbf{14.82}$  \\
    & & 100 &  $\underline{64.06}$ &	$\mathbf{88.48}$ &	$\underline{39.99}$ &	$\mathbf{73.29}$ &	$\mathbf{12.67}$ &	$\mathbf{35.97}$ &	$\mathbf{3.29}$ &	$\mathbf{10.69}$  \\
    & & 300 &  $\underline{64.05}$ &	$\underline{88.48}$ &	$\underline{39.52}$ &	$\mathbf{72.81}$ &	$\mathbf{12.66}$ &	$\mathbf{34.63}$ &	$\mathbf{2.90}$ &	$\mathbf{8.78}$  \\

    \midrule

    \multirow{7}{*}{\shortstack{SEA: \\only CosPGD (with Sigmoid) \\in \cite{robustbench}}} & \multirow{7}{*}{APGD} & 10 &  $64.48$ &	$\underline{88.60}$ &	$48.60$ &	$79.47$ &	$31.92$	& $65.45$	& $21.59$ &	$53.70$   \\
    & & 20 &  $64.43$ &	$\underline{88.59}$ &	$46.31$ &	$77.72$ &	$26.37$ &	$57.98$ &	$15.35$ &	$41.19$  \\
    & & 30 &  $64.41$	& $88.58$	& $45.78$ &	$77.22$ &	$24.35$ &	$54.46$ &	$13.18$ &	$34.70$  \\
    & & 40 &  $64.39$ &	$88.58$ &	$45.16$ &	$76.82$ &	$22.89$ &	$52.09$ &	$12.43$ &	$30.88$  \\
    & & 50 &  $64.39$ &	$88.58$ &	$44.95$ &	$76.57$ &	$22.54$ &	$50.91$ &	$11.59$ &	$28.78$  \\
    & & 100 &  $64.37$ &	$88.58$ &	$44.40$ &	$76.13$ &	$21.57$ &	$48.74$ &	$10.53$	& $24.87$  \\
    & & 300 &  $64.37$ &	$88.57$ &	$44.05$ &	$75.96$ &	$21.09$ &	$47.39$ &	$10.23$ &	$22.58$  \\

    \midrule

    \multirow{7}{*}{\shortstack{SEA: \\only $\underline{\textrm{SegPGD}}$\\ in \cite{robustbench}}} & \multirow{7}{*}{APGD} & 10 &     $\underline{64.38}$ &	$88.66$ &	$\underline{44.46}$ &	$\underline{77.21}$ &	$\underline{22.17}$ &	$\underline{58.12}$ &	$\underline{11.37}$ &	$\underline{45.04}$ \\
    & & 20 &  $\underline{64.23}$ &	$\underline{88.59}$ &	$\underline{42.46}$ &	$\underline{75.74}$ &	$\underline{17.89}$ &	$\underline{51.40}$ &	$\underline{8.11}$ &	$\underline{33.86}$   \\
    & & 30 &  $\underline{64.21}$ &	$\underline{88.56}$ &	$\underline{41.71}$ &	$\underline{75.09}$ &	$\underline{16.11}$ &	$\underline{48.30}$ &	$\underline{6.61}$ &	$\underline{28.27}$  \\
    & & 40 &  $\mathbf{64.09}$ &	$\underline{88.52}$ &	$\underline{40.85}$ &	$\underline{74.52}$ &	$\underline{45.05}$ &	$\underline{14.84}$ &	$\underline{5.63}$ &	$\underline{23.90}$  \\
    & & 50 &  $\mathbf{64.01}$ &	$\underline{88.49}$ &	$\mathbf{40.46}$ &	$\underline{74.30}$ &	$\underline{13.98}$ &	$\underline{42.97}$ &	$\underline{4.90}$ &	$\underline{20.85}$  \\
    & & 100 &  $\mathbf{63.95}$ &	$\underline{88.45}$ &	$\mathbf{39.47}$ &	$\underline{73.54}$ &	$\underline{12.78}$ &	$\underline{39.34}$ &	$\underline{4.04}$ &	$\underline{16.26}$  \\
    & & 300 &  $\mathbf{63.80}$ &	$\mathbf{88.41}$ &	$\mathbf{38.69}$ &	$\underline{72.90}$ &	$\underline{11.27}$ &	$\underline{35.85}$ &	$\underline{3.36}$ &	$\underline{12.17}$
  \\

    \bottomrule
    
    \end{tabular}
    }
    \label{tab:UPerNet_SEA_lower_iterations}
\end{table}
In \Cref{tab:UPerNet_SEA_lower_iterations}, we present this evaluation on \cite{croce2023robust} against their robust ``UPerNet~\cite{xiao2018unified} with a ConvNext-tiny backbone'' encoder checkpoint that they make available in their \href{https://github.com/nmndeep/robust-segmentation}{official GitHub repository}. 
We modify their Segmentation Ensemble Attack (SEA)~\cite{croce2023robust} to only include the respective attack mentioned for the given number of attack iterations. 
The optimizer they used is always APGD.

\begin{table}[htb]
    \centering
    \caption{Attacking Robust UPerNet with a ConvNeXt-tiny encoder from \cite{croce2023robust} with CosPGD for extremely high number of iterations i.e. 1200 iterations with $\epsilon=\frac{4}{255}$}
    \scalebox{0.9}{
    \begin{tabular}{ccc|cc}
        \toprule
        \multirow{2}{*}{Attack Method} & \multirow{2}{*}{Optimizer Used} & \multirow{2}{*}{\shortstack{Attack\\Iterations}}& \multicolumn{2}{c}{$\epsilon$=$\frac{4}{255}$} \\
        & & & mIoU (\%) & mAcc (\%) \\
        \midrule

        SEA reported by \cite{croce2023robust} & \multirow{3}{*}{APGD} & \multirow{3}{*}{1200} &  63.800 &	88.300  \\
        SEA~\cite{croce2023robust} reproduced by us & & & 63.670	& 88.320   \\
        replacing SegPGD with CosPGD(softmax) in SEA~\cite{croce2023robust} & & &   63.700 &	88.300  \\

        \bottomrule
    \end{tabular}
    }
    \label{tab:UPerNet_SEA_1200_itrs}
\end{table}
We extent \Cref{tab:UPerNet_SEA_lower_iterations} in \Cref{tab:UPerNet_SEA_1200_itrs}, here we report the results for $\epsilon=\frac{4}{255}$ and observe that the performance is comparable at the extremely high number of iterations i.e. 1200 attack iterations.

W.r.t. the comparison to \cite{croce2023robust} for $\epsilon=4/255$ and very high number of iterations, we would like to highlight that, since the model is trained for this value, the differences between the attacks are actually small. 
Indeed, for high attack iterations, SegPGD is slightly stronger, yielding a maximum difference of 0.25\% in mAcc for 300 iterations versus CosPGD, while at 10 attack iterations, CosPGD is also only slightly stronger than SegPGD in the same range.
However, assuming that \cite{croce2023robust} does not only aim for robustness w.r.t. $\epsilon=4/255$ but aims to generalize (which we infer from their evaluation), it is fair to consider the range of improvement CosPGD reaches over SegPGD for $\epsilon=12/255$ or $\epsilon=16/255$ (scenarios considered in \cite{croce2023robust} as well). 
There, CosPGD decreases the mAcc by almost 10\% more than SegPGD (for 30 iterations), and be more than 3\% more for 300 iterations. 
The general tendency is also that with really high numbers of attack iterations ($>$100 iterations: not commonly considered by peer-reviewed white-box attack works), the differences between CosPGD and SegPGD become smaller, even for $\epsilon$ bounds for which the model has not been trained. 
This is in line with our expectation, coming from the point that CosPGD has smoother gradients and allows to compute better attacks with few iterations, as discussed in \Cref{sec:method}.

\subsection{Evaluating Attacks against SAM}
\label{subsec:appendix:semseg:attacking_sam}
\begin{table}[htb]
    \centering
    \caption{Transfer Attack from DeepLabV3 to SAM~\cite{segment_anything} in a black-box setting on some random samples from PASCAL VOC2012 validation dataset. All Attacks are with $\epsilon=\frac{8}{255}$ and $\alpha$=0.01 with 100 attack iterations. DeepLabV3 was trained for Semantic Segmentation using PASCAL VOC2012 train split.}
    \scalebox{0.7}{
    \begin{tabular}{cccccc}
    \toprule
        \multirow{3}{*}{Original Image} & \multirow{3}{*}{\shortstack{PASCAL VOC2012 \\Ground Truth Mask}} & \multirow{3}{*}{\shortstack{Attacked Image}} & \multirow{3}{*}{\shortstack{Mask predicted by \\DeepLabV3 with \\ResNet50 Backbone}} & \multirow{3}{*}{\shortstack{Mask Predicted by \\Segment Anything \\Model (SAM)}} & \multirow{3}{*}{\shortstack{Observations: SAM’s failure caused only \\under CosPGD Attack (Red Rectangle on \\SAM output under CosPGD attack)}} \\
        & & & & & \\
        & & & & & \\
        \midrule

        \multirow{3}{*}{\includegraphics[width=3cm]{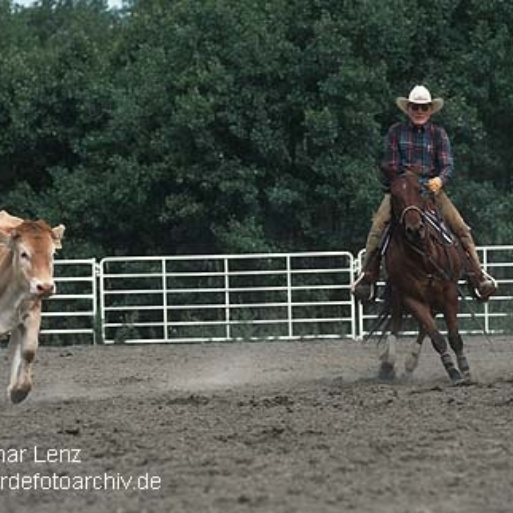}} & \multirow{3}{*}{\includegraphics[width=3cm]{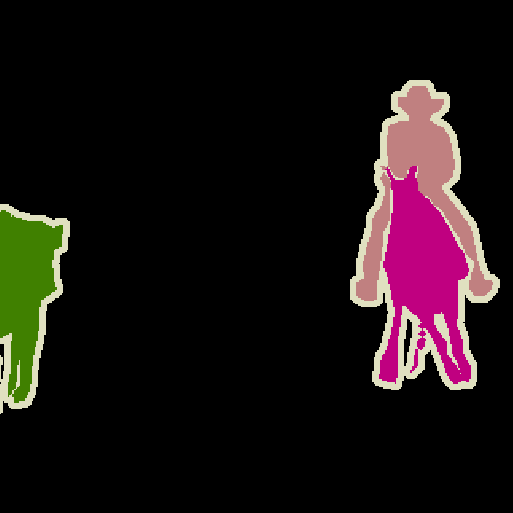}} & \shortstack{No Attack\\ \includegraphics[width=2.5cm]{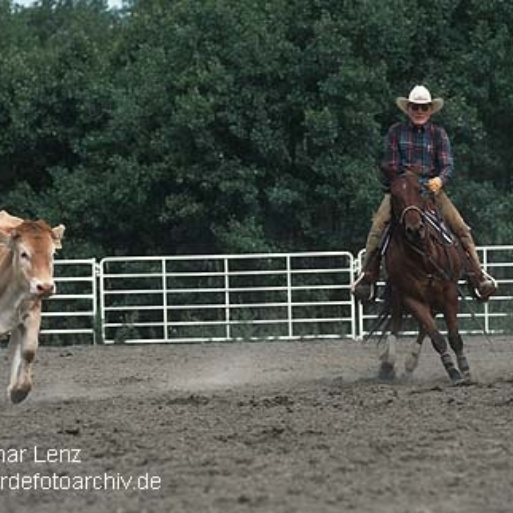}} & \includegraphics[width=3cm]{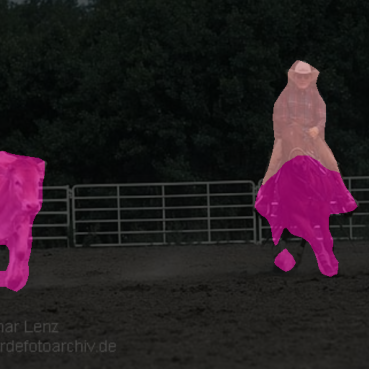} & \includegraphics[width=3cm]{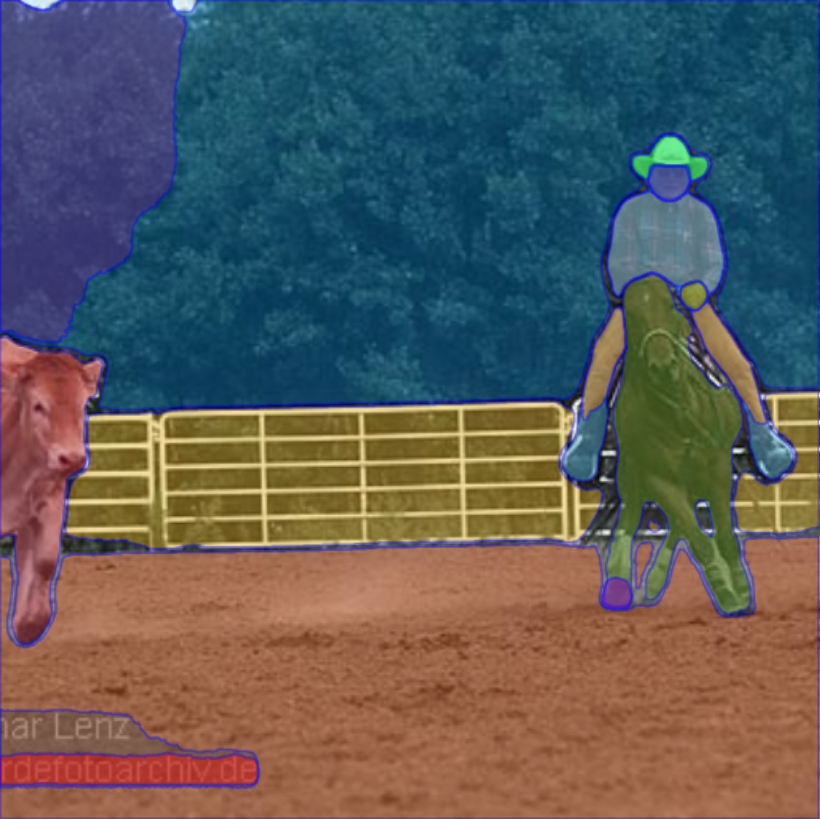} & \multirow{3}{*}{\shortstack{Under No Attack and SegPGD attack, \\SAM is able to detect the cow and \\segment it out in the image. \\However, \textbf{under CosPGD attack,} \\ \textbf{SAM is not able to detect the animal}.}} \\

        & & \shortstack{SegPGD\\ \includegraphics[width=2.5cm]{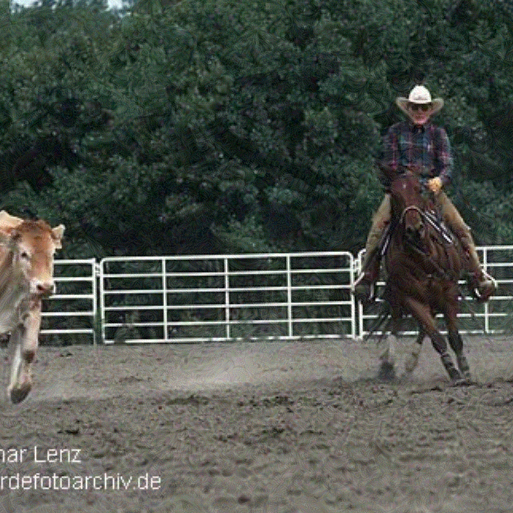}} & \includegraphics[width=3cm]{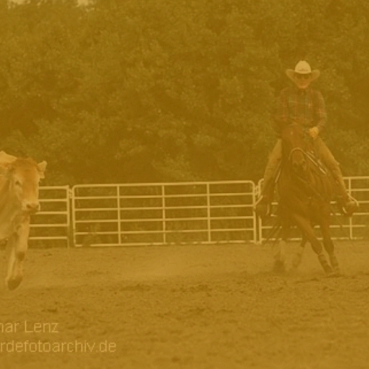} & \includegraphics[width=3cm]{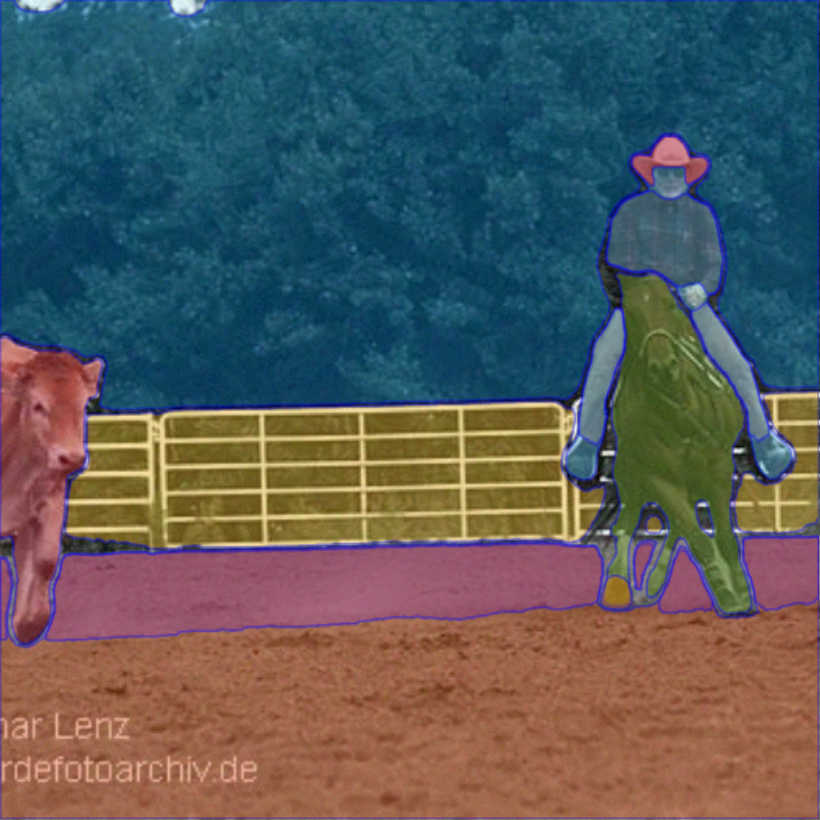} & \\

        & & \shortstack{\textbf{CosPGD}\\ \includegraphics[width=2.5cm]{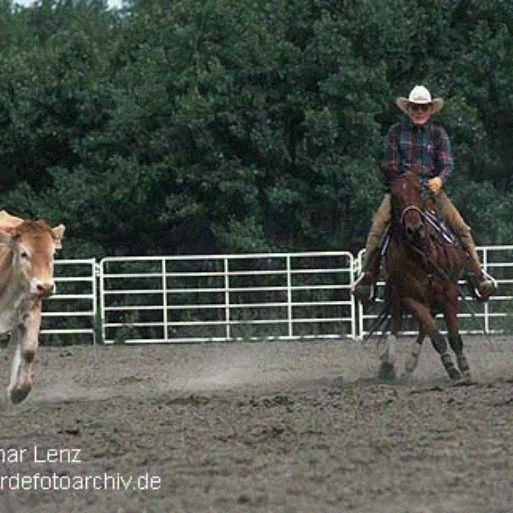}} & \includegraphics[width=3cm]{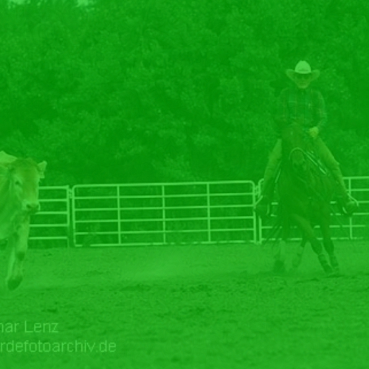} & \includegraphics[width=3cm]{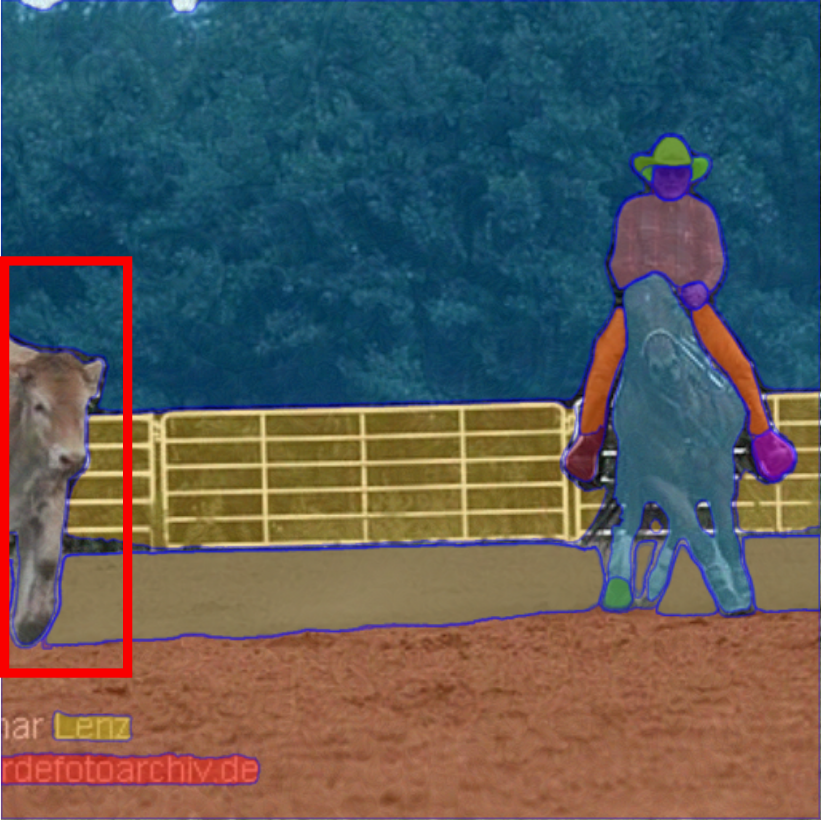} & \\

        \midrule

        \multirow{3}{*}{\includegraphics[width=3cm]{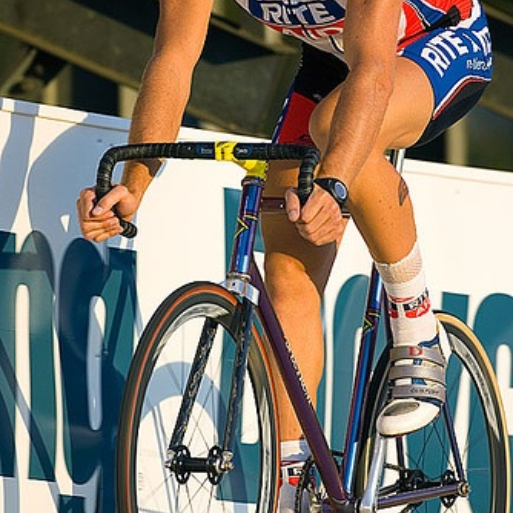}} & \multirow{3}{*}{\includegraphics[width=3cm]{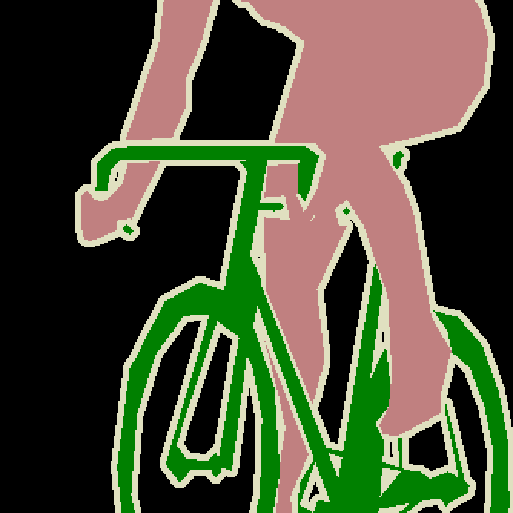}} & \shortstack{No Attack\\ \includegraphics[width=2.5cm]{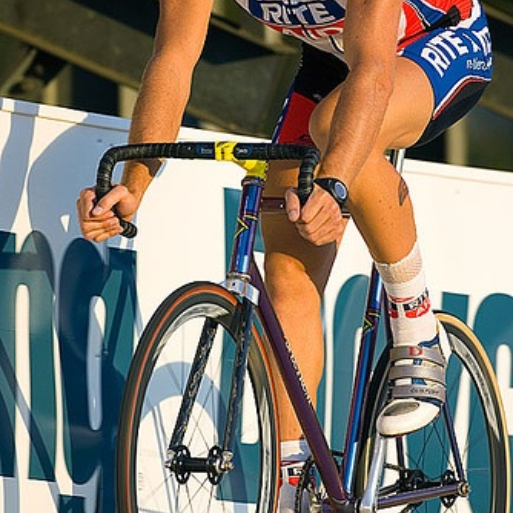}} & \includegraphics[width=3cm]{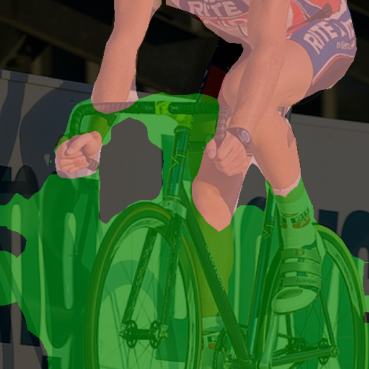} & \includegraphics[width=3cm]{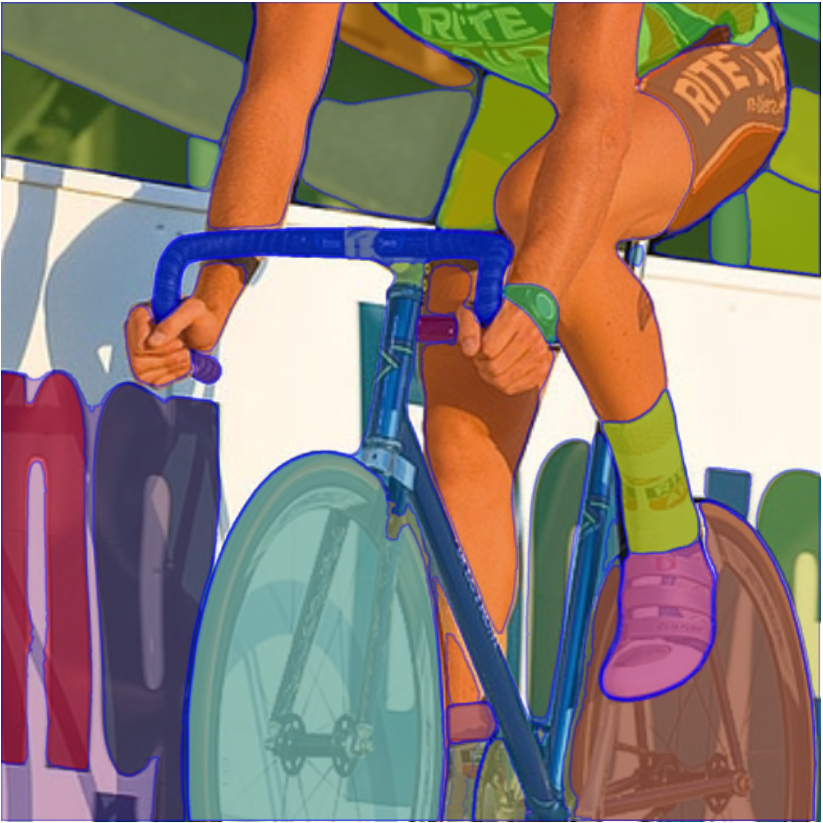} & \multirow{3}{*}{\shortstack{Under No Attack and SegPGD attack, \\SAM is able to detect the the right thigh \\of the cyclist as a single object. \\However, \textbf{under CosPGD attack,} \\ \textbf{SAM is not able to detect this} \\and performs an unnatural split in the \\segmentation masks on the right thigh.}}  \\

        & & \shortstack{SegPGD\\ \includegraphics[width=2.5cm]{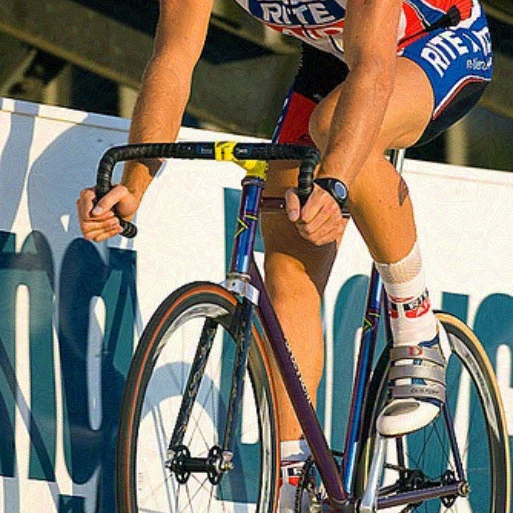}} & \includegraphics[width=3cm]{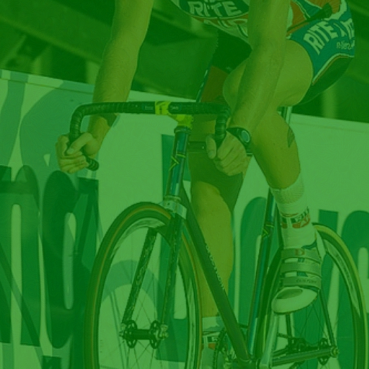} & \includegraphics[width=3cm]{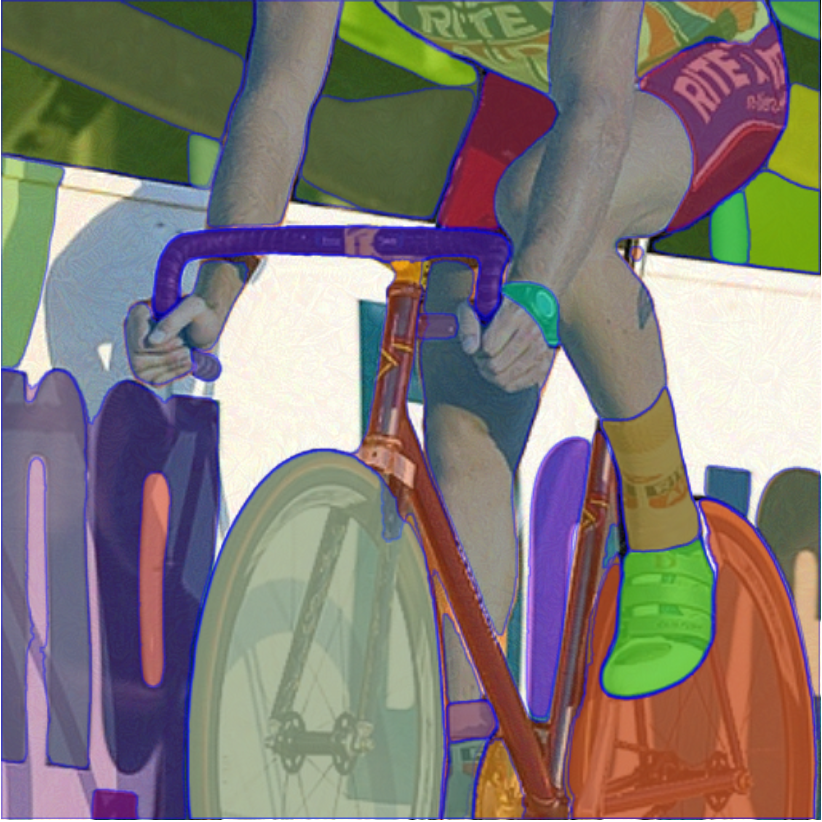} & \\

        & & \shortstack{\textbf{CosPGD}\\ \includegraphics[width=2.5cm]{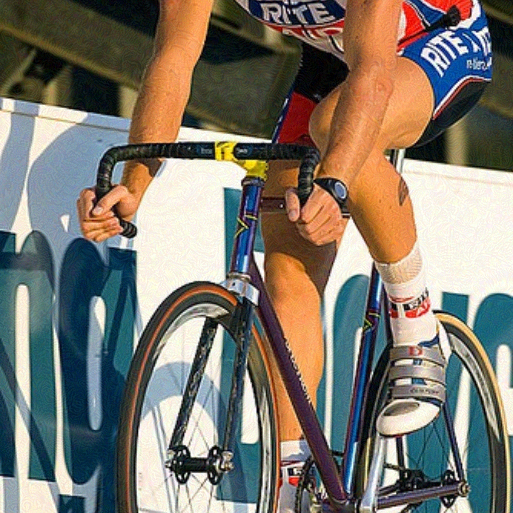}} & \includegraphics[width=3cm]{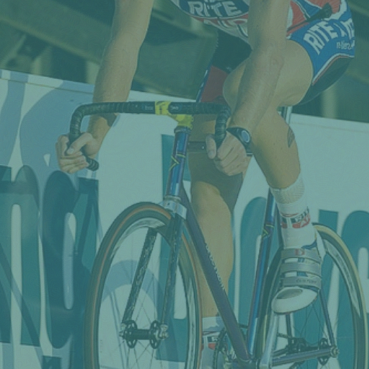} & \includegraphics[width=3cm]{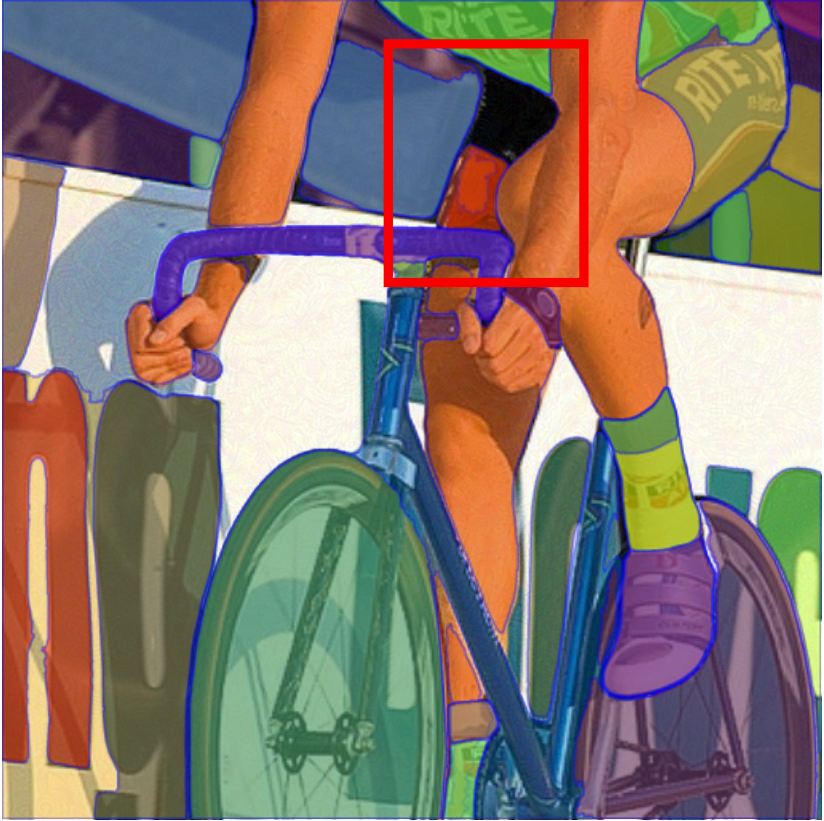} & \\

        \midrule

        \multirow{3}{*}{\includegraphics[width=3cm]{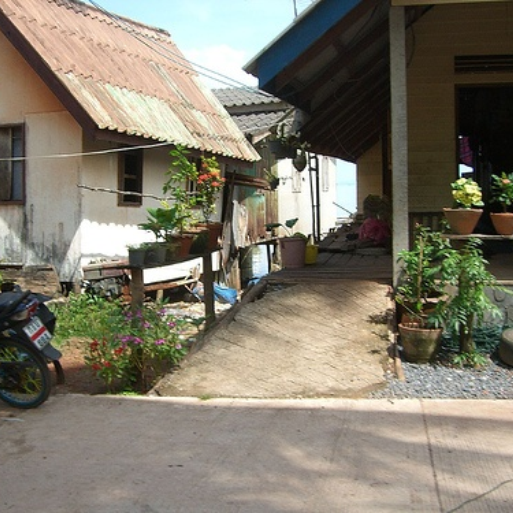}} & \multirow{3}{*}{\includegraphics[width=3cm]{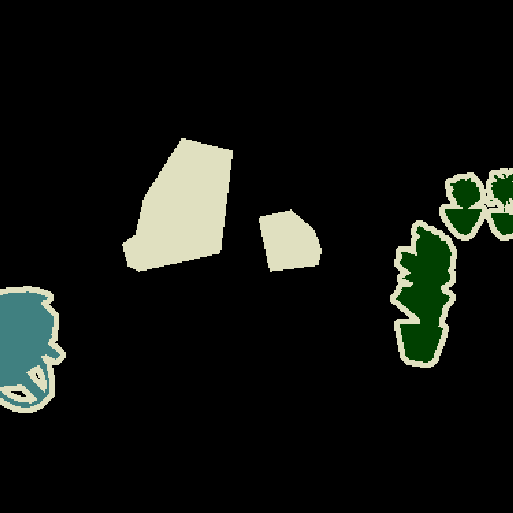}} & \shortstack{No Attack\\ \includegraphics[width=2.5cm]{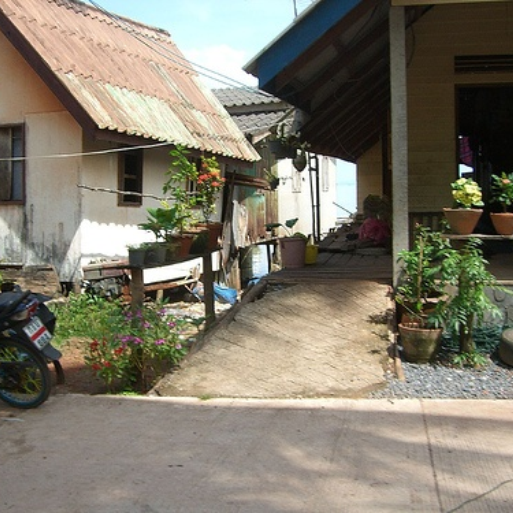}} & \includegraphics[width=3cm]{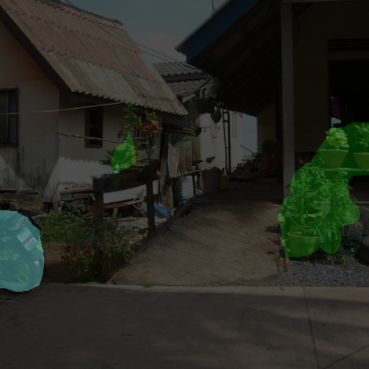} & \includegraphics[width=3cm]{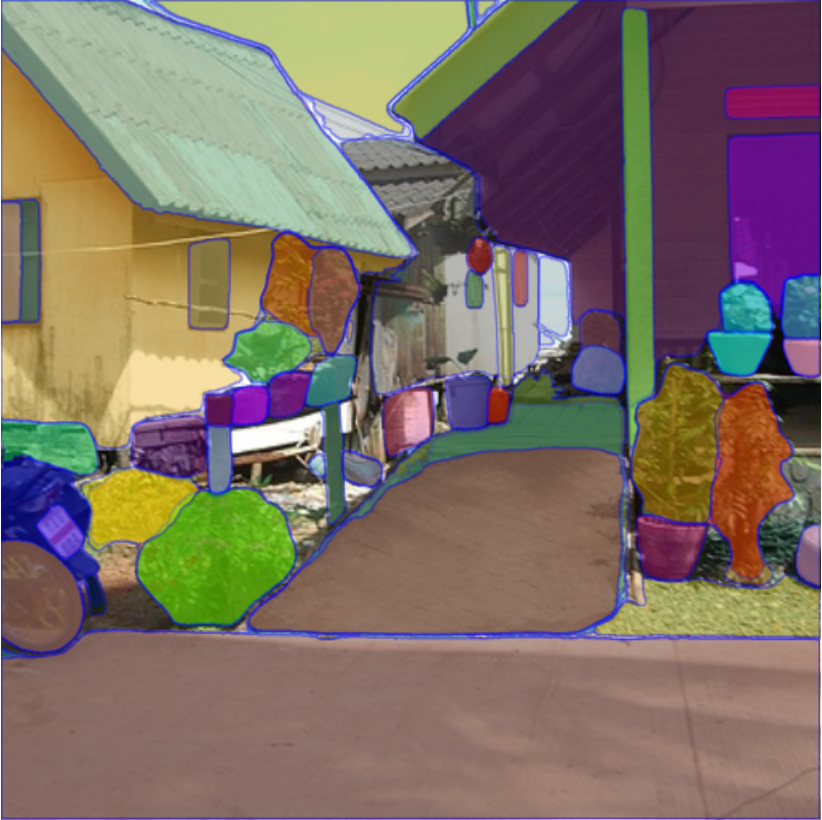} & \multirow{3}{*}{\shortstack{Under No Attack and SegPGD attack, \\SAM is able to detect the road in the centre \\of the frame and segment it out in the image. \\However, \textbf{under CosPGD attack,} \\ \textbf{SAM is not able to detect the road}.}} \\

        & & \shortstack{SegPGD\\ \includegraphics[width=2.5cm]{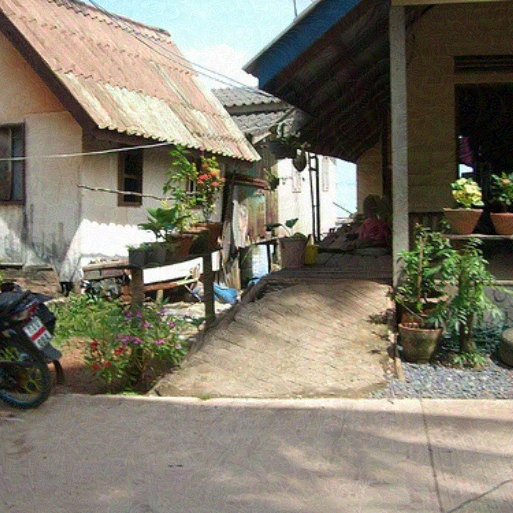}} & \includegraphics[width=3cm]{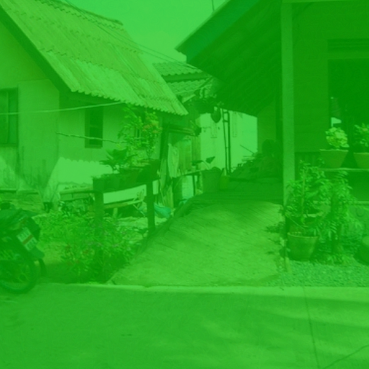} & \includegraphics[width=3cm]{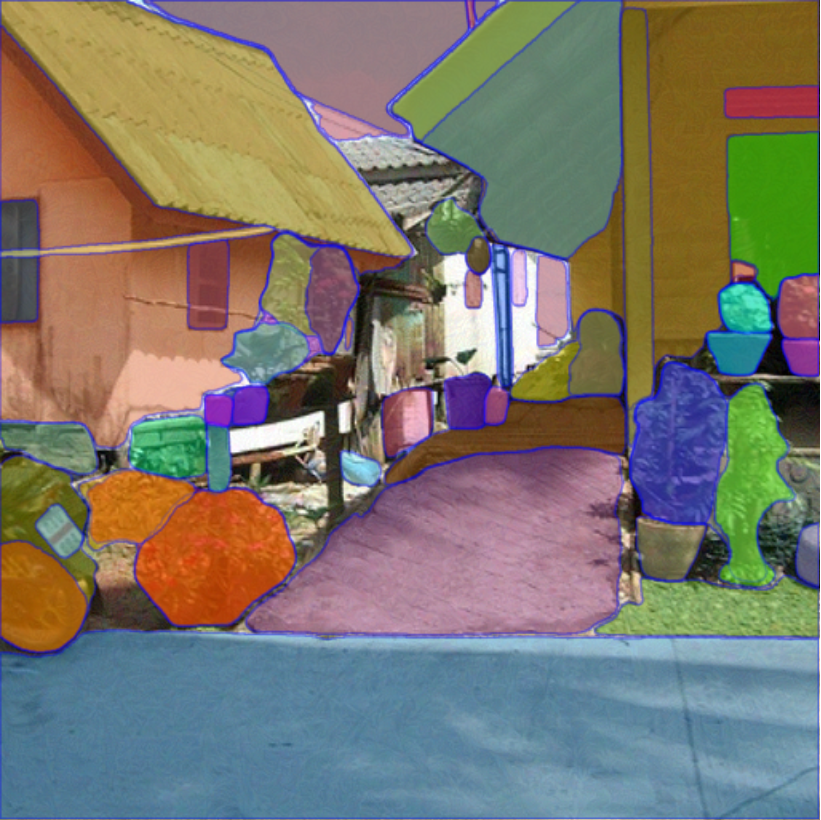} & \\

        & & \shortstack{\textbf{CosPGD}\\ \includegraphics[width=2.5cm]{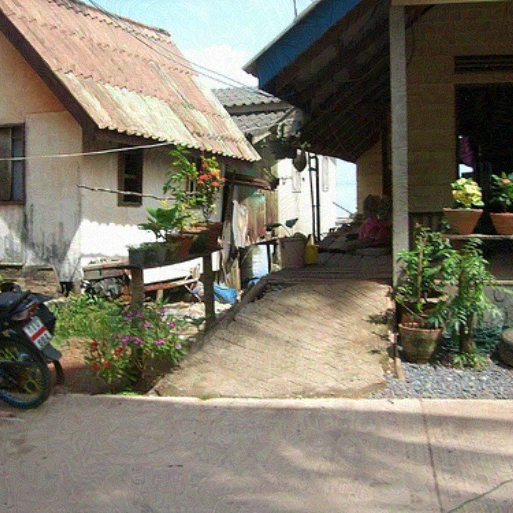}} & \includegraphics[width=3cm]{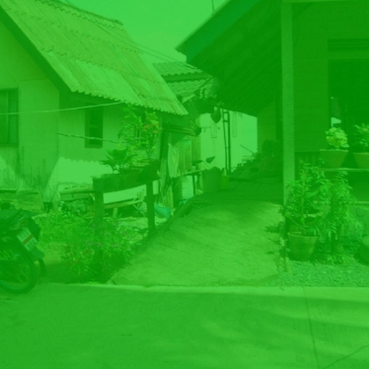} & \includegraphics[width=3cm]{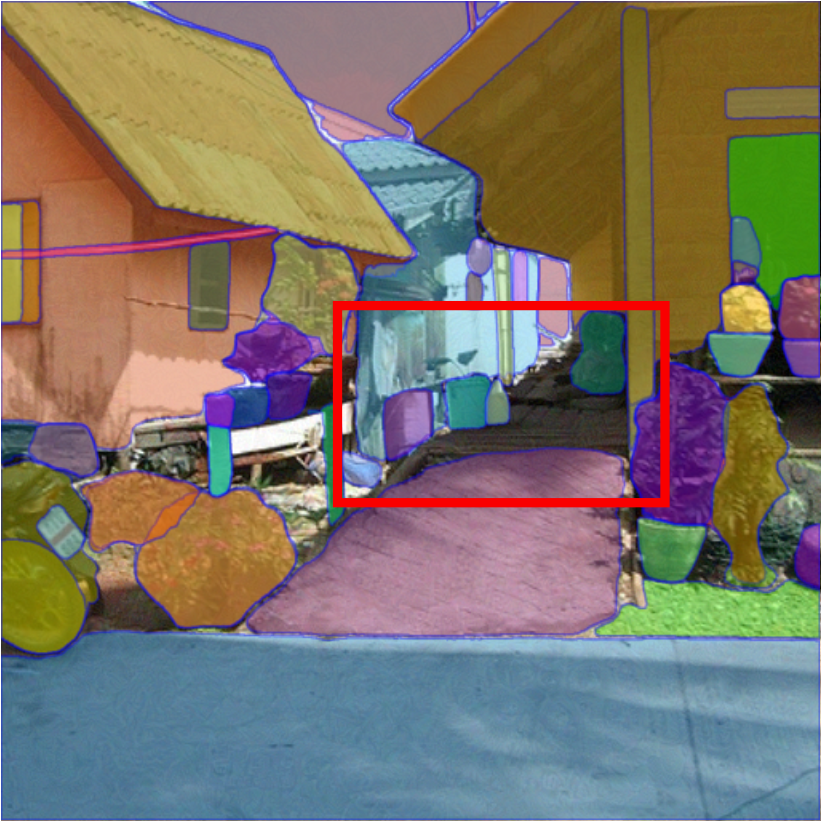} & \\

        \bottomrule
    \end{tabular}
    }
    \label{tab:transger_attack_SAM}
\end{table}

In \Cref{tab:transger_attack_SAM}, we show that when we attack a DeepLabV3 with a ResNet50 encoder on PASCAL VOC2012 images, and transfer the 100 iterations attack to SAM~\cite{segment_anything}, only the CosPGD attack can cause failures in the segmentation masks. 
SegPGD fails to create failures in the segmentation masks of SAM, when compared to its segmentation masks on a clean image. 

Note that these are just random sample results, as quantitative evaluation would be invalid.
This is because the publicly available version of SAM does not perform semantic segmentation (which is segmentation with class labels). 
SAM merely predicts segmentation masks without assigning them any class labels, and current variants of SAM used for Semantic Segmentation, for example in this \href{https://github.com/fudan-zvg/Semantic-Segment-Anything}{GitHub repository} perform worse than the other models we considered for this task. 
Furthermore, the masks produced by SAM are often finer than the ground truth masks of most datasets, making the calculation of metrics like mIoU invalid.

\subsection{Semantic Segmentation with UNet on Cityscapes}
\label{subsec:exp_results:unet}
In the following, we provide extra results on semantic segmentation with UNet on the Cityscapes dataset.

\subsubsection{Implementation Details}
In this evaluation, we use a UNet architecture~\citep{unet} with a ConvNeXt\_tiny encoder~\citep{convnext}. 
We extend the implementation from \cite{unet_convnext_github}(\url{www.github.com}) to implement CosPGD, PGD, and SegPGD non-targeted 
$l_{\infty}$-norm and $l_{2}$-norm attacks.

We do these evaluations on the Cityscapes dataset~\citep{cordts2016cityscapes}.
Cityscapes contains a total of 5000 high-quality images and pixel-wise annotations for urban scene understanding.
The dataset is split into 2975, 500, and 1525 images for training, validation, and testing respectively.
The model is trained on the test split and attacks are evaluated on the validation split.

\subsubsection{Experimental Results and Discussion}
\label{subsubsec:exp_results:unet:results}
\begin{figure*}
    \centering
    \includegraphics[width=0.435\textwidth]{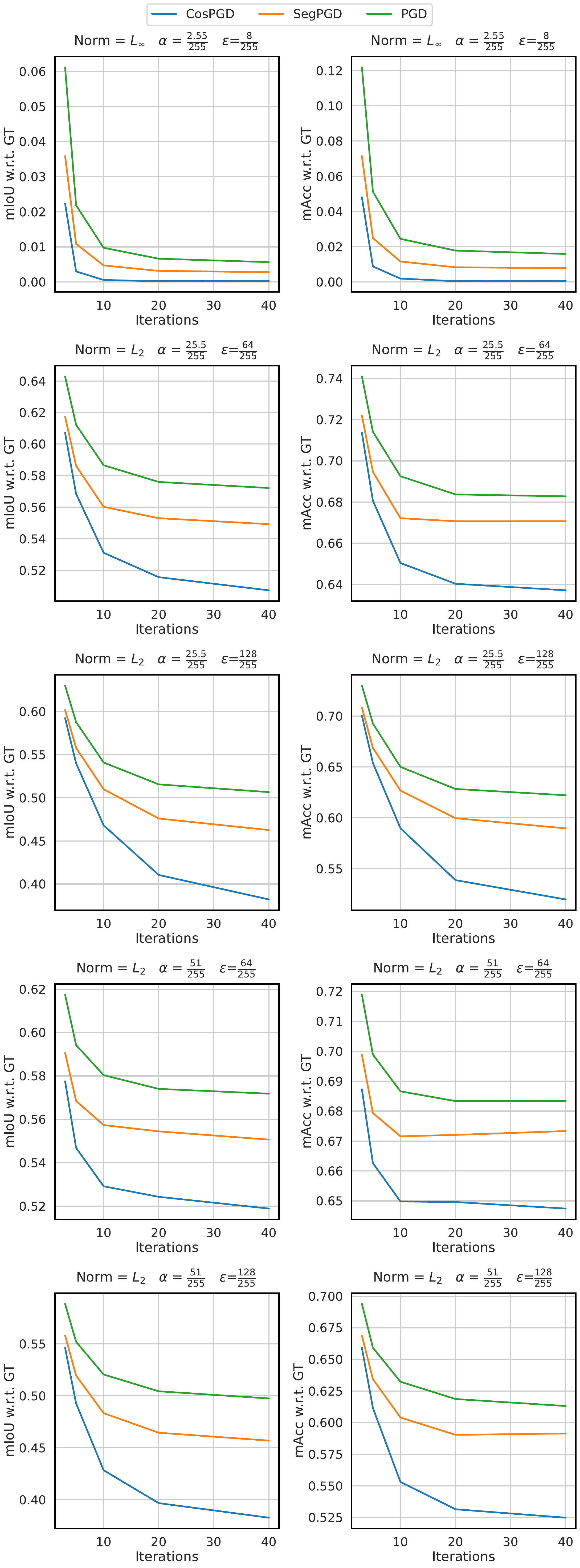}
    \caption{Comparing non-targeted CosPGD to PGD and SegPGD attacks across iterations and $l_p$-norm constraints, and $\alpha$ and $\epsilon$ values using UNet architecture with a ConvNeXt tiny encoder on Cityscapes validation dataset. CosPGD significantly outperforms previous methods by a large margin, even at few attack iterations. }
    \label{fig:unet_cityscapes}
\end{figure*}

In Figure~\ref{fig:unet_cityscapes}, we report results from the comparison of non-targeted CosPGD to PGD and SegPGD attacks across iterations and across $l_p$-norm constraints: $l_{\infty}$-norm and $l_2$-norm using UNet architecture with a ConvNeXt tiny encoder on Cityscapes validation dataset.
For the $l_{\infty}$-norm constraint, we use the same $\alpha=0.01$ and $\epsilon\approx\frac{8}{255}$ as in all previous evaluations.
For the $l_{2}$-norm constraint we follow common work~\citep{robustbench, l2norm} and use the same $\epsilon$ for CosPGD, SegPGD, and PGD i.e. $\epsilon\approx$\{$\frac{64}{255}, \frac{128}{255}$\} and $\alpha=$\{0.1, 0.2\}.

Note, SegPGD has been proposed as an $l_{\infty}$-norm constrained attack. 
We extend it to the $l_{2}$-norm constraint merely for complete comparison and curiosity.

We observe in Figure~\ref{fig:unet_cityscapes} that CosPGD is a significantly stronger attack than both PGD and SegPGD, across iterations and $l_p$-norm constraints, and $\alpha$ and $\epsilon$ values. Even at low attack iterations, it outperforms previous methods significantly, making it particularly efficient. 
Especially as an $l_{2}$-norm constrained attack, as shown before in Figure~\ref{fig:all_deeplab_l2} for DeepLabV3 on PASCAL VOC 2012 dataset and discussed before in Section~\ref{subsec:exp:semseg}, as attack iterations increase, CosPGD can increase the performance gap quite significantly. 

\subsection{Ablation on Attack Step Size $\alpha$}
\label{subsec:exp_results:deeplabl2}
Further, we provide additional experimental results and ablation studies using DeepLabV3 for semantic segmentation on the PASCAL VOC 2012 validation dataset.

\subsubsection{$l_{2}$-norm constrained adversarial attacks}
\label{subsec:exp_results:l_2}
\begin{figure}
    \centering
    \includegraphics[width=0.485\textwidth]{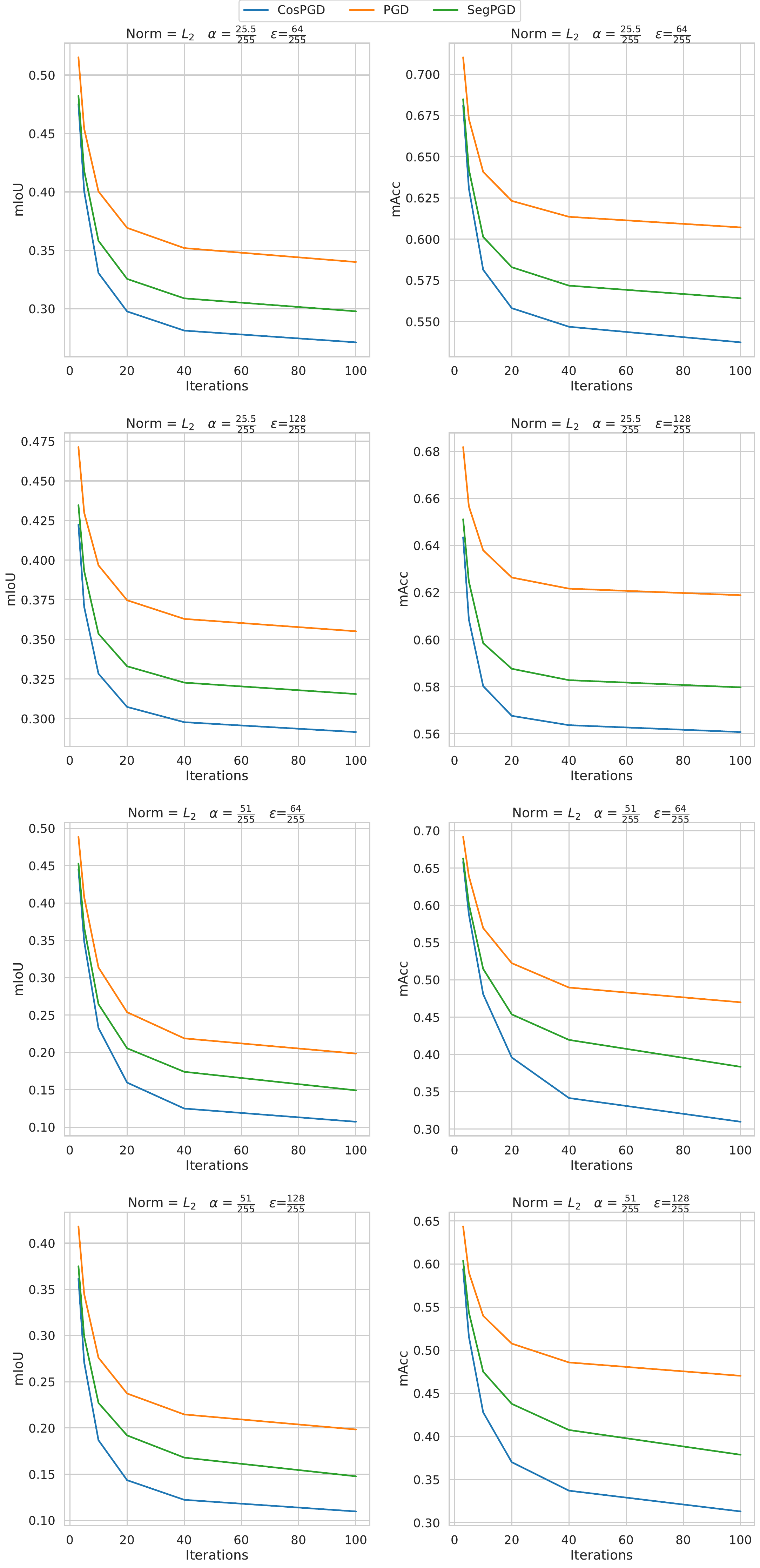}
    \caption{Comparing CosPGD to PGD and SegPGD across iterations as $l_2$-norm constrained attacks, and across $\alpha$ and $\epsilon$ values using DeepLabV3 architecture with a ResNet50 on PASCAL VOC 2012 validation dataset. Again, CosPGD outperforms previous attacks be a large margin at all attack iterations.}
    \label{fig:all_deeplab_l2}
\end{figure}

Further in Figure~\ref{fig:all_deeplab_l2}, 
we report $l_2$-norm constrained attack evaluations on commonly used \citep{robustbench, l2norm} values of $\epsilon\approx$\{$\frac{64}{255}, \frac{128}{255}$\} and $\alpha=$\{0.1, 0.2\}.

\textcolor{black}{Additionally, in Table~\ref{tbl:exp:semseg_l2_cw} we provide comparison to C\&W~\cite{c_and_w} and other $l_2$-norm constrained adversarial attacks with $\alpha$=0.2 and $epsilon\approx\frac{128}{255}$ on PASCAL VOC 2012 validation dataset using DeepLabV3 with a ResNet50 backbone.}
\begin{table*}[htb]
\caption{\textcolor{black}{Comparison of performance of CosPGD to SegPGD, PGD and C\&W as a $l_2$-norm constrained attack with $\alpha$=0.2 and $\epsilon\approx\frac{128}{255}$ where applicable for semantic segmentation over PASCAL VOC2012 validation dataset. We observe that CosPGD is a significantly stronger attack compared to all the other attacks for both metrics.}}
\label{tbl:exp:semseg_l2_cw}
\centering
\scalebox{.5}{
\begin{tabular}{p{1.5cm}ccc|cc|cc|cc|cc|cc}
\toprule
 \multirow{3}{3cm}{\textbf{Network}} & \multirow{3}{3cm}{\textbf{Attack method}} &  \multicolumn{12}{c}{\textbf{Attack iterations}} \\
& & \multicolumn{2}{c}{3} & \multicolumn{2}{c}{5} & \multicolumn{2}{c}{10}  &  \multicolumn{2}{c}{20} &  \multicolumn{2}{c}{40} &  \multicolumn{2}{c}{100} \\
 &  & mIoU(\%) & mAcc(\%) &  mIoU(\%) & mAcc(\%) & mIoU(\%) & mAcc(\%) & mIoU(\%) & mAcc(\%) & mIoU(\%) & mAcc(\%)  & mIoU(\%) & mAcc(\%)\\
\midrule

\multirow{3}{1cm}{\textbf{DeepLabV3}} & \textbf{C\&W}~(c=1) & 72.35 & 84.32 & 72.02 & 84.13 & 71.87 & 84.05 & 71.81 & 84.02 & 71.78 & 84.01 & 71.77 & 84.00   \\

& \textbf{PGD} & 41.81 & 64.36 & 34.5 & 59.03 & 27.61 & 54.0 & 23.73 & 50.77 & 21.47 & 48.58 & 19.84 & 47.04 \\

& \textbf{SegPGD} & 37.51 & 60.4 & 29.9 & 54.4 & 22.72 & 47.51 & 19.2 & 43.78 & 16.8 & 40.75 & 14.77 & 37.88 \\
 
  & \textbf{CosPGD} & \textbf{36.17} & \textbf{59.41} & \textbf{27.12} & \textbf{51.6} & \textbf{18.68} & \textbf{42.8} & \textbf{14.35} & \textbf{37.02} & \textbf{12.23} & \textbf{33.71} & \textbf{10.97} & \textbf{31.3}  \\
\bottomrule

\end{tabular}
}
\end{table*}

\subsubsection{\revision{$l_{\infty}$-norm constrained adversarial attacks}}
\label{subsec:exp_results:l_inf}
\begin{figure*}
    \centering
    \includegraphics[width=.55\textwidth]{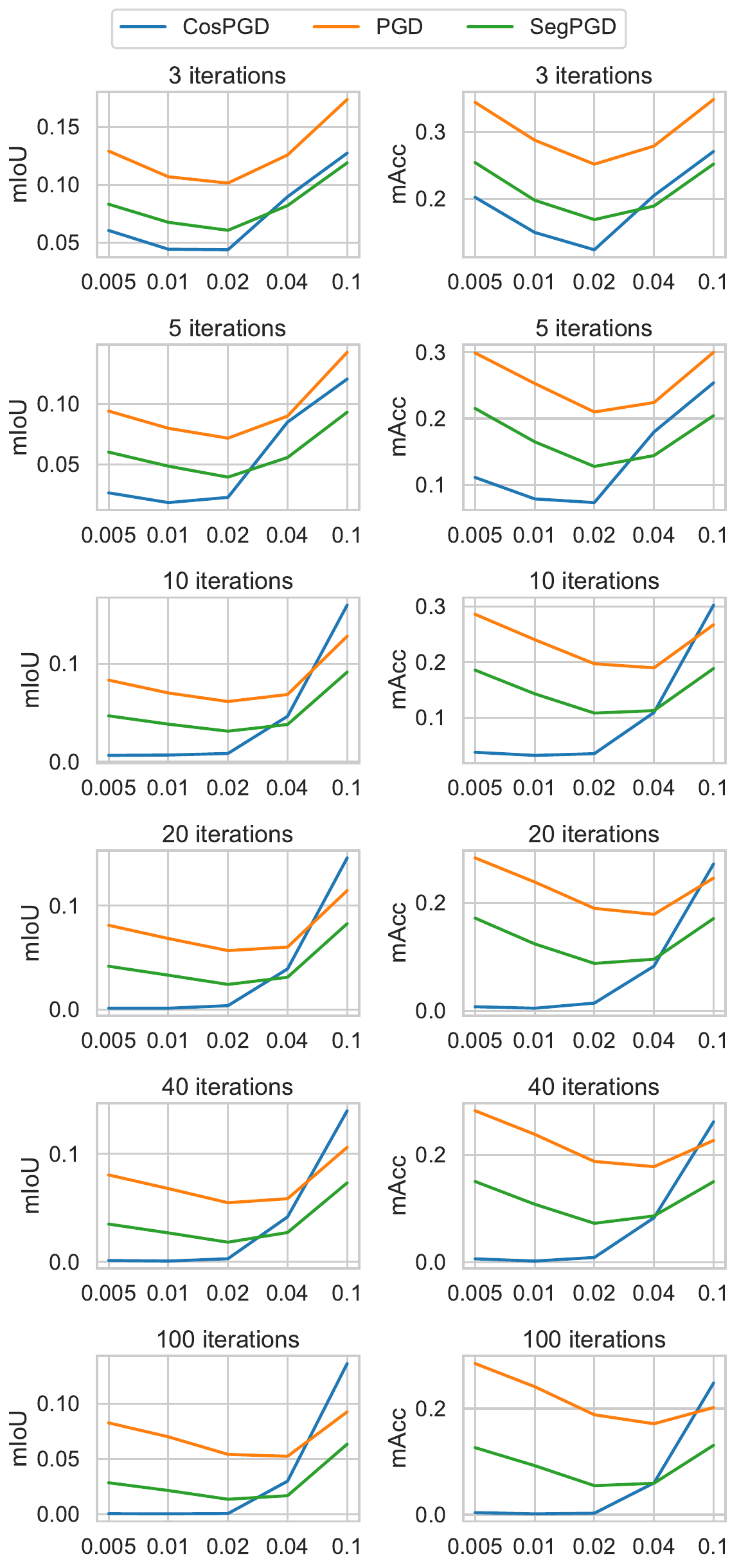}
    \caption{\revision{
    We ablate step sizes $\alpha$ for $l_{\infty}$-norm constrained CosPGD, SegPGD, and PGD attacks given different number of iterations $\in\{3,5,10,20,40,100\}$ by attacking DeepLabV3 trained on the PASCAL VOC2012 dataset with maximal perturbation of $\epsilon=0.03$.
    We can observe that the scaling in CosPGD ensures less susceptibility to the choice of step size given that it is set small enough ($\alpha \leq \epsilon$).
    }}
    \label{fig:ablation_alpha}
\end{figure*}

\revision{Following, we ablate over the attack step size $\alpha$ for the $l_{\infty}$-norm constrained adversarial attacks and report the findings in Figure~\ref{fig:ablation_alpha}.
We consider $\alpha\in\{0.005, 0.01, 0.02, 0.04, 0.1\}$.
We can observe that the scaling in CosPGD ensures less susceptibility to the choice of step size given that it is set small enough ($\alpha \leq \epsilon$).
In our work, we use step size $\alpha$=0.01 to maintain consistency with previous work~\citep{pgd, segpgd}.}

\subsection{Tabular Results}
\label{subsec:appendix:results:tables}
Here we report the quantitative results that have already been presented in the main paper in Figures~\ref{fig:semseg:semseg_pgd}in tabular form. 
For the results reported in Figure~\ref{fig:semseg:semseg_pgd}, we report the results in tables~\ref{tbl:exp:semseg_pgd}. Here we observe that at low attack iterations (iterations=3) SegPGD implies that PSPNet is more adversarially robust than both DeepLabV3. However, after more attack iterations (iterations $\geq$ 5), SegPGD correctly implies that DeepLabV3 is more robust than PSPNet.
Contrary to this, CosPGD even at low attack iterations correctly predicts DeepLabV3 to be more robust than PSPNet.
This is an insight that CosPGD provides with considerably less computation.

\begin{table*}[htb]
\caption{Comparison of performance of CosPGD to SegPGD for semantic segmentation over PASCAL VOC2012 validation dataset. We observe that CosPGD is a significantly stronger attack compared to SegPGD for both metrics and all models.
\label{tbl:exp:semseg_pgd}}
\centering
\scalebox{.6}{
\begin{tabular}{p{1.5cm}ccc|cc|cc|cc|cc|cc}
\toprule
 \multirow{3}{3cm}{\textbf{Network}} & \multirow{3}{3cm}{\textbf{Attack method}} &  \multicolumn{12}{c}{\textbf{Attack iterations}} \\
& & \multicolumn{2}{c}{3} & \multicolumn{2}{c}{5} & \multicolumn{2}{c}{10}  &  \multicolumn{2}{c}{20} &  \multicolumn{2}{c}{40} &  \multicolumn{2}{c}{100} \\
 &  & mIoU(\%) & mAcc(\%) &  mIoU(\%) & mAcc(\%) & mIoU(\%) & mAcc(\%) & mIoU(\%) & mAcc(\%) & mIoU(\%) & mAcc(\%)  & mIoU(\%) & mAcc(\%)\\
\midrule
 \multirow{2}{1cm}{\textbf{UNet}} & \textbf{SegPGD} & 12.38 & 32.41 & 7.75 & 25.27 & 4.46 & 18.36  & 2.98  & 14.24 & 2.20 & 11.66 & 1.55 & 8.66 \\
  & \textbf{CosPGD} & \textbf{9.67}  & \textbf{29.46} & \textbf{3.71} & \textbf{15.89}  & \textbf{0.61} & \textbf{3.39} & \textbf{0.06} & \textbf{0.38} & \textbf{0.03} & \textbf{0.16} & \textbf{0.01} & \textbf{0.04} \\
\midrule
 \multirow{3}{1cm}{\textbf{PSPNet}} & \textbf{PGD} & 13.79 & 31.91 & 7.59 & 21.15 & 5.44 & 16.96 & 4.48 & 14.78  & 3.80  & 13.13 & 3.72 & 13.21  \\
  & \textbf{SegPGD} & 9.19 & 23.25 & 4.70 & 14.25 & 2.72 & 9.50 & 1.82 & 7.39  & 1.30  & 5.77 & 0.83 & 3.86  \\
 
  & \textbf{CosPGD} & \textbf{7.03}  & \textbf{19.73} & \textbf{2.15} & \textbf{7.60} & \textbf{0.408} & \textbf{1.44} & \textbf{0.04} & \textbf{0.11} & \textbf{0.005} & \textbf{0.021} & \textbf{0.0002} & \textbf{0.0007} \\
  \midrule
\multirow{5}{1cm}{\textbf{DeepLabV3}} & \textbf{PGD} & 10.69  & 28.76 & 8.00 & 25.29 & 7.02 & 24.05 & 6.84  & 23.87  & 6.79  &  23.81 & 7.01 & 24.13 \\

& \textbf{BIM} & 10.86 & 29.39 & 7.75 & 24.97  & 6.95 & 24.06 & 6.67 & 23.52 & 6.57 & 23.48 & -- & -- \\

& \textbf{APGD} & 13.74 & 29.79 & 8.67 & 22.46  & 6.50 & 19.82 & 6.11 & 18.99 & 5.30 & 17.04 & 5.14 & 16.72 \\

& \textbf{SegPGD} & 6.76  & 19.78 & 4.86 & 16.49 & 3.84 & 14.29 & 3.31  & 12.40  & 2.69  &  10.81 & 2.15 & 9.25 \\
 
  & \textbf{CosPGD} & \textbf{4.44}  & \textbf{14.97} & \textbf{1.84} & \textbf{7.89} & \textbf{0.69} & \textbf{3.18} & \textbf{0.12} & \textbf{0.48} & \textbf{0.08} & \textbf{0.25} & \textbf{0.005} & \textbf{0.16} \\
\bottomrule

\end{tabular}
}
\end{table*}

\subsection{\textcolor{black}{Adversarial Training}}
\label{subsec:appendix:adversarial_training}
\begin{figure*}
    \centering 
    \begin{tabular}{@{}c@{\hspace{0.18cm}}c@{\hspace{0.04cm}}c@{\hspace{0.04cm}}c@{}}
    Clean Image and GT Mask &  &  Trained with CosPGD&  Trained with SegPGD \\
\includegraphics[width=0.3\linewidth]{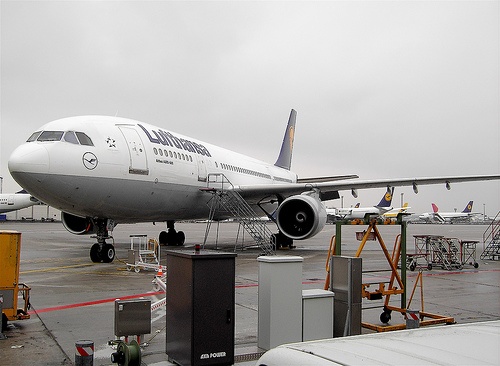}
&\rotatebox{90}{\phantom{sus}CosPGD Attack}& \includegraphics[width=0.3\linewidth]{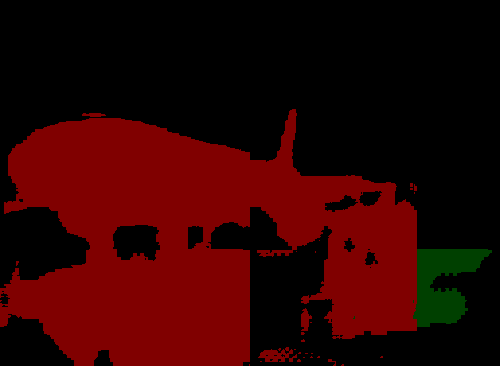}
&
\includegraphics[width=0.3\linewidth]{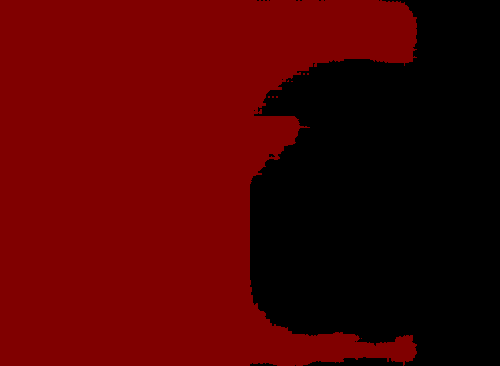}
 \\
\includegraphics[width=0.31\linewidth]{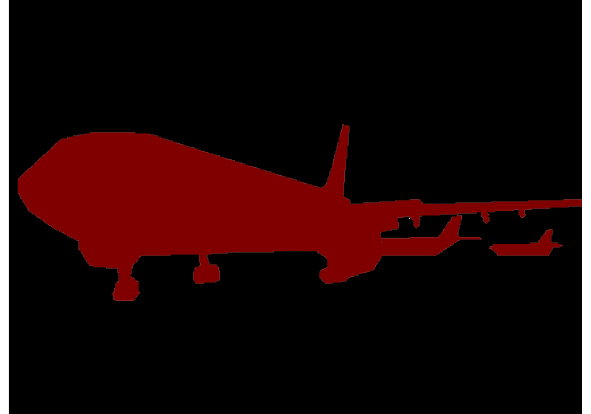}
&\rotatebox{90}{\phantom{sus}SegPGD Attack}&
\includegraphics[width=0.3\linewidth]{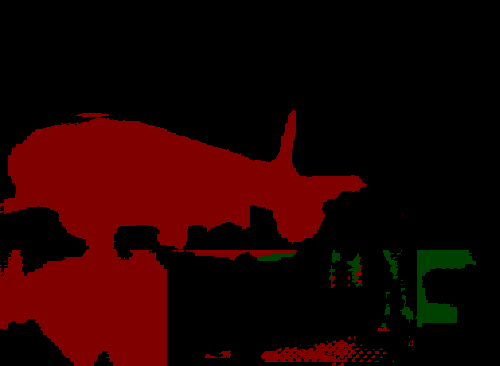}
&
\includegraphics[width=0.3\linewidth]{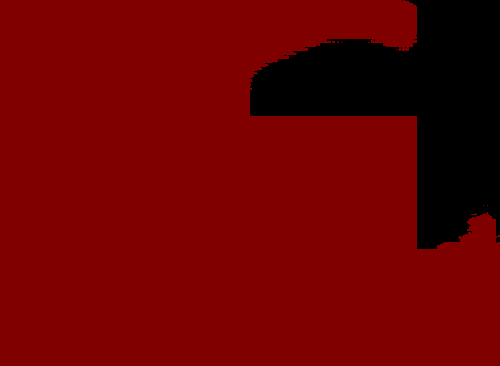}
 \\
\end{tabular}

\caption{\textcolor{black}{Predictions using UNet with ConvNeXt backbone on PASCAL VOC2012 validation dataset after 100 iterations adversarial attacks on adversarially trained models. We observe that the models adversarially trained with CosPGD are predicting reasonable masks even after 100 attack iterations, while the model trained with SegPGD is providing much worse results under both SegPGD and CosPGD attacks.}}
\label{fig:adv_training_unet}
\end{figure*}

\textcolor{black}{In Figure~\ref{fig:adv_training_unet} we show the segmentation masks predicted by UNet after being adversarially trained. We observe that even after 100 attack iterations, the model adversarially trained using CosPGD is making reasonable predictions. 
However, the model trained with SegPGD is merely predicting a blob.}

\textcolor{black}{In Table~\ref{table:adv_training_unet} we report the performance of models trained with various adversarial attacks against different commonly used adversarial attacks across multiple attack iterations. 
We observe that the model trained with CosPGD performs the best against all considered adversarial attacks. 
The models were trained with 3 attack iterations of the respective ``Training Method" attack during training.}

\begin{figure*}
    \centering
    \includegraphics[width=\linewidth]{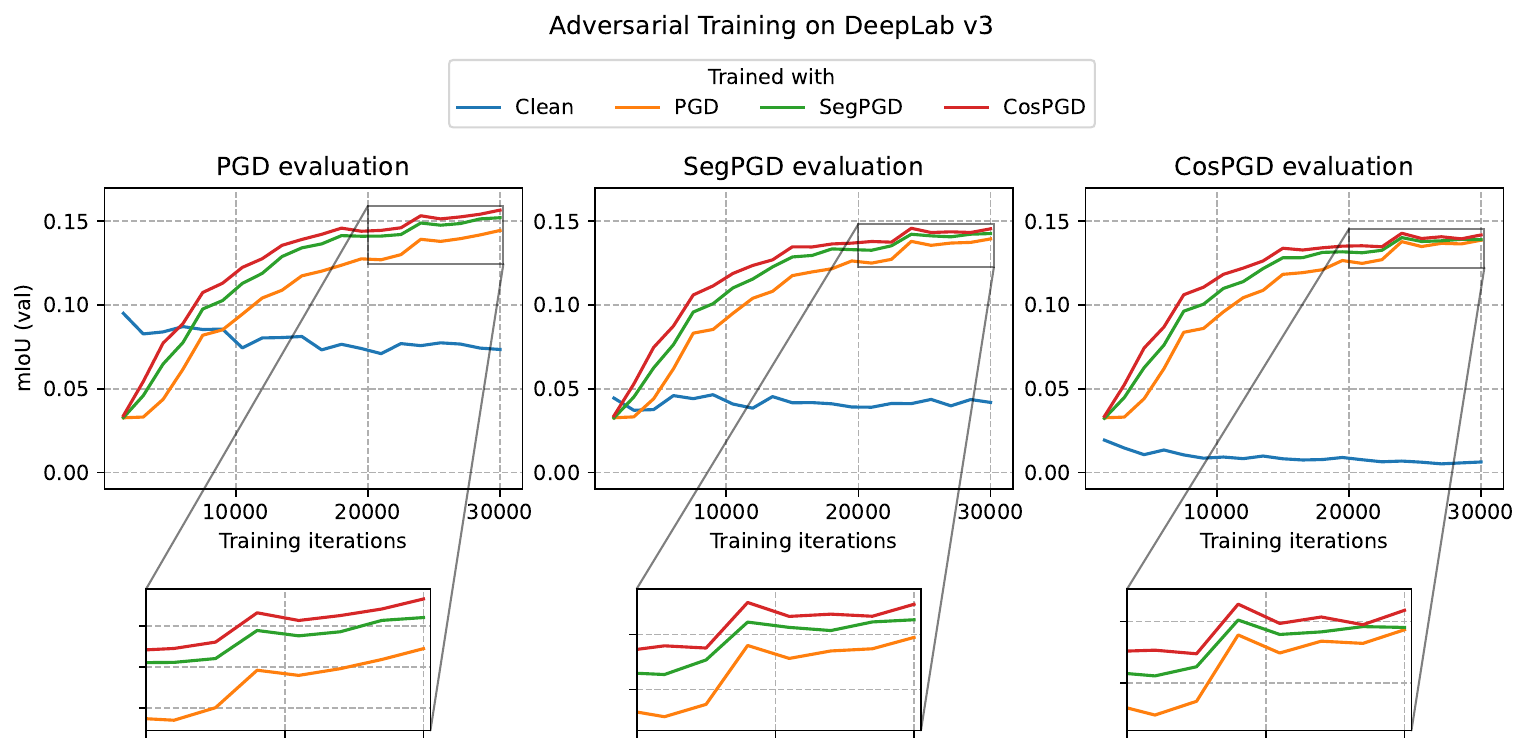}
    \caption{DeepLabV3 adversarially trained using different adversarial attacks for 3 iterations during training using 50\% of the minibatch for generating adversarial samples. All checkpoints are evaluated against 10 attack iterations of the respective attacks. We observe that the model trained with CosPGD outperforms all other adversarial training methods considered against all attacks.}
    \label{fig:adv_training:deeplabv3}
\end{figure*}

\textcolor{black}{In Figure~\ref{fig:adv_training:deeplabv3} we present the training curves for training DeepLabV3 on the PASCAL VOC2012 training dataset using adversarial training with 50\% minibatch being used for generating adversarial samples.
All models are evaluated against 10 attack iterations of the respective attack.}

\begin{table*}[htb]
    \centering
    \caption{\textcolor{black}{Evaluating the adversarial performance of models on PASCAL VOC2012 validation dataset that are adversarially trained using PASCAL VOC2012 training dataset. ``Training method" specifies the adversarial attack used during training, such that ``Clean" stands for no adversarial attack being used during training. During training, 3 attack iterations were used for all adversarial attacks with $\alpha$=0.01 and $\epsilon\approx\frac{8}{255}$. These models were evaluated against multiple adversarial attacks denoted by ``Attack method". We observe that models trained with CosPGD substantially outperform all the other adversarial training methods.}}
    \scalebox{0.57}{
    \begin{tabular}{p{1.5cm}cccc|cc|cc|cc|cc|cc}
\toprule
 \multirow{3}{3cm}{\textbf{Network}} & \multirow{3}{3cm}{\textbf{Training method}} & \multirow{3}{3cm}{\textbf{Attack method}} &  \multicolumn{12}{c}{\textbf{Attack iterations}} \\
& & & \multicolumn{2}{c}{3} & \multicolumn{2}{c}{5} & \multicolumn{2}{c}{10}  &  \multicolumn{2}{c}{20} &  \multicolumn{2}{c}{40} &  \multicolumn{2}{c}{100} \\
 &  & & mIoU(\%) & mAcc(\%) &  mIoU(\%) & mAcc(\%) & mIoU(\%) & mAcc(\%) & mIoU(\%) & mAcc(\%) & mIoU(\%) & mAcc(\%)  & mIoU(\%) & mAcc(\%)\\
\midrule
    \multirow{12}{*}{UNet} & Clean & \multirow{4}{*}{PGD} & 23.18 & 46.64 & 14.58 & 35.89 & 8.21 & 24.99 & 5.57 & 18.57 & 4.14 & 14.53 & 3.6 & 11.72  \\
    
    & PGD & & 29.26 & 57.52 & 21.28 & 51.06 & 13.74 & 41.57 & 9.29 & 32.51 & 7.47 & 27.46 & 6.38 & 22.43 \\

    & SegPGD & & 31.77 & 63.91  & 22.77 & 57.82 & 14.86 & 48.09 & 11.03 & 40.25 & 8.98 & 34.29 & 7.45 & 28.4 \\
    
    & \textbf{CosPGD} & & \textbf{47.35} & \textbf{68.67} & \textbf{43.75} & \textbf{66.34} & \textbf{38.1} & \textbf{62.85} & \textbf{34.33} & \textbf{60.06} & \textbf{32.28} & \textbf{58.64} & \textbf{30.55} & \textbf{57.51} \\
\cmidrule{2-15}
    & Clean & \multirow{4}{*}{SegPGD} &12.38 & 32.41 & 7.75 & 25.27 & 4.46 & 18.36  & 2.98  & 14.24 & 2.20 & 11.66 & 1.55 & 8.66 \\
    
    & PGD & & 29.38 & 57.82 & 21.31 & 51.35 & 13.77 & 41.72 & 9.39 & 33.15 & 7.45 & 26.98 & 6.38 & 22.26 \\
    
    & SegPGD & & 31.69 & 63.94 & 22.47 & 57.07 & 14.82 & 47.94 & 10.9 & 40.32 & 9.09 & 34.68 & 7.33 & 27.99 \\
    
    & \textbf{CosPGD} & & \textbf{47.16} & \textbf{68.51} & \textbf{43.85} & \textbf{66.41} & \textbf{37.64} & \textbf{62.58} & \textbf{33.99} & \textbf{59.8} & \textbf{31.91} & \textbf{58.31} & \textbf{30.48} & \textbf{57.01} \\
\cmidrule{2-15}
    & Clean & \multirow{4}{*}{CosPGD} & 9.67  & 29.46 & 3.71 & 15.89  & 0.61 & 3.39 & 0.06 & 0.38 & 0.03 & 0.16 & 0.01 & 0.04   \\
    
    & PGD & & 29.23 & 57.71 & 21.09 & 50.73 & 13.49 & 40.91 & 9.28 & 32.68  & 7.36 & 27.02 & 6.29 & 22.0 \\
    
    & SegPGD & & 31.53 & 63.96 & 22.46 & 57.23 & 14.81 & 48.09 & 10.86 & 40.26 & 9.20 & 35.33 & 7.28 & 28.03 \\
    
    & \textbf{CosPGD} & & \textbf{47.07} & \textbf{68.39} & \textbf{43.95} & \textbf{66.52} & \textbf{37.64} & \textbf{62.38} & \textbf{34.01} & \textbf{60.03} & \textbf{32.0} & \textbf{58.47} & \textbf{30.55} & \textbf{57.28} \\

\midrule
    \multirow{12}{*}{DeepLabV3}         
    & Clean & \multirow{4}{*}{PGD}    & 11.02 & 30.96 &  8.50 & \textbf{27.34} &  7.63 & \textbf{26.35} &  7.57 & \textbf{26.30} &  7.59 & \textbf{26.19} &  7.39 & \textbf{25.98} \\
    & PGD &                           & 21.05 & 29.07 & 16.74 & 24.61 & 14.45 & 22.19 & 13.82 & 21.56 & 13.58 & 21.32 & 13.42 & 21.17 \\
    & SegPGD &                        & 22.67 & 31.87 & 17.85 & 26.99 & 15.21 & 24.26 & 14.42 & 23.47 & 14.11 & 23.16 & 13.90 & 22.93 \\
    & CosPGD &                        & \textbf{23.13} & \textbf{32.21} & \textbf{18.33} & \textbf{27.34} & \textbf{15.68} & 24.60 & \textbf{14.80} & 23.61 & \textbf{14.49} & 23.29 & \textbf{14.27} & 23.06 \\

\cmidrule{2-15}
    & Clean & \multirow{4}{*}{SegPGD} &  6.78 & 20.50 &  5.05 & 17.40 &  3.99 & 14.95 &  3.32 & 12.94 &  2.60 & 10.57 &  1.80 &  8.05 \\
    & PGD &                           & 20.62 & 28.54 & 16.12 & 23.79 & 13.95 & 21.42 & 13.41 & 20.84 & 13.20 & 20.61 & 13.04 & 20.42 \\
    & SegPGD &                        & 22.06 & 31.37 & 16.89 & 26.02 & 14.27 & \textbf{23.23} & 13.57 & \textbf{22.50} & 13.33 & \textbf{22.23} & 13.09 & 21.92 \\
    & {CosPGD} &                      & \textbf{22.33} & \textbf{31.48} & \textbf{17.15} & \textbf{26.07} & \textbf{14.54} & 23.18 & \textbf{13.89} & 22.45 & \textbf{13.67} & 22.22 & \textbf{13.54} & \textbf{22.15} \\
 
\cmidrule{2-15}
    & Clean & \multirow{4}{*}{CosPGD} &  4.71 & 16.35 &  1.94 &  8.09 &  0.61 &  3.32 &  0.24 &  1.59 &  0.09 &  0.53 &  0.08 &  0.59 \\
    & PGD &                           & 20.56 & 28.48 & 16.05 & 23.75 & 13.87 & 21.45 & 13.38 & 20.92 & 13.18 & 20.72 & \textbf{13.07} & 20.59 \\
    & SegPGD &                        & 21.87 & 31.19 & 16.62 & 25.77 & 13.91 & 22.93 & 13.19 & 22.17 & 12.92 & 21.87 & 12.78 & 21.72 \\
    & {CosPGD} &                      & \textbf{22.14} & \textbf{31.33} & \textbf{16.88} & \textbf{25.85} & \textbf{14.18} & \textbf{22.99} & \textbf{13.48} & \textbf{22.21} & \textbf{13.20} & \textbf{21.90} & 13.05 & \textbf{21.76} \\

\bottomrule
    \end{tabular}  
    }
    \label{table:adv_training_unet}
\end{table*}

\FloatBarrier

\section{Optical flow estimation}
\label{subsec:appendix:optical}
\subsection{Tabular Results}
\label{subsec:appendix:optical_tabular}
\begin{figure}[t]
    \centering 
  \includegraphics[width=\linewidth]{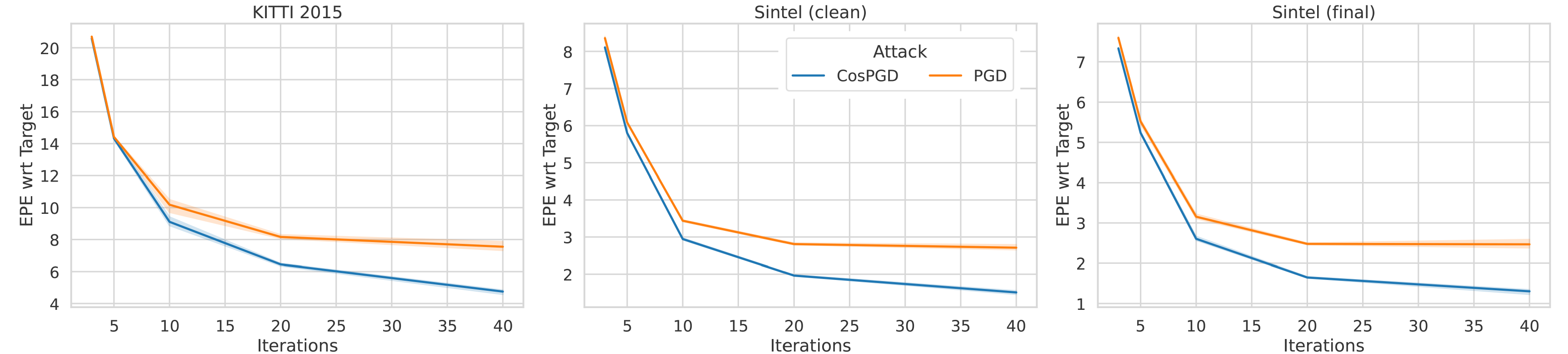}
\caption{An extension to \Cref{fig:optical:target:l_infinity}. Comparison of performance of CosPGD to PGD for optical flow estimation over KITTI-2015~(left) and Sintel~(clean $\rightarrow$ left and final $\rightarrow$ right) validation datasets as $\ell_{\infty}$-norm constrained targeted attacks using RAFT.
CosPGD is a stronger targeted attack than PGD for optical flow. We also report these results in \autoref{tbl:exp:optical_pgd} in Appendix~\ref{subsec:appendix:optical}. \label{fig:optical:target:l_infinity_large}}
\end{figure}
\begin{table*}[ht]
\caption{Comparison of performance of CosPGD to PGD  as a targeted attack for optical flow estimation over KITTI15 and Sintel validation datasets using RAFT for different numbers of attack iterations. $epe$ values are compared, with respect to both, the \textbf{Target} i.e.~$\overrightarrow{0}$ where a lower $epe$ indicates a better attack and Initial flow prediction (optical flow estimated by the model before any adversarial attack) where a higher $epe$ indicates a better attack. CosPGD and PGD perform similarly for a low number of iterations, where CosPGD fits the target slightly better. CosPGD significantly outperforms PGD from the $10^{th}$ iteration onwards on both metrics.}
\label{tbl:exp:optical_pgd}
\centering
\scalebox{.6}{
\begin{tabular}{@{}p{1.5cm}cc|cc|cc||cc|cc|cc|cc|cc|cc@{}}
\toprule
\textbf{Attack} &  \multicolumn{6}{c}{\textbf{KITTI 2015}} &  \multicolumn{12}{c}{\textbf{MPI Sintel}} \\
 & & & & & & & \multicolumn{6}{c|}{\textbf{clean}}  & \multicolumn{6}{c}{\textbf{final}}\\
\multirow{2}{3cm}{\textbf{Iterations}} & \multicolumn{2}{c}{\textcolor{black}{SegPGD}} &  \multicolumn{2}{c}{PGD} & \multicolumn{2}{c||}{CosPGD} & \multicolumn{2}{c}{\textcolor{black}{SegPGD}} &  \multicolumn{2}{c}{PGD}  &  \multicolumn{2}{c|}{CosPGD} & \multicolumn{2}{c}{\textcolor{black}{SegPGD}} &  \multicolumn{2}{c}{PGD}  &  \multicolumn{2}{c}{CosPGD} \\
& Target\textcolor{green}{$\downarrow$} & Initial\textcolor{green}{$\uparrow$} &  Target\textcolor{green}{$\downarrow$} & Initial\textcolor{green}{$\uparrow$} & 
Target\textcolor{green}{$\downarrow$} & Initial\textcolor{green}{$\uparrow$} &  Target\textcolor{green}{$\downarrow$} & Initial\textcolor{green}{$\uparrow$} & Target\textcolor{green}{$\downarrow$} & Initial\textcolor{green}{$\uparrow$} & Target\textcolor{green}{$\downarrow$} & Initial\textcolor{green}{$\uparrow$} & Target\textcolor{green}{$\downarrow$} & Initial\textcolor{green}{$\uparrow$}& Target\textcolor{green}{$\downarrow$} & Initial\textcolor{green}{$\uparrow$} & Target\textcolor{green}{$\downarrow$} & Initial\textcolor{green}{$\uparrow$} \\
\midrule
\textbf{3} & \textbf{20.57} & 11.28 & 20.7  & \textbf{11.4}  & 20.6  & 11.2  & 8.35  &  \textbf{6.83}  & 8.3  & 6.8  &  \textbf{8.1} & 6.6  & 7.58 & \textbf{7.52}  &  7.6 & 7.3 & \textbf{7.5} & 7.3 \\
\textbf{5} & 14.33 &  17.75 & 14.4  &  \textbf{17.8} & \textbf{14.3}  & 17.7  & 6.06 & 8.97   & 6.1  & \textbf{9.0}  & \textbf{5.8}  & 8.8  & 5.44 &  \textbf{9.43} & 5.6  & 9.4& \textbf{5.2}  & 9.3 \\
\textbf{10} & 11.08 &  21.36  &  10.5 &  22.1 & \textbf{9.0}  & \textbf{23.4} & 3.51 &  11.16  & 3.4  & 11.2  &  \textbf{2.9} & \textbf{11.4} & 3.13 &  11.32  & 3.1  & 11.3 & \textbf{2.6} & \textbf{11.5} \\
\textbf{20} & 7.76 & 24.55  & 8.1  & 24.6  &  \textbf{6.5} & \textbf{25.8} & 2.97 & 11.61   & 2.8  & 11.7  & \textbf{2.0}  &  \textbf{12.1} & 2.62 & 11.7  &  2.5 & 11.8 & \textbf{1.6} & \textbf{12.1} \\
\textbf{40} & 7.53 & 24.89  & 7.3  & 25.0  & \textbf{4.8}  & \textbf{27.4}  & 2.66 & 11.8   & 2.8  &  11.7 &  \textbf{1.6} &  \textbf{12.4} & 2.4 & 11.83  & 2.6  & 12.3 & \textbf{1.3} & \textbf{12.3} \\

\bottomrule

\end{tabular}
}
\end{table*}
Here we report the extended results from Figure~\ref{fig:optical:target:l_infinity} comparing CosPGD to PGD as a targeted attack using RAFT for KITTI15 and Sintel datasets in \Cref{fig:optical:target:l_infinity_large}  and in tabular form in Table~\ref{tbl:exp:optical_pgd}.
We observe that CosPGD is more effective than PGD to change the predictions toward the targeted prediction.
During a low number of iterations (iterations = 3 and 5), PGD is on par with CosPGD in increasing the $epe$ values of the predictions compared to the initial predictions on non-attacked images.
However, as the number of iterations increases, CosPGD outperforms PGD for this metric as well.
In the following, we report further results and compare CosPGD to a recently proposed sophisticated $l_2$-norm constrained targeted attack PCFA.

\subsection{Non-targeted attacks for optical flow estimation}
\label{subsec:appendix:limitations:l_inf_kitti}
\begin{figure}
    \centering
    \includegraphics[width=0.475\textwidth]{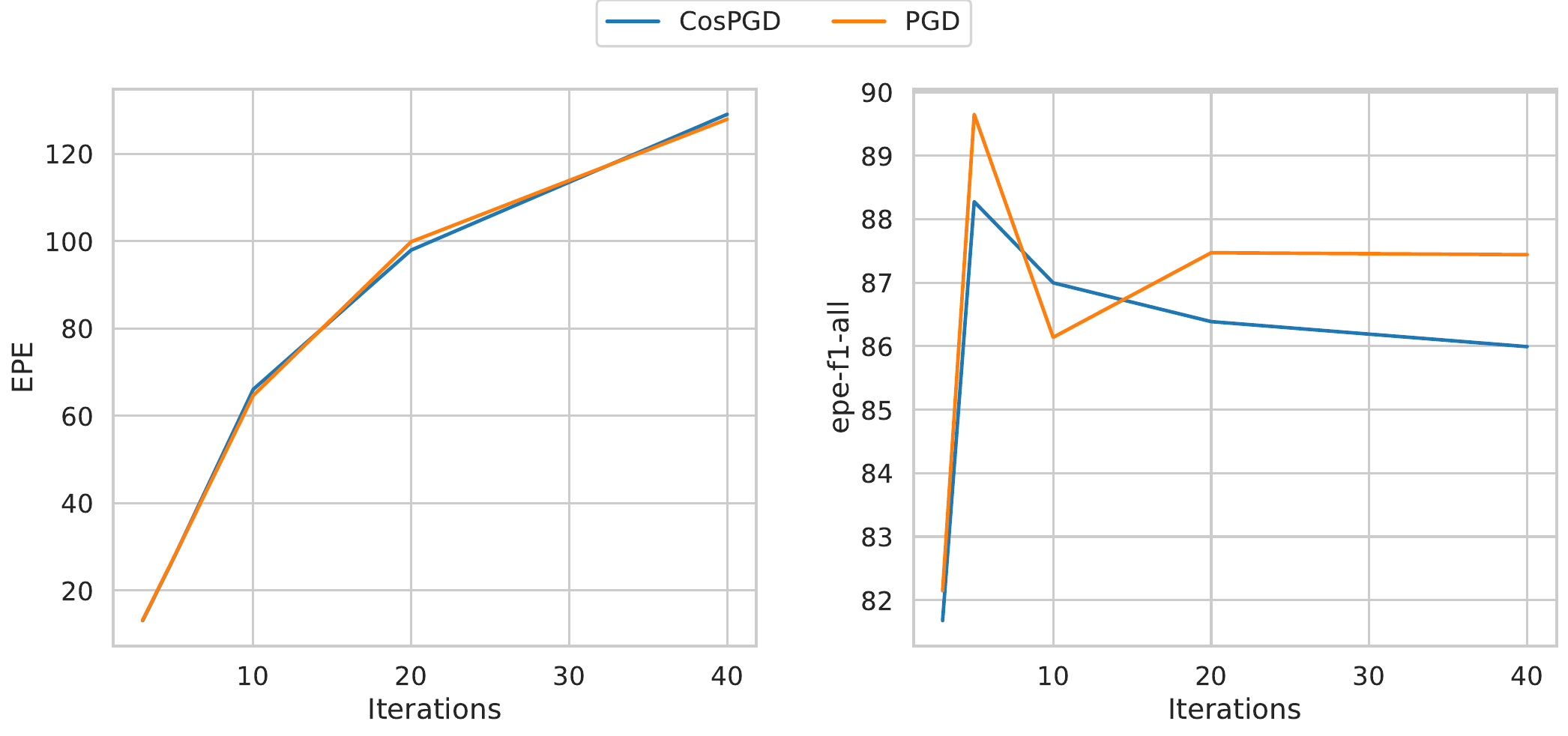}
    \caption{Comparing CosPGD and PGD as $l_{\infty}$-norm constrained non-targeted attacks for optical flow estimation using RAFT on KITTI 2015 validation dataset.}
    \label{fig:kitti_nontargeted}
\end{figure}

\begin{figure*}[t]
    \centering 

   \begin{tabular}{c@{}c@{\hspace{0.1cm}}}
    PGD attack & CosPGD attack \\

  \includegraphics[width=0.5\linewidth]{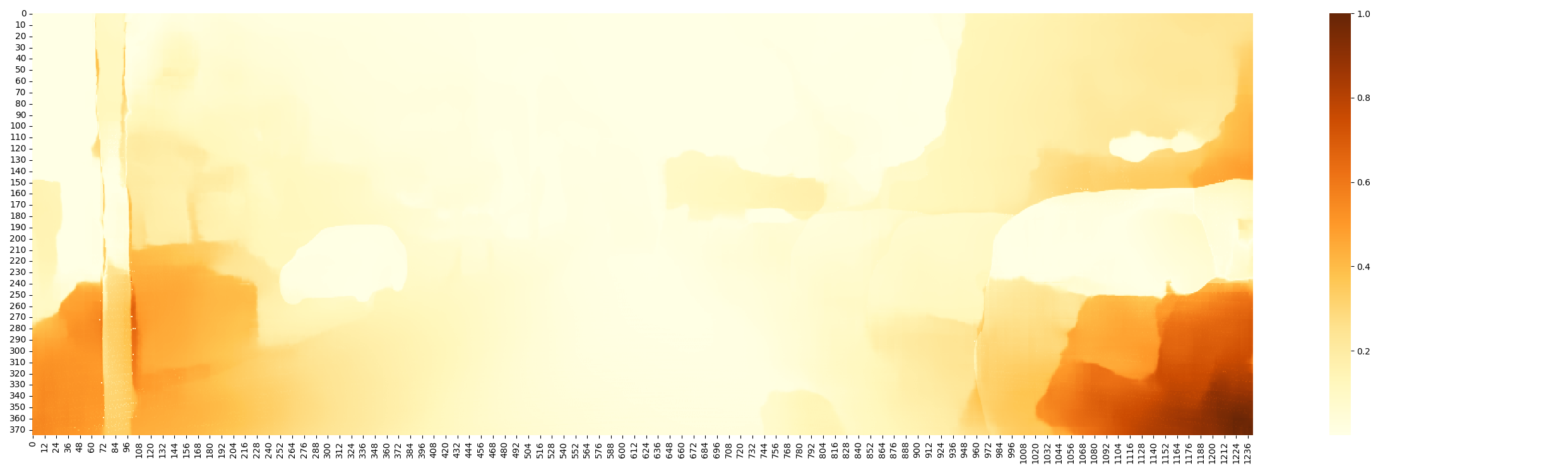} &

  \includegraphics[width=0.5\linewidth]{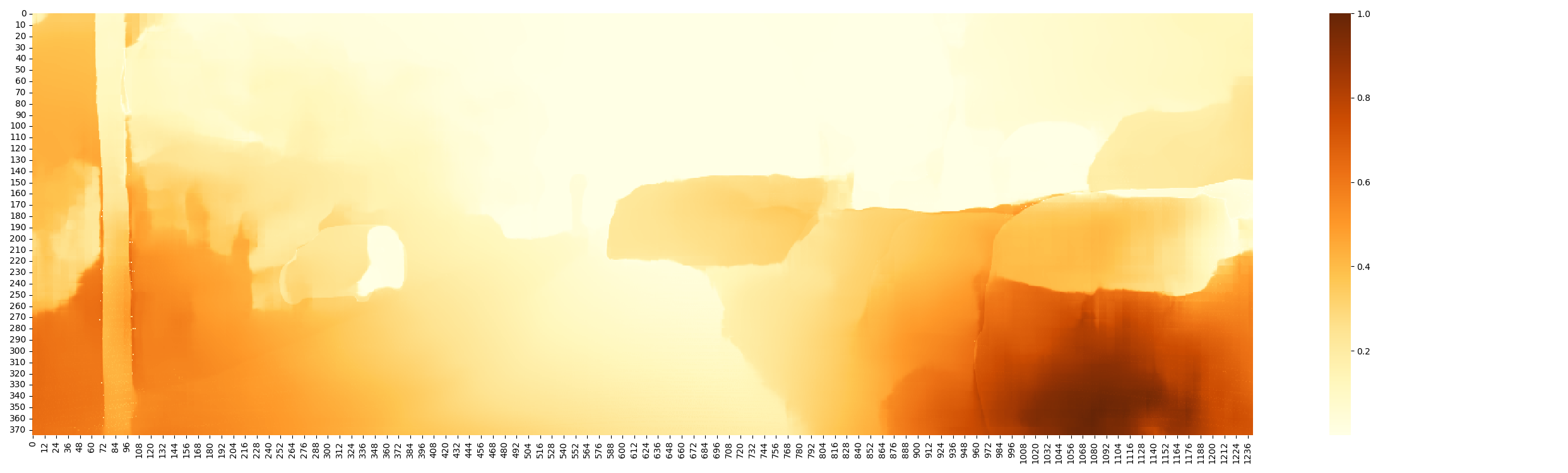}\\

  \includegraphics[width=0.5\linewidth]{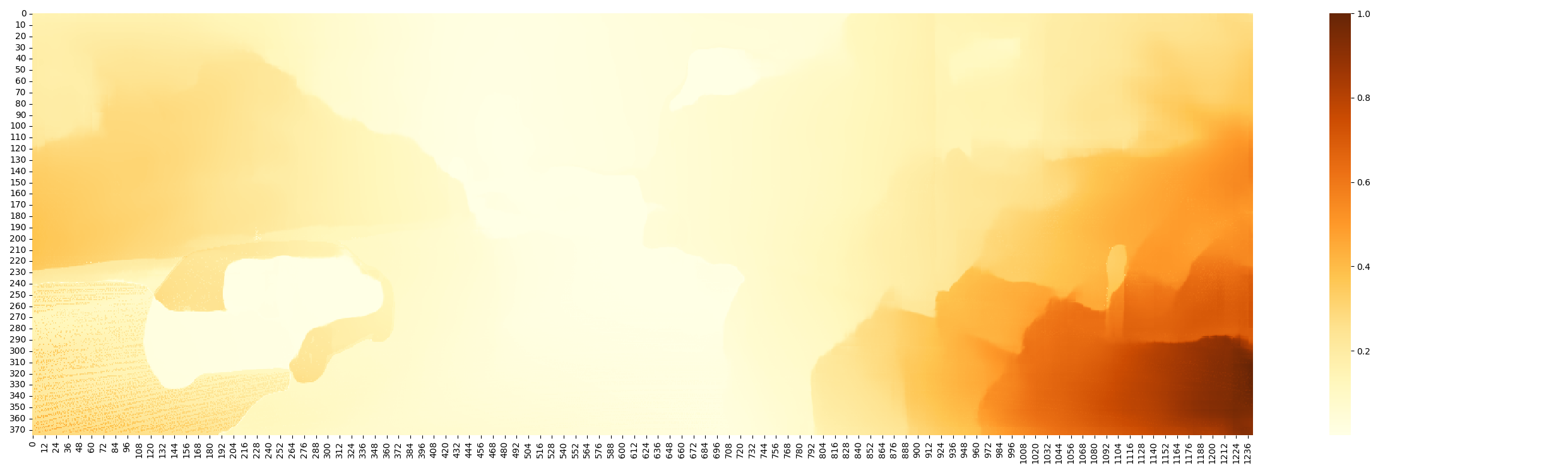} &

  \includegraphics[width=0.5\linewidth]{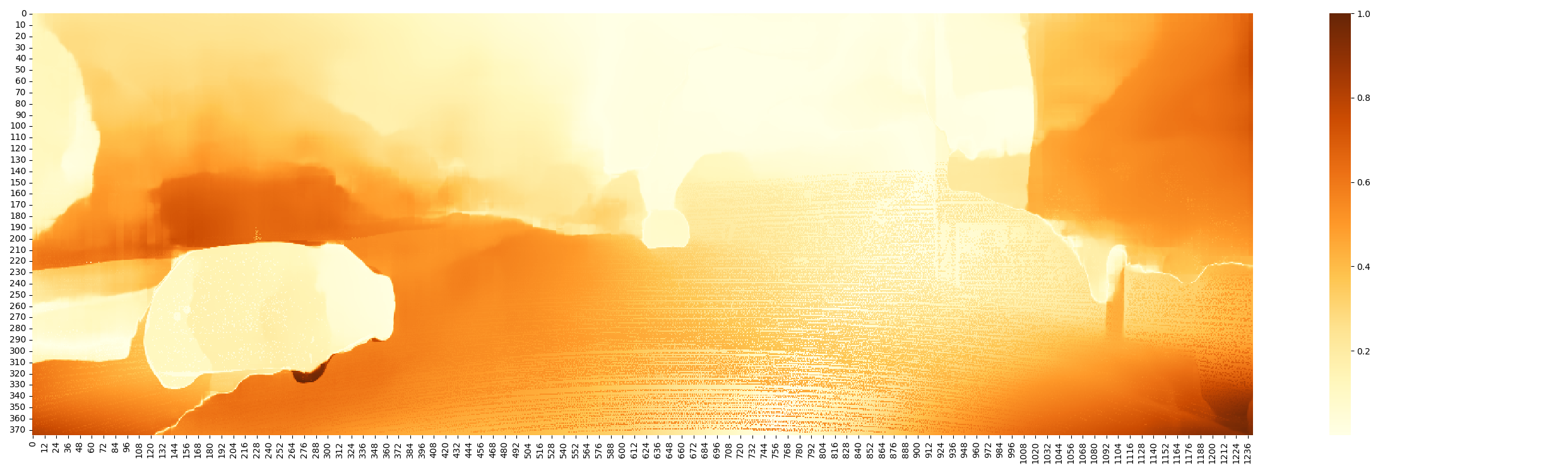}\\

  \includegraphics[width=0.5\linewidth]{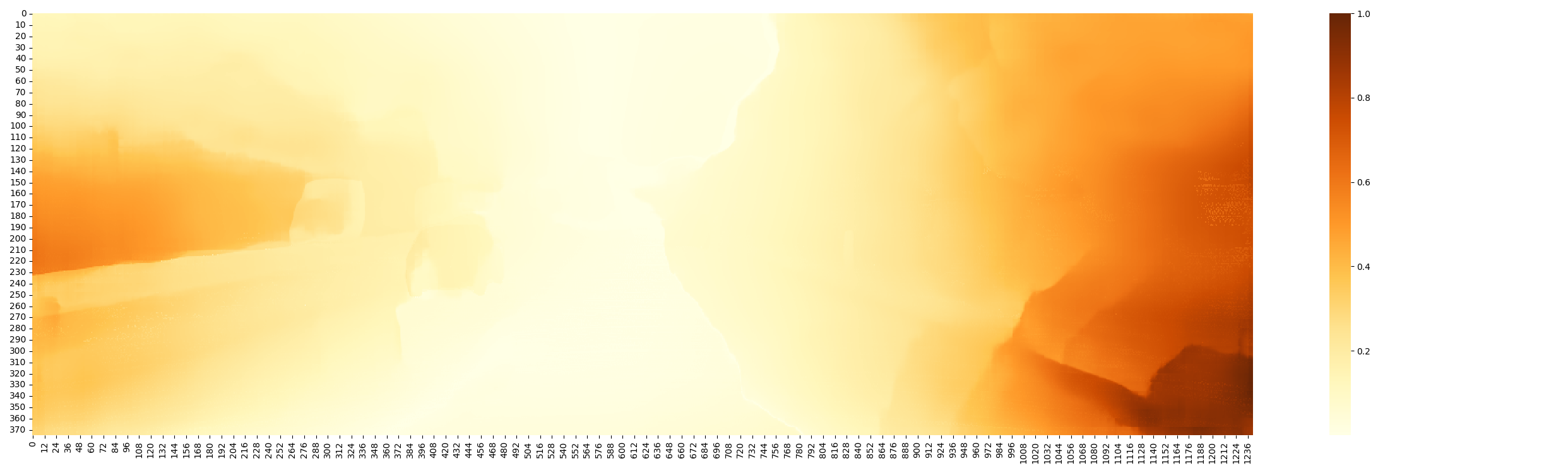} &

  \includegraphics[width=0.5\linewidth]{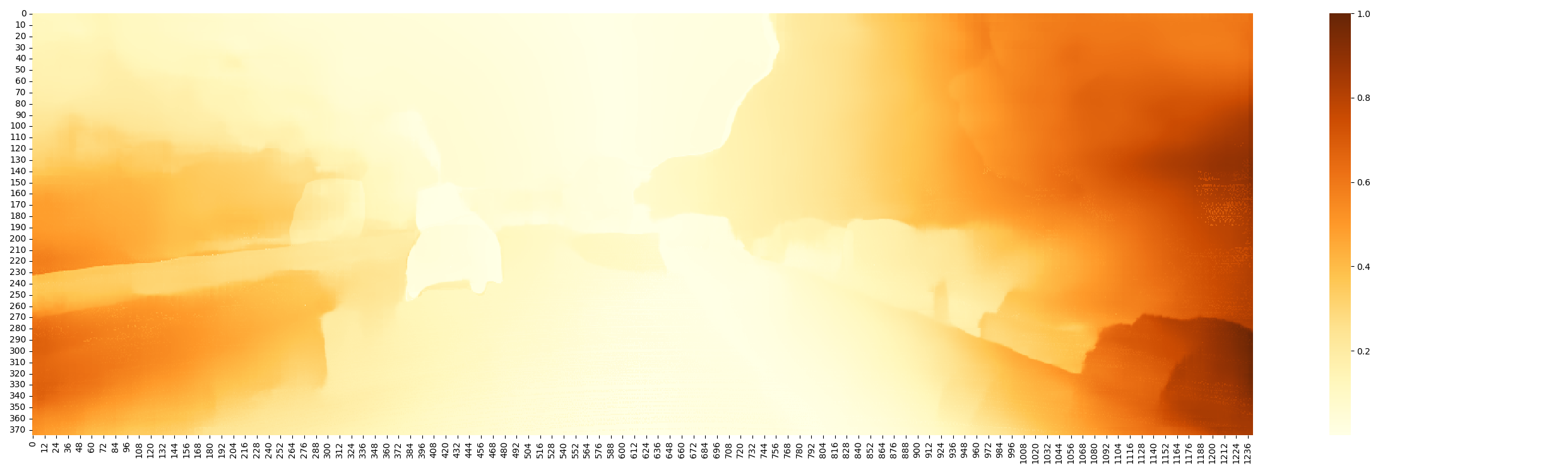}\\

\end{tabular}

\caption{Comparing change in pixel-wise $epe$ values w.r.t. initial $epe$ values after 40 iterations of PGD and CosPGD as non-targeted $\ell_{\infty}$-norm constrained attacks on RAFT using KITTI15 validation set. The values for each image are: $\frac{|epe_{adv}-epe_{initial}|}{max(epe_{adv})}$ where $epe_{adv}$ \& $epe_{initial}$ are pixel-wise $epe$ values of the final adversarial sample and the initial non-attacked image, respectively.}
\label{fig:kitti_non_targeted_epes}
\end{figure*}

For $l_{\infty}$-norm constrained non-targeted attacks, CosPGD changes pixels values temperately over a larger region of the image, while PGD changes it drastically but only for a small region in the image.
This can be observed in Figure~\ref{fig:kitti_nontargeted} when CosPGD and PGD are compared as $l_{\infty}$-norm constrained non-targeted attacks for optical flow estimation.
We observe that both CosPGD and PGD are performing at par as both have very similar $epe$ values across iterations.
However, CosPGD across iterations has a lower $epe$-$f1$-$all$ value.
As shown by Equation~\ref{eqn:out} in Section~\ref{subsec:appendix:epe_f1_all}, $epe$-$f1$-$all$ is the measure of average overall $epe$ values that are above a modest threshold.
Therefore, both CosPGD and PGD have very similar $epe$ scores while CosPGD has a significantly lower $epe$-$f1$-$all$ compared to PGD.
This implies that CosPGD and PGD are performing at par, however, PGD is drastically changing $epe$ values at certain pixels, while CosPGD is changing $epe$ values temperately over considerably more pixels.
Figure~\ref{fig:kitti_non_targeted_epes} shows this qualitatively for 4 randomly chosen samples.

\subsection{Comparison to PCFA}
\label{subsec:appendix:PCFA}
Further, we compare CosPGD as a $l_2$-norm constrained targeted attack to the recently proposed \emph{state-of-the-art} $l_2$-norm constrained targeted attack PCFA~\citep{pcfa}.
For comparison. we use the same settings as those used by the authors for both attacks, for 20 attack iterations (steps), generating adversarial patches for each image individually, bounded under the change of variables methods proposed by \citet{pcfa}.
Here, we observe that a sophisticated $l_2$-norm constrained targeted attack, PCFA that does not utilise pixel-wise information for generating adversarial patches over all considered networks and datasets, performs similar to CosPGD.
We compare over the performance over RAFT, PWCNet~\citep{sun2018pwcnet}, GMA~\citep{gma_jiang2021learning} and SpyNet~\citep{spynet2017}
We consider both targeted settings proposed by \citet{pcfa}, i.e. target being a zero vector $\overrightarrow{0}$ and target being the negative of the initial prediction (\emph{negative flow}).
We compare the average $epe$ over all images.
A lower $AEE$ is w.r.t. Target and higher $AEE$ w.r.t. initial indicate a stronger attack.
In Table~\ref{tbl:pcfa_cospgd_kitti15}(currently included at the end of the appendix to not disturb the table numbers), we compare PCFA and CosPGD on multiple datasets, multiple networks over 3 random seeds.

\begin{figure*}
    \centering
    \includegraphics[width=\textwidth]{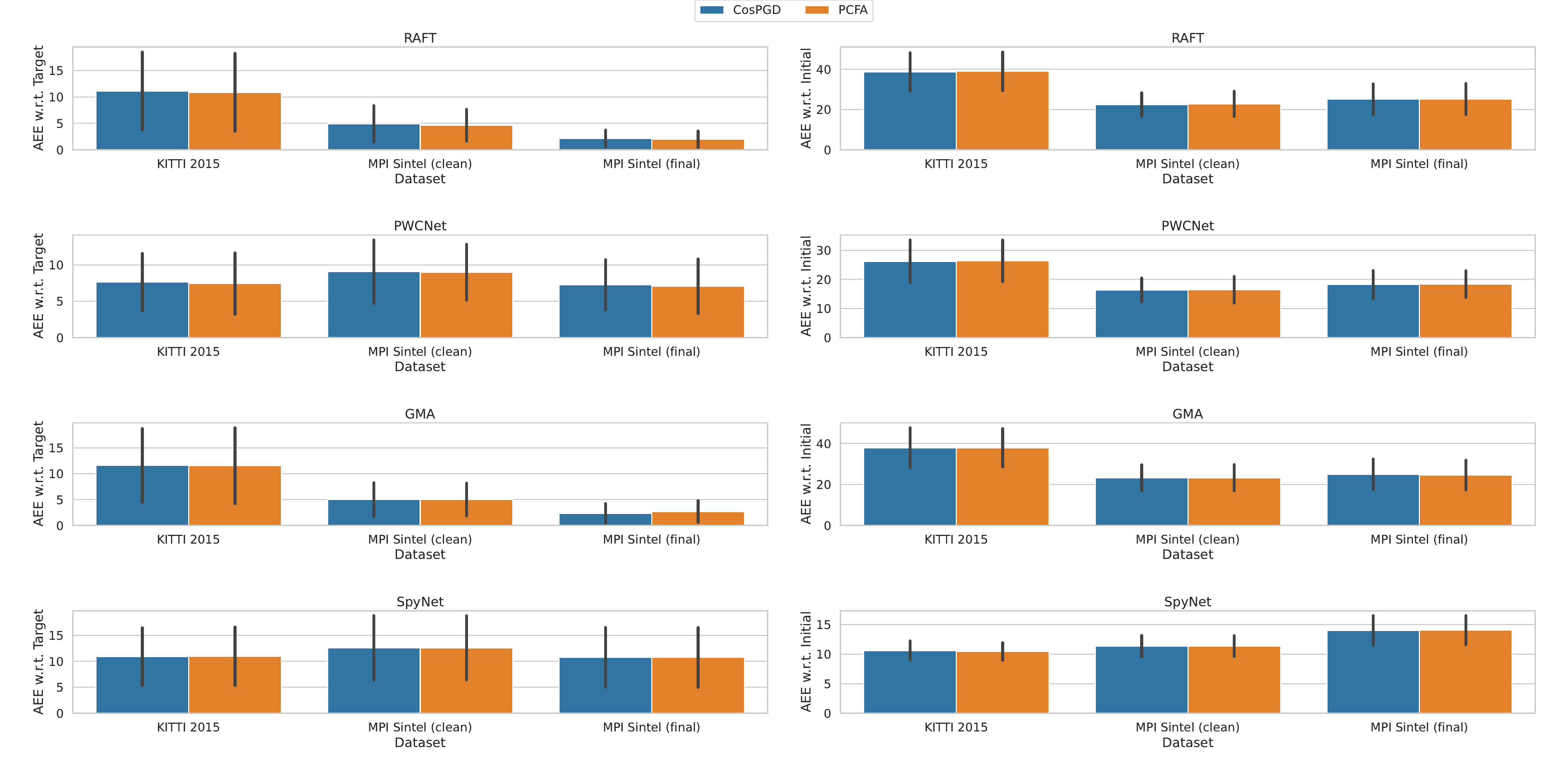}\\
    AEE w.r.t.~Target, lower is better \hspace{2cm} AEE w.r.t.~Initial, higher is better
    \caption{Comparison of mean and standard deviation of the results using different targets, $\overrightarrow{0}$ and \emph{negative flow} for CosPGD and PCFA. A lower $AEE$ is w.r.t. Target and a higher $AEE$ w.r.t. initial indicate a stronger attack.}
    \label{fig:pcfa_both}
\end{figure*}

\begin{figure*}
    \centering
    \includegraphics[width=\textwidth]{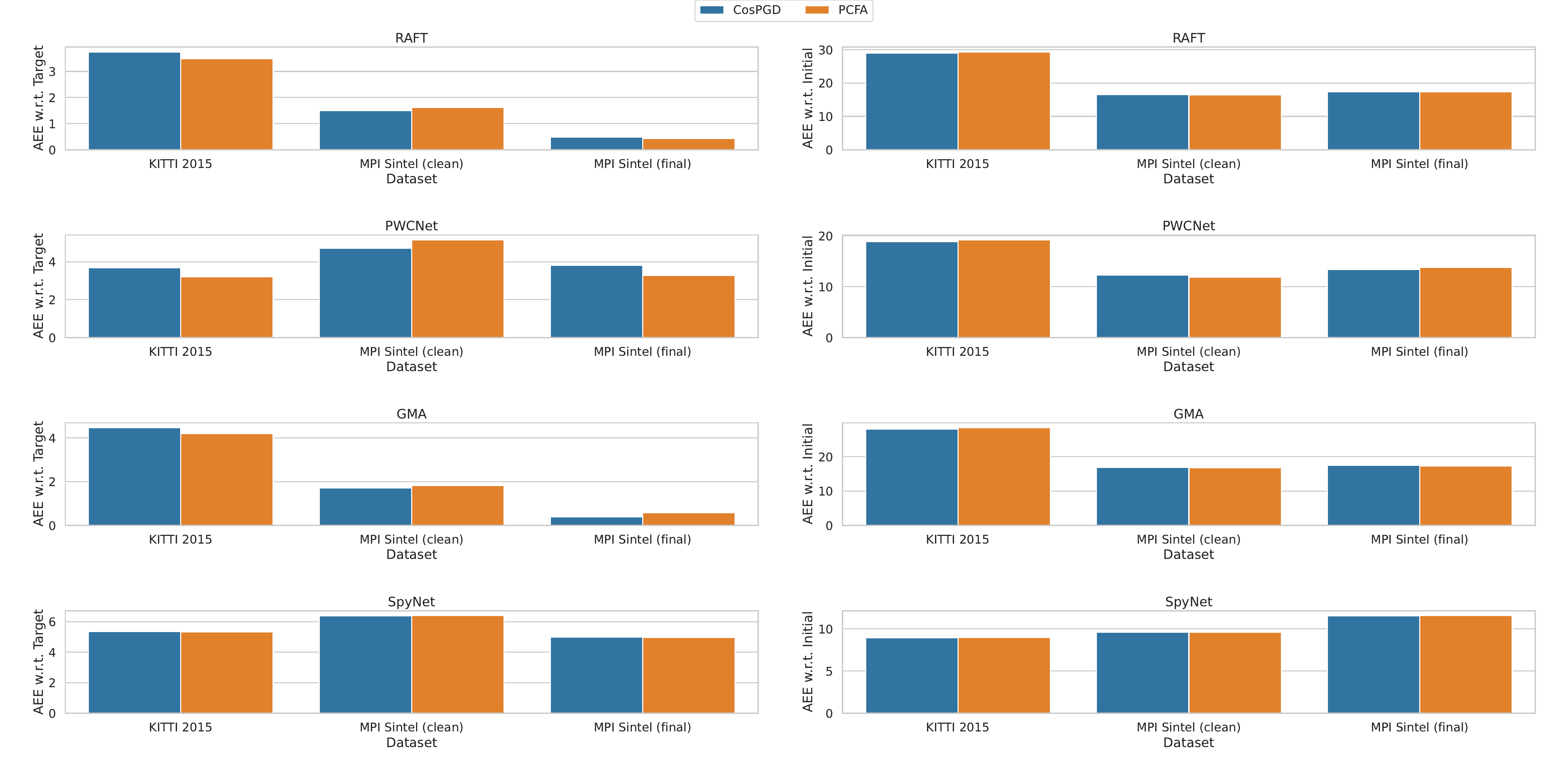}
    \\
    AEE w.r.t.~Target, lower is better \hspace{2cm} AEE w.r.t.~Initial, higher is better
    \caption{Comparison of PCFA and CosPGD when using $\overrightarrow{0}$ as the target. A lower $AEE$ is w.r.t. Target and a higher $AEE$ w.r.t. initial indicate a stronger attack.}
    \label{fig:pcfa_zero}
\end{figure*}

\begin{figure*}
    \centering
    \includegraphics[width=\textwidth]{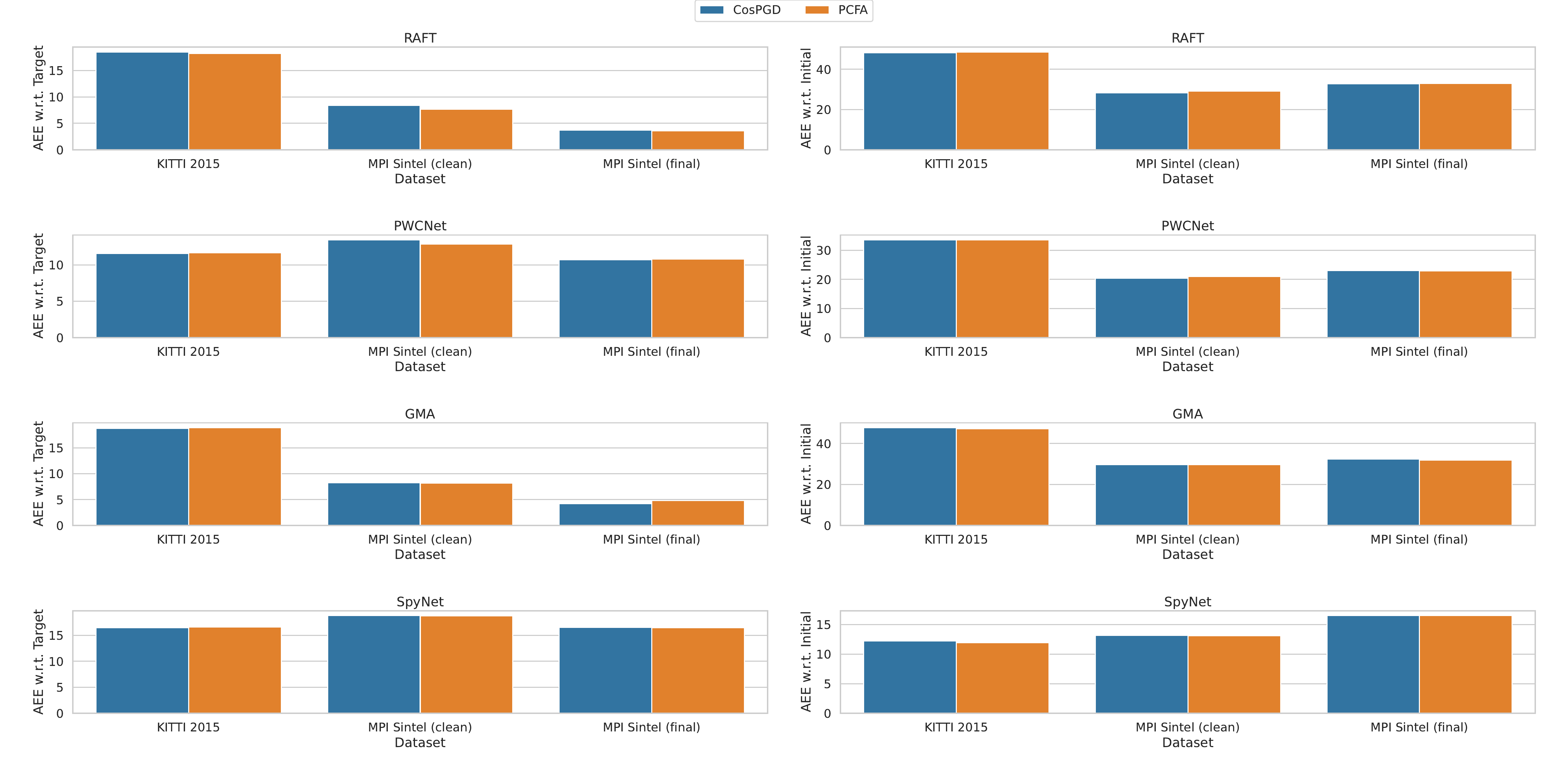}
    \\
    AEE w.r.t.~Target, lower is better \hspace{2cm} AEE w.r.t.~Initial, higher is better
    \caption{Comparison of PCFA and CosPGD when using \emph{negative flow} as the target. A lower $AEE$ is w.r.t. Target and a higher $AEE$ w.r.t. initial indicate a stronger attack.}
    \label{fig:pcfa_neg_flow}
\end{figure*}

Figure~\ref{fig:pcfa_both}, provides an overview of the comparison between the two methods, using targets as $\overrightarrow{0}$ and \emph{negative flow}.
Figures~\ref{fig:pcfa_zero},~\ref{fig:pcfa_neg_flow}, provide further details compares both methods when using $\overrightarrow{0}$ and \emph{negative flow} as the target, respectively.

In \Cref{tbl:pcfa_cospgd_kitti15}, we include the results in a tabular form.
\begin{table*}[h]
\caption{Comparison of performance of CosPGD to PCFA as a targeted $l_2$-norm constrained attack for optical flow estimation over KITTI2015 and Sintel validation datasets using different optical flow models over 3 random seeds. Average $epe$ values are compared, with respect to both, the \textbf{Target}  where a lower $epe$ indicates a better attack and \textbf{Initial flow prediction} (optical flow estimated by the model before any adversarial attack) where a higher $epe$ indicates a better attack. 
We compare over both targets used by \cite{pcfa}, i.e. zero vector ~$\overrightarrow{0}$ and Negative of the Initial Flow.
\textbf{CosPGD and PCFA performance is very comparable.}
}
\label{tbl:pcfa_cospgd_kitti15}
\centering
\scalebox{.72}{
\begin{tabular}{@{}p{1.5cm}cccc|cccc@{}}

\toprule


\multirow{3}{3cm}{\textbf{Model}} &  \multicolumn{4}{c}{Target $\overrightarrow{0}$} & \multicolumn{4}{c}{Negative Initial Flow} \\

 &  \multicolumn{2}{c}{AEE wrt Target\textcolor{green}{$\downarrow$}} & \multicolumn{2}{c}{\textcolor{black}{AEE wrt Initial\textcolor{green}{$\uparrow$}}} &  \multicolumn{2}{c}{AEE wrt Target\textcolor{green}{$\downarrow$}} & \multicolumn{2}{c}{AEE wrt Initial\textcolor{green}{$\uparrow$}}  \\

& CosPGD & PCFA & CosPGD & PCFA & CosPGD & PCFA & CosPGD & PCFA \\
\midrule

& \multicolumn{8}{c}{\textbf{KITTI 2015}}  \\
\midrule

GMA & 28.69 $\pm$ 0.12 &	28.67 $\pm$ 0.17&	3.89 $\pm$ 0.09&	3.89 $\pm$ 0.15 & 47.00 $\pm$ 0.40&	47.08 $\pm$ 0.69&	19.22 $\pm$ 0.53&	19.20 $\pm$ 0.57\\

PWCNet & 19.13 $\pm$ 0.04	&18.96 $\pm$ 0.08	&3.25 $\pm$ 0.08&	3.47 $\pm$ 0.14 & 33.13 $\pm$ 0.25&	33.13 $\pm$ 0.26&	12.01 $\pm$ 0.20&	12.02 $\pm$ 0.22  \\

RAFT & 29.09 $\pm$ 0.03&	29.17 $\pm$ 0.11&	3.75 $\pm$ 0.05&	3.63 $\pm$ 0.10 &  48.83 $\pm$ 0.35&	48.93 $\pm$ 0.29&	17.97 $\pm$ 0.29&	17.81 $\pm$ 0.27 \\

SpyNet & 9.00 $\pm$ 0.01&	9.01 $\pm$ 0.03&	5.31 $\pm$ 0.01&	5.35 $\pm$ 0.06 & 12.10 $\pm$ 0.02&	12.08 $\pm$ 0.05&	16.47 $\pm$ 0.03&	16.44 $\pm$ 0.05 \\

\midrule

& \multicolumn{8}{c}{\textbf{MPI Sintel (clean)}} \\
\midrule

GMA & 16.87 $\pm$ 0.14&	16.76 $\pm$ 0.11&	1.75 $\pm$ 0.15&	1.85 $\pm$ 0.10 & 29.25 $\pm$ 0.38&	29.05 $\pm$ 0.38&	8.58 $\pm$ 0.34&	8.82 $\pm$ 0.37 \\
PWCNet & 12.20 $\pm$ 0.21	& 12.18 $\pm$ 0.07	& 4.87 $\pm$ 0.17 &	4.75 $\pm$ 0.12 & 20.57 $\pm$ 0.21 &	20.43 $\pm$ 0.21 &	13.20 $\pm$ 0.13 &	13.21 $\pm$ 0.29 \\
RAFT &  16.42 $\pm$ 0.03	& 16.46 $\pm$ 0.05 & 	1.69 $\pm$ 0.04 & 	1.65 $\pm$ 0.06 & 29.01 $\pm$ 0.11 & 	29.20 $\pm$ 0.01	& 7.67 $\pm$ 0.11 & 	7.47 $\pm$ 0.05 \\
SpyNet & 9.69 $\pm$ 0.01 &	9.75 $\pm$ 0.07 & 	6.40 $\pm$ 0.05	 & 6.35 $\pm$ 0.00 & 13.08 $\pm$ 0.01 & 13.17 $\pm$ 0.03 & 	18.75 $\pm$ 0.02 &	18.76 $\pm$ 0.06 \\

\midrule

& \multicolumn{8}{c}{\textbf{MPI Sintel (final)}}  \\
\midrule

GMA & 17.34 $\pm$ 0.07&	17.31 $\pm$ 0.11&	0.53 $\pm$ 0.07&	0.54 $\pm$ 0.11 & 32.11 $\pm$ 0.20&	32.04 $\pm$ 0.24&	4.57 $\pm$ 0.22&	4.64 $\pm$ 0.24 \\
PWCNet & 13.61 $\pm$ 0.10	&13.44 $\pm$ 0.14&	3.52 $\pm$ 0.13&	3.66 $\pm$ 0.12 & 23.00 $\pm$ 0.30&	23.01 $\pm$ 0.06&	10.84 $\pm$ 0.28&	10.75 $\pm$ 0.05 \\
RAFT & 17.38 $\pm$ 0.04	& 17.36 $\pm$ 0.03 &	0.55 $\pm$ 0.09 &	0.50 $\pm$ 0.03 & 32.72 $\pm$ 0.22 &	32.72 $\pm$ 0.14 &	3.71 $\pm$ 0.21	 & 3.75 $\pm$ 0.13 \\
SpyNet & 11.56 $\pm$ 0.01	& 11.59 $\pm$ 0.03 &	4.97 $\pm$ 0.01 &	4.97 $\pm$ 0.01 & 16.51 $\pm$ 0.01	& 16.55 $\pm$ 0.06 & 	16.52 $\pm$ 0.01 &	16.47 $\pm$ 0.05 \\
\bottomrule
\end{tabular}
}
\end{table*}

It would be interesting to extend these evaluations to newer optical flow datasets such as Spring~\cite{mehl2023spring}.

\section{Image Restoration Tasks}
\label{subsec:appendix:image_restoration}
Following, we provide further results and discussion on the two considered image restoration tasks namely, Image Deblurring in Section~\ref{subsec:appendix:image_deblurring} and Image Denoising in Section~\ref{subsec:appendix:limitations:l_inf_ssid}

\subsection{\revision{Image Deblurring models}}
\label{subsec:appendix:image_deblurring}
\begin{figure*}
    \centering
 \includegraphics[width=\textwidth]{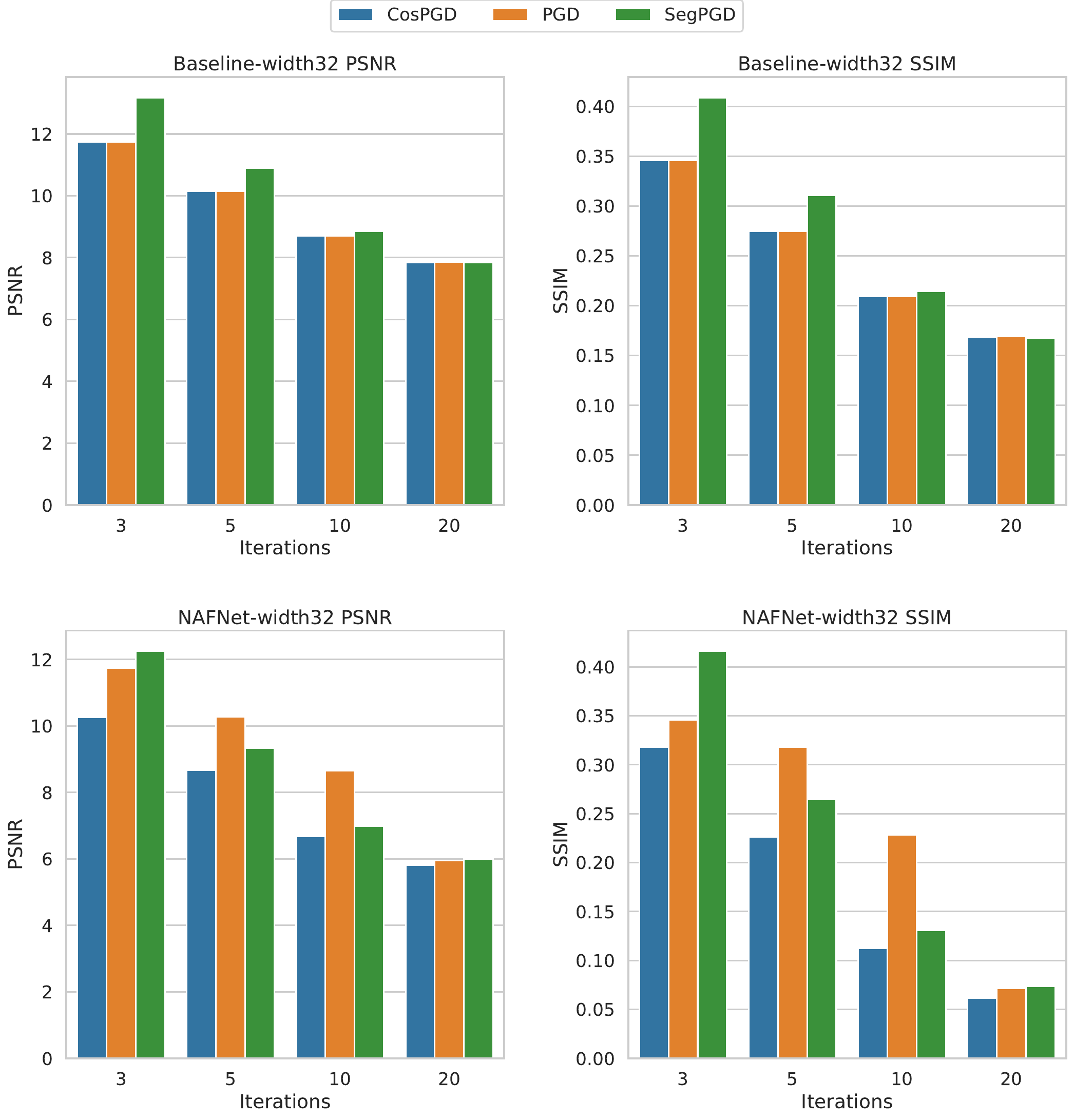}
    \caption{Non-targeted $l_{\infty}$-norm constrained  CosPGD, PGD, and SegPGD attacks on the ``Baseline network" and NAFNet for image deblurring task on the GoPro dataset, recently proposed by \cite{nafnet} as the state-of-the-art networks for image restoration tasks. The ``Baseline network" is significantly more robust than the NAFNet and thus the performance of the Baseline network against CosPGD attack is at par with its performance against PGD.
    However, PGD indicates at low attack iterations~(iterations $\leq$ 10) that NAFNet is more robust than ``Baseline network" and only after 20 attack iterations its correctly indicates that NAFNet is less robust.
    However, CosPGD is able to draw this conclusion at merely 3 attack iterations.
    }
    \label{fig:deblurring}
\end{figure*}
In \autoref{fig:deblurring} for the Baseline network, we observe that both CosPGD and PGD are performing at par.
While for the newly proposed NAFNet, PGD is still estimating NAFNet's adversarial robustness to be very similar to the Baseline network and only after 20 attack iterations it is estimating correctly that NAFNet is not as robust as the Baseline network.
However, CosPGD reveals that NAFNet is not as robust as the baseline even at a low number of iterations~(3 attack iterations). 
This valuable insight regarding model robustness of newly proposed transformer-based image restoration models is provided by CosPGD with considerably less computation.

To enable the applicability of SegPGD on this task, we implement SegPGD by comparing the equality of the pixel values to use their proposed loss for comparison. 
\revision{Following the discussion from Section~\ref{subsec:exp:gopro}, in Figure~\ref{fig:gopro} for the Baseline network we also observe that SegPGD here is significantly weaker due to its limitation to image classification tasks as discussed in Section~\ref{sec:method}.
However, for NAFNet, from 5 attack iterations onwards SegPGD is outperforming PGD, while still being weaker than CosPGD.
This, interesting improvement in the performance of SegPGD as an adversarial attack can be attributed to the pixel-wise nature of the attack, similar to CosPGD further highlighting the benefits of utilizing pixel-wise information when crafting adversarial attacks for pixel-wise prediction tasks.}

\revision{Additionally, we report the findings on many recently proposed state-of-the-art image restoration models using CosPGD in Table~\ref{tab:image_restoration}.}
\begin{table*}[h]
    \centering
    \caption{Comparison of clean and adversarial performance of image reconstruction models, as considered by \cite{agnihotri2023unreasonable}.
    `+ADV' denotes FGSM adversarial training with a 50-50 mini-batch split for generating an adversarial sample.}
    \scalebox{0.61}{
    \begin{tabular}{@{}l|cc|cc|cc|cc|cc|cc|cc@{}}
    \toprule
    \multirow{3}{*}{Architecture} & \multicolumn{2}{c|}{Clean} & \multicolumn{6}{c|}{CosPGD} & \multicolumn{6}{c}{PGD} \\
    & &  & \multicolumn{2}{c|}{5 attack itrs } & \multicolumn{2}{c|}{10 attack itrs } & \multicolumn{2}{c|}{20 attack itrs } & \multicolumn{2}{c|}{5 attack itrs } & \multicolumn{2}{c|}{10 attack itrs } & \multicolumn{2}{c}{20 attack itrs } \\
    & PSNR & SSIM & PSNR & SSIM & PSNR & SSIM & PSNR & SSIM & PSNR & SSIM & PSNR & SSIM & PSNR & SSIM \\
    \toprule
         \textbf{Restormer}\citep{Zamir2021Restormer} & 31.99 & \textbf{0.9635} & 11.36 & 0.3236 & 9.05 & 0.2242 & 7.59 & 0.1548 & 11.41 & 0.3256 & 9.04 & 0.2234 & 7.58 & 0.1543 \\
         ~~~~ + \textbf{ADV} & 30.25 & 0.9453 &\textbf{24.49} & \textbf{0.81} & \textbf{23.48} & \textbf{0.78} & \textbf{21.58} & \textbf{0.7317} & \textbf{24.5} & \textbf{0.8079} & \textbf{23.5} & \textbf{0.7815} & \textbf{21.58}  & \textbf{0.7315} \\
          \midrule
        Baseline\citep{nafnet} & 32.48 & 0.9575 & 10.15 & 0.2745  &  8.71 & 0.2095  &  7.85 & 0.1685  &  10.15 & 0.2745 & 8.71 & 0.2094 & 7.85 & 0.1693 \\
          ~~~~ + ADV & 30.37 & 0.9355 & 15.47 & 0.5216  &  13.75 & 0.4593  &  12.25 & 0.4032  & 15.47 & 0.5215  &  13.75 & 0.4592  & 12.24 & 0.4026   \\
          \midrule
        NAFNet\citep{nafnet} & \textbf{32.87} & 0.9606 & 8.67 & 0.2264 & 6.68 & 0.1127  & 5.81 & 0.0617 & 10.27 & 0.3179  & 8.66 & 0.2282  &  5.95 & 0.0714\\
          ~~~~ + ADV & 29.91 & 0.9291 &  17.33 & 0.6046 & 14.68 & 0.509 &  12.30 & 0.4046  & 15.76 & 0.5228  & 13.91 & 0.4445  & 12.73 & 0.3859 \\
%
    \bottomrule
    \end{tabular}    
    }
    \label{tab:image_restoration}
\end{table*}

\subsection{Non-targeted Attacks for Image Denoising Task}
\label{subsec:appendix:limitations:l_inf_ssid}
\paragraph{Dataset. }For the image denoising task, following work from \cite{nafnet,Zamir2021Restormer} we use the Smartphone Image Denoising Dataset (SSID)~\citep{ssid}.
This dataset consists of 160 noisy images taken from 5 different smartphones and their corresponding high-quality ground truth images.
Similar to the image deblurring task, we report the $PSNR$ and $SSIM$ values as metrics for this image restoration task as well.

\begin{figure*}
    \centering
    \includegraphics[width=\textwidth]{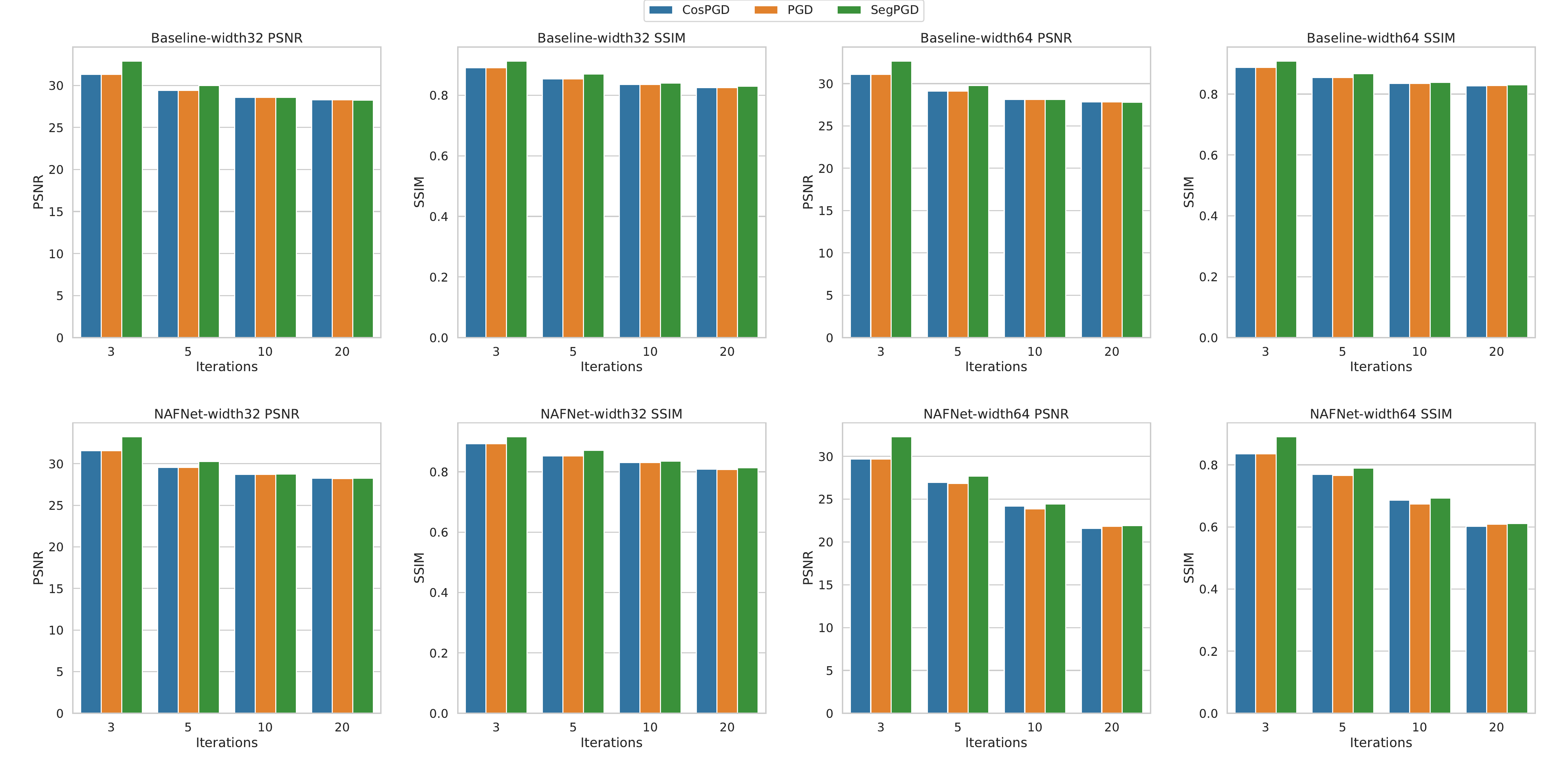}
    \caption{Comparing CosPGD to PGD and SegPGD as $l_{\infty}$-norm constrained non-targeted attacks for the image denoising task using Baseline network~(top row) and NAFNet~(bottom row) on SSID dataset.
            A lower value of PSNR and SSIM indicate a stronger attack.}
    \label{fig:ssid_nontargeted}
\end{figure*}

\paragraph{Discussion. }Further extending the findings from Section~\ref{subsec:appendix:limitations:l_inf_kitti} we report $l_{\infty}$-norm constrained non-targeted attacks for the image denoising on the SSID dataset using the Baseline network and NAFNet~(as proposed by \cite{nafnet}) in Figure.~\ref{fig:ssid_nontargeted}.
We observe that both CosPGD and PGD are performing at par for both, the Baseline network and NAFNet.
Additionally, similar to findings in Section~\ref{subsec:exp:gopro}, SegPGD is unable to perform at par with CosPGD and PGD.

After both CosPGD and PGD attacks it appears that the image denoising networks are relatively more robust than image deblurring networks.
These findings also correlate with \cite{xie2019feature_denoinsing}, as they report that feature denonising improves model robustness against adversarial attacks.

\section{Discussion on limitations of CosPGD}
\label{sec:appendix:limitations}

Similar to most white-box adversarial attacks~\citep{fgsm, pgd, pgdl2, apgd, segpgd}, CosPGD currently requires access to the model's gradients for generating adversarial examples.
While this is beneficial for generating adversaries, it limits the applications of the non-targeted settings as many benchmark datasets~\citep{kitti15, sintel1, sintel2, pascal-voc-2012} do not provide the ground truth for test data. 
Evaluations of the validation datasets certainly show the merit of the attack method. 
CosPGD mitigates this limitation by also being applicable as an effective targeted attack.
Nevertheless, 
it would be interesting to study the attack on test images as well in an untargeted setting, due to the potential slight distribution shifts pre-existing in the test data.
While CosPGD is significantly more efficient than other existing adversarial attacks, all white-box adversarial attacks are time and memory consuming and benchmarking them across multiple downstream tasks, datasets, and networks is a very time-consuming process.

Additionally, there are settings, especially for non-targeted attacks, where approaches like pixel-wise PGD would work at par with CosPGD as the $epe$ can be increased equally well by either changing all pixel-wise regression estimates slightly~(sophisticated attack like CosPGD) or by changing only a few of them drastically~(brute force attacks like PGD).
This can also be seen in the results in \ref{subsec:appendix:limitations:l_inf_kitti}.

\end{document}